\documentclass{article} % For LaTeX2e
\usepackage{iclr2025_conference,times}

\usepackage{amsmath,amsfonts,bm}

\def\eqref#1{equation~\ref{#1}}
\def\1{\bm{1}}

\DeclareMathAlphabet{\mathsfit}{\encodingdefault}{\sfdefault}{m}{sl}
\SetMathAlphabet{\mathsfit}{bold}{\encodingdefault}{\sfdefault}{bx}{n}

\newcommand\ie{\textit{i.e.,}}
\newcommand\eg{\textit{e.g.,}}

\newcommand{\norm}[1]{\left\lVert#1\right\rVert}

\newcommand{\beq}{\begin{equation}}
\newcommand{\eeq}{\end{equation}}
\newcommand{\beqnn}{\begin{equation*}}
\newcommand{\eeqnn}{\end{equation*}}
\newcommand{\beqy}{\begin{eqnarray}}
\newcommand{\eeqy}{\end{eqnarray}}
\newcommand{\beqynn}{\begin{eqnarray*}}
\newcommand{\eeqynn}{\end{eqnarray*}}
\newcommand{\bit}{\begin{itemize}}
\newcommand{\eit}{\end{itemize}}
\newcommand{\ben}{\begin{enumerate}}
\newcommand{\een}{\end{enumerate}}
\newcommand{\bex}{\begin{example}}
\newcommand{\eex}{\end{example}}

\newcommand{\balg}[1]{\begin{algorithm} \caption{#1}}
\newcommand{\ealg}{\end{algorithm}}

\newcommand{\balgc}{\begin{algorithmic}[1]}
\newcommand{\ealgc}{\end{algorithmic}}

\newcommand{\bary}{\begin{array}}
\newcommand{\eary}{\end{array}}
\newcommand{\bmx}{\begin{bmatrix}}
\newcommand{\emx}{\end{bmatrix}}
\newcommand{\bsmx}{\left[\begin{smallmatrix}}
\newcommand{\esmx}{\end{smallmatrix}\right]}
\newcommand{\bmxc}[1]{\left[\begin{array}{@{}#1@{}}}
\newcommand{\emxc}{\end{array}\right]}
\newcommand{\bcn}{\begin{center}}
\newcommand{\ecn}{\end{center}}

\renewcommand{\b}{\mathbf{b}}

\providecommand{\norm}[1]{\lVert#1\rVert}

\providecommand{\abs}[1]{\left| #1 \right|}

\newenvironment{theorem}[2][Theorem]{\begin{trivlist}
		\item[\hskip \labelsep {\bfseries #1}\hskip \labelsep {\bfseries #2.}]}{\end{trivlist}}

\usepackage{hyperref}
\usepackage{url}
\usepackage{graphicx}
\usepackage{booktabs}
\usepackage{array}

\usepackage{multirow}						% tabular cells spanning multiple rows
\usepackage{amsfonts}						% blackboard math symbols
\usepackage{duckuments}						% sample images
\usepackage{subfig}
\usepackage{makecell}
\usepackage{amssymb,bm,amsthm,nccmath,amsfonts,amsmath}
\usepackage{xcolor}
\definecolor{green}{rgb}{0, 0.64, 0.31}
\definecolor{red}{rgb}{1, 0, 0} 
\definecolor{cyan}{rgb}{0, 0.6, 0.8} 
\let\b\boldsymbol

\title{Rethinking Structure Learning For Graph Neural Networks}

\author{
Yilun Zheng$^1$,
Zhuofan Zhang$^2$, 
Ziming Wang$^1$, 
Xiang Li$^3$,
Sitao Luan$^4$ $^*$,
Xiaojiang Peng$^3$,
Lihui Chen$^1$
\thanks{
Corresponding Author.
Email address:
yilun001@e.ntu.edu.sg,
2210413014@email.szu.edu.cn,
sitao.luan@mail.mcgill.ca,
xiaojiangpeng@sztu.edu.cn,
elhchen@ntu.edu.sg. } \\
$^1$Nanyang Technological University, Centre for Info. Sciences and Systems,
$^2$School of Software Technology, \\ Zhejiang University,
$^3$College of Big Data and Internet, Shenzhen University, \\
$^4$Mila - Quebec Artificial Intelligence Institute.
}

\iclrfinalcopy % Uncomment for camera-ready version, but NOT for submission.
\begin{document}

\maketitle

\begin{abstract}

To improve the performance of Graph Neural Networks (GNNs), Graph Structure Learning (GSL) has been extensively applied to reconstruct or refine original graph structures, effectively addressing issues like heterophily, over-squashing, and noisy structures. While GSL is generally thought to improve GNN performance, it often leads to longer training times and more hyperparameter tuning. Besides, the distinctions among current GSL methods remain ambiguous from the perspective of GNN training, and there is a lack of theoretical analysis to quantify their effectiveness. Recent studies further suggest that, under fair comparisons with the same hyperparameter tuning, GSL does not consistently outperform baseline GNNs. This motivates us to ask a critical question: \textit{is GSL really useful for GNNs?} To address this question, this paper makes two key contributions. First, we propose a new GSL framework, which includes three steps: GSL base (\ie{} the representation used for GSL) construction, new structure construction, and view fusion, to better understand the effectiveness of GSL in GNNs. Second,  after graph convolution, we analyze the differences in mutual information (MI) between node representations derived from the original topology and those from the newly constructed topology. Surprisingly, our empirical observations and theoretical analysis show that no matter which type of graph structure construction methods are used, after feeding the same GSL bases to the newly constructed graph, there is no MI gain compared to the original GSL bases. To fairly reassess the effectiveness of GSL, we conduct ablation experiments and find that it is the pretrained GSL bases that enhance GNN performance, and in most cases, GSL itself does not contribute to the improved performance. This finding encourages us to focus on exploring essential components, such as self-training and structural encoding, in GNN design rather than only relying on GSL.
%by integrating GSL into GNN baselines and removing GSL from state-of-the-art models, while using the same GSL bases. The results indicate
% Based on our theoretical justifications and experimental results, we conclude that GSL bases cannot improve GNN performance on node classification tasks. 

% These insights contribute to a deeper understanding of GSL and prompt a re-evaluation of the essential components in the design of GNNs moving forward. 

% \sitao{"based on ...". State how novel your new framework is. What novel perspective do you use for the new categorization}
% \sitao{For all GSL methods or some kinds of GSL?}
% \sitao{Conduct ablation study?}
% \sitao{Make thing more specific. For example, "we point out the problems of the existing xxx based GSL methods with theoretical justifications and re-evaluated empirical results." You need to point out something useful, for example, provide a new benchmark, build a new model... }

\end{abstract}

\section{Introduction}

%  GNN GSL

% Controversy

% Our Analysis
% MI no gain. GNN fail because MI(AGG,Y) is ineffective.
% That explains why some ego node spliting methods: HES, CooGNN, GraphSage, GraphSIN work.

% Contribution... MI, Framework, Experiments

Graph Neural Networks (GNNs) \citep{GCN} are effective in capturing structural information from non-Euclidean data, which can be used in many applications such as recommendation \citep{gnn_app_recom1,gnn_app_recom2}, telecommunication \citep{lu2024gcepnet}, bio-informatics \citep{gnn_app_bio1,gnn_app_bio2,hua2024mudiff}, and social networks \citep{gnn_app_social1,luan2019break}. However, conventional GNNs suffer from issues including heterophily \citep{WRGAT, GGCN, lu2024flexible, luan2024heterophilic}, over-squashing \citep{brodyattentive}, adversarial attacks \citep{ProGNN,CoGSL,STABLE}, and missing or noisy structures \citep{WSGNN,NodeFormer,SUBLIME}. To address these issues, Graph Structure Learning (GSL) has been widely used \citep{gsl_survey}, which reconstructs or refines the original graph structures to enhance the performance of GNNs. However, GSL brings more hyperparameters and adds plenty of computational cost in both the construction process and the learning process. In addition, recent studies \citep{gnn_strong_baseline,hom_gnn_progress} have shown that GSL methods cannot consistently outperform traditional GNNs with the same hyperparameter tuning strategy. Therefore, an in-depth analysis of the effectiveness and necessity of GSL is highly needed.

%To better understand GSL, we propose a new framework of GSL framework. 
%In previous GSL survey papers \citep{gsl_survey, gsl_survey2, gsl_survey_engineer, gsl_survey_2018}, people mainly focus on summarizing graph structure construction methods, which are only involved in one step in GNNs learning. 
First, to better understand GSL, we propose a comprehensive framework to carefully break down GSL into $3$ steps: \textbf{GSL bases generation, new structure construction, and view fusion}. More specifically, \textbf{(1)} GSL bases are the processed node embeddings used before constructing new structures, which are constructed by either graph-aware or graph-agnostic models with fixed or learnable parameters. \textbf{(2)} Based on the GSL bases, new structures are constructed with similarity-based \citep{Geom-GCN, GLCN}, structural-based \citep{GAug, CoGSL}, or optimization-based approaches \citep{ProGNN}. Then, graph refinements are followed. \textbf{(3)} To preserve original graphs or combine multiples GSL graphs, several view fusion strategies are applied, \eg{} late fusion \citep{GEN}, early fusion \citep{STABLE}, or separation \citep{SUBLIME}. Compared with the existing categorization of GSL \citep{gsl_survey, gsl_survey2, gsl_survey_engineer, gsl_survey_2018} that mainly focuses on step \textbf{(2)}, our proposed framework carefully disentangles each component in GSL, which enhances our understanding of GSL in GNNs.

\begin{figure}[t]
    \centering
    \includegraphics[width=0.85\linewidth]{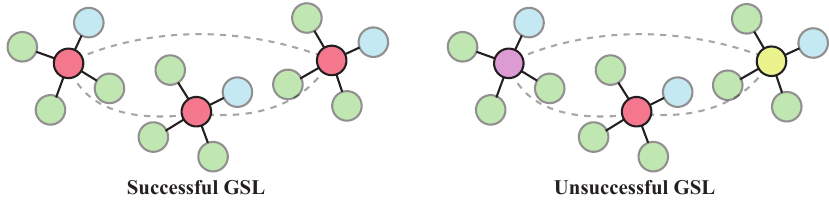}
    \caption{Examples of GSL that use neighbor distribution as GSL bases. The left shows a case of successful GSL, where new connections between \textcolor{red}{red nodes} are constructed using their GSL bases 3 \textcolor{green}{green nodes} and 1 \textcolor{cyan}{blue node}. The right shows a case of unsuccessful GSL, because the GSL bases of intra-class nodes are not consistent, and nodes with different classes are connected.}
    \label{fig:gsl_controversy}
\end{figure}

With the above framework, we explore the effectiveness of GSL in GNNs. As examples shown in Figure \ref{fig:gsl_controversy}, a GSL method creates new connections between nodes with similar GSL bases, which is denoted as the contextual information of the ego node and its neighbors in this case. When the GSL bases show high consistency with intra-class nodes, nodes within the same class are connected, which is beneficial for GNNs \citep{when_do_graph_help} and we denote it as \textbf{successful GSL}. Conversely, when the GSL bases show high consistency between inter-class nodes, nodes in different classes are likely to be connected, which is harmful to GNNs and we denote it as \textbf{unsuccessful GSL}. These examples highlight that the effectiveness of GSL is highly contingent on the quality of GSL bases. However, even if most GSL methods are supposed to be successful GSL, do we really need GSL in these cases? In this paper, our answer is \textbf{"No"}. The prerequisite of successful GSL is that the GSL bases are highly consistent within intra-class nodes, which inherently ensures a high quality of node representations \citep{gnn_neural_collapse}. Therefore, even a successful GSL is unnecessary because the GSL bases are already informative enough to provide distinguishable node embeddings.

% \sitao{What do you want to show from this example? Do you want to show that not all GSL is beneficial? Or GSL highly depends on the bases? Or something else?} 
% \sitao{This part need to be reconsidered very carefully}
% \sitao{Is the good base enough? Can GSL makes the good bases even better? You need to have very solid evidence before you claim GSL is unnecessary.}

Based on the above example, we theoretically analyze the effectiveness of GSL by mutual information between node representations and labels. Our results show that the mutual information between GSL bases and labels acts as an upper bound for the mutual information between newly aggregated node representations (derived from the new graph structures by GSL) and the labels, indicating that GSL is unnecessary in most cases. We also explain why existing GSL sometimes outperform GNNs in certain scenarios of heterophily \citep{Geom-GCN,luan2024heterophily} or inconsistent neighbor distributions \citep{disentangle_hom,is_hom_necessary}, which is missing in the previous works. Under these circumstances, GSL may outperform GNNs but still upper bound by Multilayer Perceptrons (MLP) which only depends on GSL bases. This also explains why ego node separation \citep{def_hom_edge} is an important part of model design. Our experiments show that, under the same hyperparameter tuning settings, no matter adding GSL to $4$ baseline GNNs or deleting GSL in $8$ state-of-the-art (SOTA) GSL-based methods, there are no significant changes in model performance on node classification. In conclusion, our main contributions are as follows.

% \sitao{Which is consistent with our theoretical analysis?}
% \sitao{Do you mean the new aggregated representations according to the new graph structure learn with GSL?}
% \sitao{node bases? If you use "node bases" for "node representations" in your paper, just keep it consistent.}
% \sitao{You need to explain what is GSL basis before.}
% \sitao{Do you mean GSL-augmented baseline GNNs outperform baseline GNNs?}

\begin{itemize}
    \item We propose a new GSL framework with three steps, which is more comprehensive than the previous literature and helps better understand each component in GSL.
    \item Both of our empirical experiments and theoretical analysis prove that there is no information gained by applying graph convolution on GSL bases under GSL graphs, indicating that most GSL methods are unnecessary.
    \item We fairly re-evaluate the effectiveness of GSL by adding GSL to GNN baselines and removing GSL in SOTA GSL-based methods models. The results show that GSL cannot consistently improve GNN performance.
\end{itemize}

% \sitao{Do you mean applying GSL constructed graph convolution on node representations}
% \sitao{What is non-optimization-based GSL and what is optimization-based GSL? When you introduce a new created word, make sure everybody understand it very clearly.}
% \sitao{Do you mean SOTA GSL-based methods?}

\section{Preliminary}

\paragraph{Graphs.} Suppose we have an undirected graph $\mathcal{G}=\{\mathcal{V},\mathcal{E}\}$ with node set $\mathcal{V}$ and edge set $\mathcal{E}$. Let $\b{Y} \in \mathbb{R}^{N \times C}$ denote the node labels and $\b{X} \in \mathbb{R}^{N \times M}$ represent the node features, where $N$ is the number of nodes, $C$ is the number of classes, and $M$ is the number of features. The graph structure is represented by an adjacency matrix $\b{A}$, where $\b{A}_{u,v} = \b{A}_{v,u} = 1$ indicates the existence of an edge $e_{uv}, e_{vu} \in \mathcal{E}$ between nodes $u$ and $v$. The normalized adjacency matrix is given by $\hat{\b{A}}=\b{\tilde{D}}^{-\frac{1}{2}}\b{\tilde{A}}\b{\tilde{D}}^{-\frac{1}{2}}$, where $\b{\tilde{D}} = \b{D}+\b{I_n}$ and $\b{\tilde{A}} = \b{A}+\b{I_n}$ represent the degree matrix and adjacency matrix with added self-loops. The neighbors of node $u$ is denoted as $\mathcal{N}_u = \{v|e_{uv}\in\mathcal{E}\}$. Graph Structure Learning (GSL) generates a new graph topology $\b{A}'$, where the new neighbors of node $u$ is denoted as $\mathcal{N}_u'$

\paragraph{Graph-aware models $\mathcal{M}^\mathcal{G}$ and graph-agnostic models $\mathcal{M}^{\neg\mathcal{G}}$.} Graph-aware models $\mathcal{M}^\mathcal{G}$, such as Graph Convolutional Networks (GCN) \citep{GCN}, Graph Attention Networks (GAT) \citep{GAT}, and ChebNet \citep{ChebNet}, utilize topological information in graphs through message aggregation or graph filters \citep{luan2022complete}, incorporating both $\b{X}$ and $\mathcal{G}$. In contrast, graph-agnostic models $\mathcal{M}^{\neg\mathcal{G}}$, such as Multilayer Perceptrons (MLP), only use $\b{X}$ without considering $\mathcal{G}$.

For example, the updating process of node embeddings in GCN and MLP can be represented as:
\begin{equation}
    \text{GCN}:\ \b{H^l}=\sigma(\hat{\b{A}}\b{H^{l-1}}\b{W^{l-1}}) ,\ \ 
    \text{MLP}:\ \b{H^l}=\sigma(\b{H^{l-1}}\b{W^{l-1}})
\end{equation}
where $\b{H^l}$ and $\b{W}^l$ are the node embeddings and weight matrix at the $l$-th layer, respectively, and $\sigma(\cdot)$ is an activation function.

\paragraph{Graph Homophily.} The concept of homophily originates from social network analysis and is defined as the tendency of individuals to connect with others who have similar characteristics \citep{hom_def_social}. A higher level of graph homophily makes the topological information of each node more informative, thereby improving the performance of graph-aware models $\mathcal{M}^\mathcal{G}$ \citep{when_do_graph_help, ACM_GCN, disentangle_hom}. Commonly used homophily metrics include edge homophily \citep{def_hom_edge,mixhop} and node homophily \citep{Geom-GCN}:

\begin{equation}
\begin{aligned}
\label{eq:definition_homophily_metrics}
& \resizebox{0.9\hsize}{!}{$h_{\text{edge}}(\mathcal{G},\b{Y}) = \frac{\big|\{e_{uv} \mid e_{uv}\in \mathcal{E}, Y_{u}=Y_{v}\}\big|}{|\mathcal{E}|}, \ 
h_{\text{node}}(\mathcal{G},\b{Y}) = \frac{1}{|\mathcal{V}|} \sum_{v \in \mathcal{V}}\frac{\big|\{u \mid u \in \mathcal{N}_v, Y_{u}=Y_{v}\}\big|}{\big|\mathcal{N}_v\bigl|}$} 
\end{aligned}
\end{equation}

\paragraph{Contextual Stochastic Block Models with Homophily (CSBM-H).} To study the behavior of GNNs, CSBM-H \citep{when_do_graph_help,is_hom_necessary} have been proposed to create synthetic graphs with a controlled homophily degree. Specifically, in CSBM-H, for a node $u$ with label $y$, its features $\b{X_u}\in\mathbb{R}^{M}$ are sampled from a class-wised Gaussian distribution $\b{X}_u \sim \b{N}_{Y_u}(\b{\mu}_{Y_u},\b{\Sigma}_{Y_u})$ with $\b{\mu}_{Y_u}\in\mathbb{R}^F$ and $\b{\Sigma}_{Y_u}\in\mathbb{R}^{F\times F}$, where each dimension of $\b{X}_u$ is independent from each other, \ie{$\b{\Sigma}_{Y_u}=\text{diag}(\mathbb{R}^n_{\ge 0})$}. Then, to generate graph structure $\mathcal{G}$ with given homophily degree $h$ with the range of $[0,1]$, the node $u$ has the probability $h$ to connect intra-class nodes and the probability $\frac{1-h}{C-1}$ to connect inter-class nodes. After applying neighbor sampling, both of the node homophily $h_{node}$ and edge homophily $h_{edge}$ in $\mathcal{G}$ are approximately equal to $h$.

\paragraph{Mutual Information.} Mutual Information quantifies the amount of information obtained about one random variable given another variable \citep{mutual_info}. The mutual information between variable $X$ and $Y$ can be expressed as:
\begin{equation}\label{eq:mi}
    I(\b{X};\b{Y}) = \sum_{y\in\mathcal{Y}}\sum_{x\in\mathcal{X}} p(x,y)\ \log{\frac{p(x,y)}{p(x)p(y)}}
\end{equation}
where $p(x,y)$ is joint probability, and $p(x)$ and $p(y)$ are marginal probability. 

Mutual information could be used to analyze the quality of input features by measuring how much information the inputs $\b{X}$ retain about the outputs $\b{Y}$. However, in graphs under the task of node classification, the mutual information between a discrete variable $\b{Y}$ and a continuous variable $\b{X}$ cannot be directly measured by Eq. (\ref{eq:mi}). Therefore, in this paper, we measure the mutual information $I(\b{X};\b{Y})$ based on entropy estimation from k-nearest neighbors distances following \citep{est_mutual_info,est_mutual_info2,est_mutual_info_ori}.

\section{Graph Structure Learning}

\begin{figure}[t]
    \centering
    \includegraphics[width=1.0\linewidth]{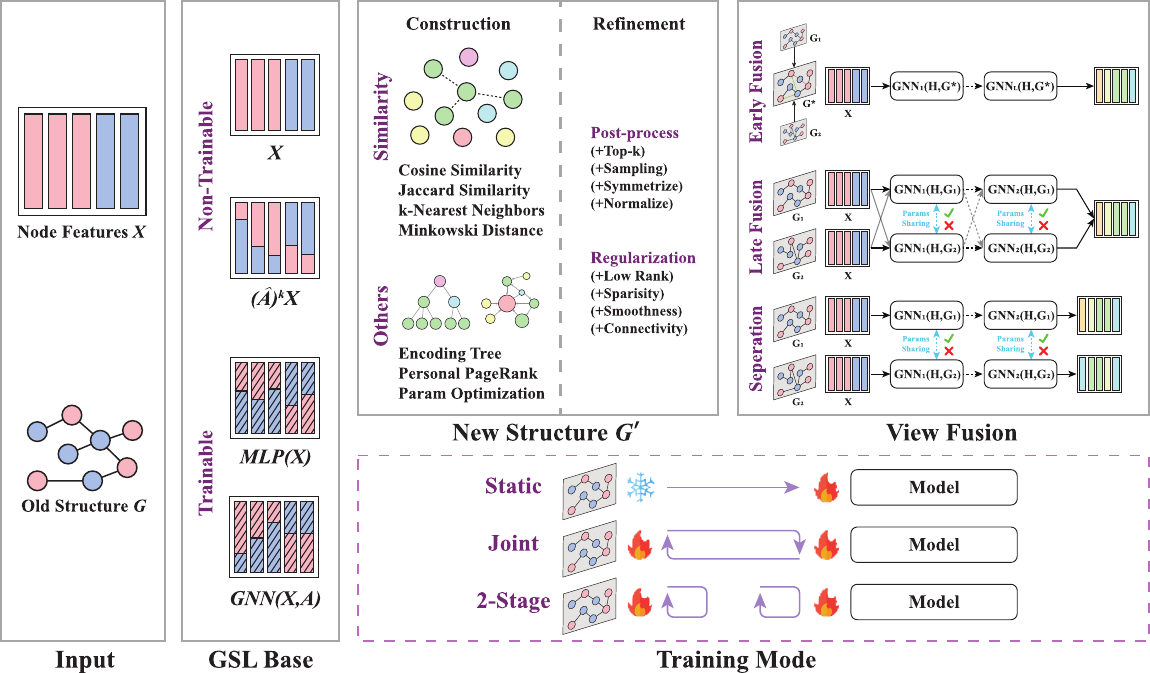}
    \caption{A framework of GSL.}
    \label{fig:taxomony}
\end{figure}

This section introduces our proposed Graph Structure Learning (GSL) framework. Previous surveys \citep{gsl_survey,gsl_survey2,gsl_survey_2018} only focus on new structure construction, constituting one step in GNNs learning. To provide a comprehensive understanding of GSL with GNNs, as shown in Figure \ref{fig:taxomony}, our framework includes three steps: GSL bases generation, new structure construction, and view fusion. Then we describe the pipeline of the GSL framework: First, GSL bases $\b{B}$ is constructed based on node features $\b{X}$ (and input graphs $\mathcal{G}$). Then, new graph structures $\mathcal{G}'$ are constructed with the GSL bases. Last, the information from new graph $\mathcal{G}'$ (multiple views if possible) and original graph $\mathcal{G}$ are combined with different view fusion strategies for the training of GNNs. Please refer to Appendix \ref{apd:gsl_method_explanation} for a more detailed explanation of the representative GSL methods within our proposed GSL framework.

\subsection{GSL Bases}\label{sec:GSL_basis}
The GSL bases $\b{B}$ is defined as the processed node embeddings used before constructing new structures. The quality of the GSL bases plays a crucial role in determining the performance of GNNs with GSL. For node classification tasks, an effective GSL bases $\b{B}$ should exhibit consistency among intra-class nodes, as shown in the left part of Figure \ref{fig:gsl_controversy}, is expected to be consistent among intra-class nodes. From the embedding training perspective, the construction of $\b{B}$ can be categorized as either non-parametric approaches \citep{LDS,Geom-GCN,SEGSL}, which generate static $\b{B}$, or parametric approaches \citep{ProGNN,IDGL,GRCN}, which updates $\b{B}$ dynamically during training. From the perspective of information usage, the construction of $\b{B}$ can be categorized into graph-agnostic approaches \citep{LDS,ProGNN,SEGSL} or graph-aware approaches \citep{Geom-GCN,GRCN,GEN}. Combining these two perspectives, in Figure \ref{fig:taxomony}, we show the diagrams of four types of of $\b{B}$ construction: $\b{B}=\b{X}$, $\b{B}=(\b{A})^k\b{X}$, $\b{B}=\text{MLP}(\b{X})$, and $\b{B}=\text{GNN}(\b{X},\b{A})$.

\subsection{New Structure Construction}
The construction of the new structure $\mathcal{G}'$, based on $\b{B}$, is a key element of GSL. From the perspective of relation extraction, methods for constructing $\mathcal{G}'$ can be categorized into similarity-based \citep{Geom-GCN, GLCN, ASC}, structure-based \citep{GAug, CoGSL, SEGSL}, and parametric optimization-based \citep{ProGNN, SUBLIME, GloGNN} approaches. Similarity-based methods are the most prevalent, and the choice of similarity measurement, such as k-Nearest Neighbors \citep{LDS}, cosine similarity \citep{IDGL}, or Minkowski distance \citep{SUBLIME}, plays a critical role in the quality of the reconstructed graphs. However, the initial $\mathcal{G}'$ produced by these methods often results in a coarse graph structure, which may not be optimal for GNN training. Thus, further refinements are often necessary, such as sampling \citep{GAug, STABLE, SUBLIME}, symmetrization \citep{GRCN, SLAPS, SUBLIME}, normalization \citep{GAug, SUBLIME, GLCN}, or applying graph regularization \citep{ProGNN, GLCN, GloGNN}.

\subsection{View Fusion}
In cases where the methods \citep{SLAPS, SEGSL, GLCN} already implicitly fuse the information from $\mathcal{G}$ into $\mathcal{G}'$, further view fusion is unnecessary. However, for other approaches, the fusion of information from the original graph structure $\mathcal{G}$ and the reconstructed structure $\mathcal{G}'$ is crucial. Based on the fusion stage, methods can be classified as early fusion \citep{STABLE, WSGNN, CoGSL}, late fusion \citep{GEN, SUBLIME, HiGNN}, or separation \citep{SUBLIME}. Early fusion, often seen as "graph editing", modifies $\mathcal{G}$ by adding or removing edges with $\mathcal{G}'$ before training. Late fusion keeps both views as input, fusing node embeddings either at each layer or in the final layer. Separation methods, typically paired with contrastive learning, maintain multiple views without embedding fusion during GNN training. Additionally, view fusion methods can be further distinguished by whether they involve parameter sharing across layers during training.

\subsection{Training Mode}
In addition to the previous three steps, the training mode of $\mathcal{G}'$ plays a crucial role in GSL and can be categorized into static, joint, and 2-stage approaches. Most methods \citep{ProGNN, GloGNN, GGCN} use joint training where $\mathcal{G}'$ and model parameters are optimized simultaneously. In contrast, some methods \citep{GEN, CoGSL, LDS} follow a 2-stage mode, iteratively updating $\mathcal{G}'$ and model parameters. While dynamic updates offer greater flexibility for learning complex structures through parameter optimization, they also significantly increase computational complexity, especially during the bases and graph construction steps. To address this, other methods \citep{HiGNN, WRGAT, ASC} opt for a static $\mathcal{G}'$ during training. Although this fixed structure may limit performance, it avoids the time-consuming process of frequent graph updates.

\section{Effectiveness of Graph Structure Learning}

% Emperical Experiments & Observation
% Theoretical results & Proposition
% Link with MLPvsGNNs, GNNs design

This section analyzes the impact of GSL on GNN performance with empirical observations in Section \ref{sec:ana_exp} and theoretical analysis in Section \ref{sec:ana_theory}. Then, the time complexity of GSL is analyzed in Section \ref{sec:ana_complexity}.

\subsection{Empirical Observations}\label{sec:ana_exp}

\begin{figure}
    \centering
  \subfloat[$\b{B}=\b{X}$\label{fig:raw_mi_acc}]{%
       \includegraphics[width=0.48\textwidth]{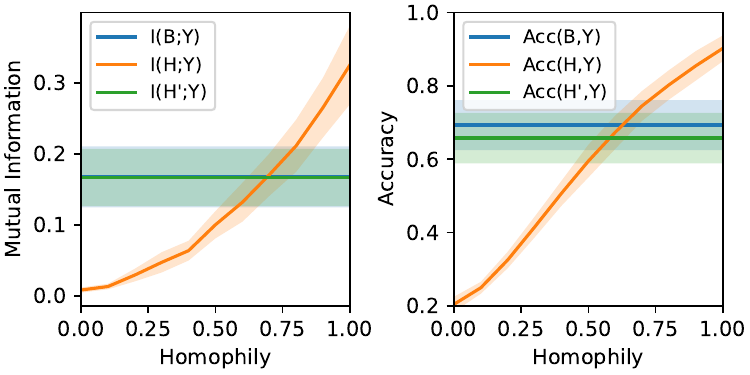}}
    \hfill
  \subfloat[$\b{B}=\b{\hat{A}}\b{X}$\label{fig:agg_mi_acc}]{%
        \includegraphics[width=0.48\textwidth]{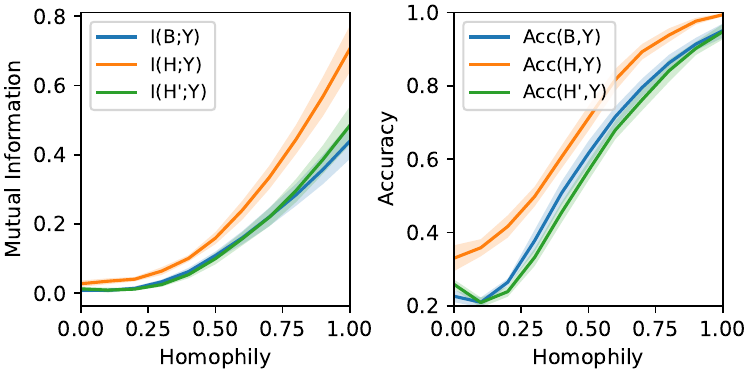}}
    \hfill
    \caption{Mutual information and accuracy of node classification on GSL bases \textcolor{cyan}{$\b{B}$}, convoluted bases of old graphs \textcolor{orange}{$\b{H}=\b{\hat{A}}\b{B}$}, convoluted bases of new graphs \textcolor{green}{$\b{H'}=\b{\hat{A}'}\b{B}$}, across varying homophily degrees. The rewriting bases $\b{B}$ is set to node features $\b{B}=\b{X}$ (left) or aggregated features $\b{B}=\b{\hat{A}}\b{X}$ (right).}
\end{figure}

\paragraph{Setting} Based on CSBM-H, we generate synthetic graphs with $10$ random seeds for each homophily degree $h\in\{0,0.1,\dots,1.0\}$ to mitigate randomness effects. Each graph $\mathcal{G}$ contains $1000$ nodes, with each node characterized by $10$ features, $5$ balanced classes, and a degree sampled from the range $[2, 10]$. Then, for the GSL, we apply k-Nearest-Neighbors algorithm on GSL bases $\b{B}$ with $k=5$ to generate new graphs, \ie{} $\mathcal{G}'=\text{kNN}(\b{B})$. To inspect the effectiveness of GSL, based on $\b{B}$, $\mathcal{G}$, and $\mathcal{G}'$, we can get several forms of node representation: $\b{B}$, $\b{H}=\text{GCN}(\b{B},\mathcal{G})$, and $\b{H}'=\text{GCN}(\b{B},\mathcal{G}')$, corresponding to the node representation of MLP, GCN, GCN+GSL respectively. We assess the performance of these methods on node classification tasks using mutual information, $\text{MI}(\cdot)$, as a non-parametric measure, and accuracy, $\text{Acc}(\cdot)$, as a parametric measure.

To consider both the graph-agnostic bases and graph-aware bases as discussed in Section \ref{sec:GSL_basis}, the GSL bases is selected as $\b{B}=\b{X}$ and $\b{B}=\b{\hat{A}}\b{X}$ as shown in Figure \ref{fig:raw_mi_acc} and Figure \ref{fig:agg_mi_acc} respectively. For example, on the left part of Figure \ref{fig:raw_mi_acc}, the performance of \textcolor{cyan}{MLP}, \textcolor{orange}{GCN}, \textcolor{green}{GCN+GSL} is shown as mutual information \textcolor{cyan}{$I(\b{B};\b{Y})$}, \textcolor{orange}{$I(\b{H};\b{Y})$}, and \textcolor{green}{$I(\b{H}';\b{Y})$} respectively. Based on these results, we make several observations as follows:

\textbf{Observation 1. Mutual information is an effective non-parametric measure of model performance.} As shown in Figure \ref{fig:raw_mi_acc} and Figure \ref{fig:agg_mi_acc}, the trend of mutual information $I(\cdot)$ (left) closely mirrors the model accuracy $\text{ACC}(\cdot)$ (right). Additionally, mutual information effectively distinguishes performance differences between methods. Since mutual information is non-parametric, it offers a flexible and reliable measure, making it suitable for theoretical analysis in the next section.

\textbf{Observation 2. MLP performs comparably to GCN+GSL under the same GSL bases.} In Figure \ref{fig:raw_mi_acc}, the mutual information \textcolor{cyan}{$I(\b{B};\b{Y})$} and classification accuracy \textcolor{cyan}{$\text{ACC}(\b{B}, \b{Y})$} are close to \textcolor{green}{$I(\b{H'};\b{Y})$} and \textcolor{green}{$\text{ACC}(\b{H'}, \b{Y})$}, respectively, across both graph-agnostic and graph-aware GSL bases. This suggests that, contrary to the expectation that GSL might enhance performance, the results indicate that the model performance does not improve significantly after applying graph convolution on $\mathcal{G}'$, reinforcing the GSL controversy discussed in Figure \ref{fig:gsl_controversy}.

\textbf{Observation 3. GCN+GSL sometimes outperforms GCN in heterophilous graphs.} As shown in Figure \ref{fig:raw_mi_acc}, the \textcolor{orange}{$I(\b{B};\b{Y})$} or \textcolor{orange}{$\text{ACC}(\b{B},\b{Y})$} increases with a higher homophily degree, while \textcolor{green}{$I(\b{H}';\b{Y})$} or \textcolor{green}{$\text{ACC}(\b{H}',\b{Y})$} remain stable across homophily degrees. In graphs with low homophily, the neighbors identified by GSL are more likely to share the same labels as the target node compared to the original graph neighbors, which causes GCN to underperform relative to GCN+GSL. However, this effect is observed only when $\b{B} = \b{X}$ (\ref{fig:raw_mi_acc}). When $\b{B} = \b{\hat{A}}\b{X}$ (\ref{fig:agg_mi_acc}), the GSL bases does not exhibit consistency among intra-class nodes in low homophily settings, leading GCN+GSL to perform worse than GCN.

These observations highlight that even when GCN+GSL outperforms GCN, its performance remains close to MLP under the same GSL bases. Recent studies \citep{gnn_strong_baseline, hom_gnn_progress} also indicate that under consistent hyperparameter tuning, GSL does not always consistently outperform classic GNN baselines. This leads us to reconsider the necessity of GSL. Thus, in addition to the empirical observations above, we proceed with a theoretical analysis of GSL's effectiveness in the following section.

\subsection{Theoretical Analysis}\label{sec:ana_theory}

To explain the above empirical observations, in this section, we first prove that the mutual information $I(Y;H)$ of label $Y$ and aggregated features $H$ serve as a non-parametric measurement of the performance of graph convolution. Following this, we compare the mutual information between the node labels $\b{Y}$ and either the original GSL bases $\b{B}$ or the aggregated GSL bases $\b{H}'$ (on GSL graph $\mathcal{G}'$), to highlight the impact of GSL on model performance.

\begin{theorem} 1
    Given a graph $\mathcal{G}=\{\mathcal{V},\mathcal{E}\}$ with node labels $\b{Y}$ and node features $\b{X}$, the accuracy of graph convolution on node classification is upper bounded by the mutual information of node label $Y$ and aggregated node features $H$:
    \begin{equation}
        P_A \le \frac{I(Y;H) + \log 2}{\log(C)}
    \end{equation}
\end{theorem}

\begin{theorem} 2
    Given a reconstructed graph $\mathcal{G}'=\{\mathcal{V},\mathcal{E}'\}$ on a bases $\b{B}$, the mutual information $I(Y;B')$ of node label $Y$ and aggregated bases $B'_u = \frac{1}{\abs{\mathcal{N}_u}}\sum_{v\in\mathcal{N}_u}B_v$ is upper bounded by $I(Y;B)$.
\end{theorem}
where the proofs are shown in Appendix \ref{apd:proof}. 

Theorem 1 shows that the mutual information $I(Y;H)$ provides an upper bound on the accuracy of graph convolution for node classification, which justifies why mutual information serves as an effective measure of model performance, as demonstrated in Observation 1.

Based on the conclusion of mutual information in Theorem 1, we analyze the effectiveness of GSL. Theorem 2 shows that the graph convolution on new graphs generated by GSL does not increase mutual information. This explains why MLP performs similarly to, or sightly better than, GCN+GSL in Observation 2 and the GSL controversy in Figure \ref{fig:gsl_controversy}, because GSL doesn't introduce additional information compared to the original GSL bases. \footnote{Admittedly, this theoretical analysis cannot be extended to optimization-based GSL due to the complexity of non-linear optimization problems. As such, the unnecessity of GSL in these methods is confirmed through our experiments.}

To further explain Observation 3 in Section \ref{sec:ana_exp}, we refer again to Theorem 2. In conjunction with previous studies on graph homophily \citep{Geom-GCN,ACM_GCN,disentangle_hom}, we know that the performance of GCN could be inferior to MLP on heterophilous graphs. Since GCN+GSL is upper bounded by the MLP on the same GSL bases, when MLP outperforms GCN, GCN+GSL may also outperform GCN, as seen in Figure \ref{fig:raw_mi_acc}. However, even when GCN+GSL surpasses GCN in some cases, it still lags behind MLP, a much simpler model, on the same GSL bases. Therefore, we hypothesize that previous GSL improvements stem from the construction of the GSL bases or the introduction of additional model parameters. A fair comparison of GSL with other GNNs or MLP baselines should be conducted using the same GSL bases, as demonstrated in our experiments.

\subsection{Complexity Analysis}\label{sec:ana_complexity}

After investigating the difference in the performance of GCN+GSL and GCN, we then analyze the time complexity of some representative methods of GSL, such as IDGL \citep{IDGL}, GRCN \citep{GRCN}, GAug \citep{GAug}, and HOG-GCN \citep{HOG-GCN}, as shown in Table \ref{tab:gsl_framework}. Assume the dimension of node representation is $F$ for all the layers, the additional time complexity introduced by GSL generally includes: 1. Construction of GSL bases: $O(\abs{\mathcal{E}}F+\abs{\mathcal{V}}F^2)$ for graph-aware bases or $O(\abs{\mathcal{V}}F^2)$ for graph-agnostic bases, 2. Graph construction: $O(\abs{\mathcal{V}}^2F)$, 3. Graph refinement: $O(\abs{\mathcal{V}}^2)$, and 4: View Fusion $O(\abs{\mathcal{V}}^2)$. Apart from the complexity of the new graph construction in GSL, during the graph convolution, compared with GNNs without using GSL, the additional complexity is further introduced by single view GSL $O(\abs{\mathcal{E}'}F)$ or multiple view GSL $O((N_\mathcal{G}-1)(\abs{\mathcal{E}}F+\abs{\mathcal{V}}F^2))$, where $\abs{\mathcal{E}'}$ is the additional edges introduced in GSL and $N_\mathcal{G}$ is the number of views in GSL. Consider the fact that $\abs{\mathcal{V}}^2 \gg \abs{\mathcal{E}}$, we have the total additional complexity of GSL by summing up all these terms: $O(\abs{\mathcal{V}}^2F+\abs{\mathcal{V}}F^2)$. Compared with the complexity in normal GCN $O(\abs{\mathcal{E}}F+\abs{\mathcal{V}}F^2)$ \citep{complexity_gcn}, this additional complexity $O((\abs{\mathcal{V}}^2-\abs{\mathcal{E}})F)$ adds tremendous training time and grows exponentially with the number of nodes in graphs, which is shown in our experiments.

\vspace{-0.3cm}
\section{Experiments}
In this section, we evaluate the effectiveness of GSL by comparing the performance of baseline GNNs and GNNs augmented with GSL (GNN+GSL), as well as the performance of GSL-based state-of-the-art (SOTA) methods against their non-GSL counterparts (SOTA vs. SOTA-GSL) on node classification tasks. The results of these comparisons are presented in Section \ref{sec:exp_performance}. Additionally, we analyze the quality of the newly constructed graphs generated by GSL in Section \ref{sec:exp_gsl_quality} and investigate how different components of GSL impact model performance in Section \ref{sec:exp_gsl_components}.

\textbf{Settings.} Our experiments include several baseline GNNs: GCN \citep{GCN}, SGC \citep{SGC}, GraphSage \citep{GraphSage}, and GAT \citep{GAT}, and GSL-based SOTA models: GAug \citep{GAug}, GEN \citep{GEN}, GRCN \citep{GRCN}, IDGL \citep{IDGL}, NodeFormer \citep{NodeFormer}, GloGNN \citep{GloGNN}, WRGAT \citep{WRGAT}, and WRGCN \citep{WRGAT}. The datasets used in our experiments include heterophilous graphs: Squirrel, Chameleon, Actor, Texas, Cornell, and Wisconsin \citep{Geom-GCN, dataset_hetero}, homophilous graphs: Cora, PubMed, and Citeseer \citep{dataset_cora}, and Minesweeper, Roman-empire, Amazon-ratings, Tolokers, and Questions \citep{hom_gnn_progress}. We show more dataset details in Appendix \ref{apd:dataset_details}. The model performance is measured by accuracy for multi-class datasets or AUC-ROC for binary-class datasets on node classification tasks. For the data splits, we use $50\%/25\%/25\%$ in train/validation/test sets for GNN+GSL and follow the default splits in OpenGSL \citep{OpenGSL} for each SOTA or SOTA-GSL method. Please refer to Appendix \ref{apd:implement_detail} for more implementation details.

\vspace{-0.3cm}
\subsection{Performance Comparison}\label{sec:exp_performance}

\textbf{GNN+GSL.} We investigate the impact of GSL by comparing the performance of GNN and GNN+GSL. As GSL introduces significant variations in three key aspects, outlined in Table \ref{tab:gsl_framework}, we aim to comprehensively evaluate all possible GSL configurations through a combination of various GSL components, which include 1) GSL bases: original features $\b{X}$, aggregated features $\b{\hat{A}X}$, MLP-pretrained features $\text{MLP}(\b{X})$, GCN-pretrained features $\text{GCN}(\b{X},\b{A})$, GCL(Graph Contrastive Learning)-pretrained features $\text{GCL}(\b{X},\b{A})$; 2) GSL Graph Construction: Graphs constructed via cosine similarity at the graph level (cos-graph), node level (cos-node), and k-nearest neighbors (kNN); and 3) View Fusion: early fusion $\{\mathcal{G}'\}$, late fusion $\{\mathcal{G},\mathcal{G}'\}$ with parameter sharing $\theta_1 = \theta_2$ or not $\theta_1 \neq \theta_2$. To ensure a fair comparison of the performance between GNN+GSL, GNN, and MLP, we consider $5$ GSL bases as input choices and train all models on each GSL bases. Detailed explanations of these GSL modules can be found in Appendix \ref{apd:gnn_plus_gsl}.

\begin{table}[h]
  \centering
\resizebox{1\hsize}{!}{
\begin{tabular}{*{19}{c}}\toprule
Model & Construct & Fusion & Param Sharing &Mines. & Roman. & Amazon. & Tolokers & Questions & Squirrel & Chameleon & Actor & Texas & Cornell & Wisconsin & Cora & CiteSeer & PubMed & Rank\\ \toprule 
MLP & None & - & - & 79.55$\pm$1.23 & 65.45$\pm$0.99 & 46.65$\pm$0.83 & 75.94$\pm$1.38 & 74.92$\pm$1.39 & \textcolor{blue}{39.29$\pm$2.22} & \textcolor{red}{43.57$\pm$4.18} & \textcolor{red}{35.40$\pm$1.38} & \textcolor{red}{80.46$\pm$6.44} & \textcolor{red}{73.78$\pm$7.34} & \textcolor{red}{85.88$\pm$7.78} & \textcolor{blue}{87.97$\pm$1.80} & \textcolor{blue}{76.68$\pm$2.10} & 87.39$\pm$2.18 & \textcolor{blue}{3.93}\\ 
GCN & None & - & - & \textcolor{red}{90.07$\pm$5.79} & \textcolor{red}{81.46$\pm$1.25} & \textcolor{red}{50.89$\pm$1.16} & \textcolor{red}{84.61$\pm$0.99} & \textcolor{red}{77.68$\pm$1.10} & \textcolor{red}{41.26$\pm$2.47} & \textcolor{blue}{43.24$\pm$3.86} & \textcolor{blue}{34.34$\pm$1.17} & \textcolor{blue}{73.08$\pm$8.68} & \textcolor{blue}{67.03$\pm$10.54} & \textcolor{blue}{78.24$\pm$8.32} & \textcolor{red}{87.97$\pm$1.51} & \textcolor{red}{76.75$\pm$2.30} & \textcolor{red}{89.47$\pm$0.64} & \textcolor{red}{1.36}\\ 
GCN & cos-graph & $\{\mathcal{G}'\}$ & - & 77.91$\pm$5.25 & 67.40$\pm$1.02 & 46.72$\pm$1.51 & 76.11$\pm$1.52 & 72.56$\pm$1.14 & 38.15$\pm$2.45 & 39.87$\pm$4.87 & 33.47$\pm$1.61 & 63.06$\pm$9.85 & 65.68$\pm$7.76 & 72.75$\pm$5.70 & 85.21$\pm$1.39 & 75.52$\pm$1.14 & \textcolor{blue}{89.03$\pm$0.42} & 6.71\\ 
GCN & cos-graph & $\{\mathcal{G},\mathcal{G}'\}$ & $\theta_1=\theta_2$ & 52.53$\pm$6.45 & 62.57$\pm$0.81 & 41.29$\pm$1.61 & 74.22$\pm$1.79 & 69.63$\pm$1.52 & 37.62$\pm$1.74 & 39.78$\pm$4.00 & 32.74$\pm$0.92 & 57.88$\pm$8.75 & 66.49$\pm$9.12 & 73.14$\pm$5.92 & 64.68$\pm$1.61 & 67.32$\pm$1.89 & 86.43$\pm$0.76 & 9.32\\ 
GCN & cos-graph & $\{\mathcal{G},\mathcal{G}'\}$ & $\theta_1\neq\theta_2$ & 88.70$\pm$0.86 & 69.90$\pm$2.38 & 47.35$\pm$0.83 & 82.85$\pm$0.95 & 75.29$\pm$1.38 & 38.84$\pm$2.87 & 40.30$\pm$4.31 & 33.73$\pm$1.49 & 65.47$\pm$8.48 & 62.97$\pm$10.89 & 75.29$\pm$6.54 & 85.51$\pm$1.87 & 75.23$\pm$1.14 & 88.74$\pm$0.59 & 4.79\\ 
GCN & cos-node & $\{\mathcal{G}'\}$ & - & 85.57$\pm$6.63 & 68.24$\pm$2.49 & 47.56$\pm$1.32 & 77.26$\pm$1.44 & 74.16$\pm$1.80 & 38.14$\pm$2.40 & 40.16$\pm$3.13 & 34.04$\pm$1.66 & 61.13$\pm$8.19 & 61.08$\pm$8.16 & 71.18$\pm$6.98 & 86.06$\pm$1.95 & 75.76$\pm$1.39 & 88.92$\pm$0.50 & 5.93\\ 
GCN & cos-node & $\{\mathcal{G},\mathcal{G}'\}$ & $\theta_1=\theta_2$ & 52.53$\pm$6.45 & 62.57$\pm$0.81 & 41.29$\pm$1.61 & 74.22$\pm$1.79 & 69.63$\pm$1.52 & 37.62$\pm$1.74 & 39.78$\pm$4.00 & 32.74$\pm$0.92 & 57.88$\pm$8.75 & 66.49$\pm$9.12 & 73.14$\pm$5.92 & 64.68$\pm$1.61 & 67.32$\pm$1.89 & 86.43$\pm$0.76 & 9.36\\ 
GCN & cos-node & $\{\mathcal{G},\mathcal{G}'\}$ & $\theta_1\neq\theta_2$ & \textcolor{blue}{89.17$\pm$0.68} & \textcolor{blue}{72.63$\pm$1.45} & \textcolor{blue}{48.31$\pm$0.96} & 82.91$\pm$0.97 & 75.56$\pm$1.05 & 38.41$\pm$2.32 & 39.94$\pm$4.49 & 34.10$\pm$1.53 & 64.68$\pm$8.85 & 63.24$\pm$9.47 & 73.92$\pm$7.51 & 85.69$\pm$1.73 & 75.49$\pm$1.42 & 88.72$\pm$0.71 & 4.29\\ 
GCN & kNN & $\{\mathcal{G}'\}$ & - & 82.89$\pm$6.66 & 68.44$\pm$0.83 & 47.13$\pm$1.00 & 78.92$\pm$1.79 & 73.90$\pm$1.73 & 38.15$\pm$2.02 & 40.22$\pm$3.82 & 33.94$\pm$1.24 & 63.03$\pm$8.53 & 61.35$\pm$9.28 & 72.16$\pm$7.41 & 86.08$\pm$1.62 & 75.56$\pm$1.42 & 88.59$\pm$0.58 & 5.93\\ 
GCN & kNN & $\{\mathcal{G},\mathcal{G}'\}$ & $\theta_1=\theta_2$ & 52.53$\pm$6.45 & 62.57$\pm$0.81 & 41.29$\pm$1.61 & 74.22$\pm$1.79 & 69.63$\pm$1.52 & 37.62$\pm$1.74 & 39.78$\pm$4.00 & 32.74$\pm$0.92 & 57.88$\pm$8.75 & 66.49$\pm$9.12 & 73.14$\pm$5.92 & 64.68$\pm$1.61 & 67.32$\pm$1.89 & 86.43$\pm$0.76 & 9.39\\ 
GCN & kNN & $\{\mathcal{G},\mathcal{G}'\}$ & $\theta_1\neq\theta_2$ & 88.96$\pm$0.73 & 72.44$\pm$1.61 & 47.06$\pm$0.83 & \textcolor{blue}{83.10$\pm$0.80} & \textcolor{blue}{75.61$\pm$1.19} & 37.63$\pm$1.93 & 40.18$\pm$4.76 & 33.84$\pm$1.94 & 63.87$\pm$9.68 & 62.16$\pm$9.77 & 75.49$\pm$7.29 & 85.82$\pm$1.55 & 75.50$\pm$1.30 & 88.54$\pm$0.55 & 5.00\\ 
\toprule 

MLP & None & - & - & 79.55$\pm$1.23 & 65.45$\pm$0.99 & 46.65$\pm$0.83 & 75.94$\pm$1.38 & 74.92$\pm$1.39 & \textcolor{blue}{39.29$\pm$2.22} & \textcolor{red}{43.57$\pm$4.18} & \textcolor{red}{35.40$\pm$1.38} & \textcolor{red}{80.46$\pm$6.44} & \textcolor{red}{73.78$\pm$7.34} & \textcolor{red}{85.88$\pm$7.78} & \textcolor{blue}{87.97$\pm$1.80} & \textcolor{blue}{76.68$\pm$2.10} & 87.39$\pm$2.18 & \textcolor{blue}{3.71}\\ 
SGC & None & - & - & \textcolor{red}{83.45$\pm$4.47} & \textcolor{red}{78.04$\pm$0.69} & \textcolor{red}{51.38$\pm$0.68} & \textcolor{red}{84.88$\pm$1.13} & \textcolor{red}{77.39$\pm$1.23} & \textcolor{red}{41.18$\pm$2.73} & \textcolor{blue}{42.35$\pm$4.10} & 34.05$\pm$1.41 & \textcolor{blue}{73.63$\pm$6.94} & \textcolor{blue}{70.27$\pm$9.91} & \textcolor{blue}{80.59$\pm$5.13} & \textcolor{red}{88.10$\pm$1.89} & \textcolor{red}{77.52$\pm$2.20} & \textcolor{red}{89.39$\pm$0.62} & \textcolor{red}{1.57}\\ 
SGC & cos-graph & $\{\mathcal{G}'\}$ & - & 73.76$\pm$4.46 & 67.17$\pm$0.81 & 47.15$\pm$0.88 & 76.28$\pm$1.63 & 73.93$\pm$2.66 & 38.66$\pm$2.53 & 40.07$\pm$4.39 & 33.87$\pm$1.45 & 71.19$\pm$7.38 & 67.57$\pm$9.19 & 77.65$\pm$6.08 & 86.95$\pm$2.01 & 76.12$\pm$1.29 & 89.10$\pm$0.43 & 5.79\\ 
SGC & cos-graph & $\{\mathcal{G},\mathcal{G}'\}$ & $\theta_1=\theta_2$ & 52.53$\pm$4.89 & 62.97$\pm$0.78 & 42.42$\pm$1.57 & 74.29$\pm$1.79 & 70.56$\pm$1.27 & 37.56$\pm$2.25 & 39.33$\pm$3.60 & 32.85$\pm$0.90 & 57.60$\pm$7.53 & 66.49$\pm$10.37 & 71.57$\pm$4.46 & 64.82$\pm$2.11 & 67.55$\pm$1.80 & 86.58$\pm$0.72 & 9.64\\ 
SGC & cos-graph & $\{\mathcal{G},\mathcal{G}'\}$ & $\theta_1\neq\theta_2$ & 79.70$\pm$1.21 & 62.02$\pm$2.06 & 47.24$\pm$0.93 & 83.22$\pm$1.52 & \textcolor{blue}{77.19$\pm$0.99} & 38.32$\pm$1.80 & 40.85$\pm$4.61 & 33.51$\pm$1.50 & 70.34$\pm$7.31 & 64.86$\pm$9.01 & 75.29$\pm$6.82 & 87.47$\pm$1.70 & 75.70$\pm$1.28 & 88.65$\pm$0.49 & 6.14\\ 
SGC & cos-node & $\{\mathcal{G}'\}$ & - & 79.03$\pm$3.76 & 67.84$\pm$1.87 & 47.93$\pm$0.94 & 78.09$\pm$1.84 & 75.46$\pm$1.43 & 38.61$\pm$2.20 & 40.50$\pm$4.10 & 34.03$\pm$1.27 & 70.08$\pm$6.84 & 68.11$\pm$9.23 & 77.45$\pm$4.63 & 87.47$\pm$1.86 & 76.36$\pm$1.27 & \textcolor{blue}{89.37$\pm$0.41} & 4.54\\ 
SGC & cos-node & $\{\mathcal{G},\mathcal{G}'\}$ & $\theta_1=\theta_2$ & 52.53$\pm$4.89 & 62.97$\pm$0.78 & 42.42$\pm$1.57 & 74.29$\pm$1.79 & 70.56$\pm$1.27 & 37.56$\pm$2.25 & 39.33$\pm$3.60 & 32.85$\pm$0.90 & 57.60$\pm$7.53 & 66.49$\pm$10.37 & 71.57$\pm$4.46 & 64.82$\pm$2.11 & 67.55$\pm$1.80 & 86.58$\pm$0.72 & 9.57\\ 
SGC & cos-node & $\{\mathcal{G},\mathcal{G}'\}$ & $\theta_1\neq\theta_2$ & 80.12$\pm$1.36 & 66.90$\pm$1.66 & \textcolor{blue}{48.04$\pm$0.97} & \textcolor{blue}{83.53$\pm$1.43} & 77.11$\pm$1.09 & 38.52$\pm$2.29 & 40.20$\pm$4.66 & 34.20$\pm$1.79 & 68.47$\pm$8.11 & 64.59$\pm$9.74 & 75.29$\pm$6.05 & 87.54$\pm$1.63 & 75.88$\pm$1.26 & 88.68$\pm$0.43 & 5.11\\ 
SGC & kNN & $\{\mathcal{G}'\}$ & - & 75.53$\pm$4.98 & \textcolor{blue}{67.94$\pm$0.70} & 47.68$\pm$0.84 & 79.45$\pm$2.06 & 74.22$\pm$2.47 & 37.32$\pm$2.10 & 39.92$\pm$3.91 & 34.05$\pm$1.55 & 72.81$\pm$6.15 & 70.00$\pm$7.98 & 77.84$\pm$6.02 & 87.82$\pm$1.77 & 76.54$\pm$1.44 & 89.19$\pm$0.42 & 4.64\\ 
SGC & kNN & $\{\mathcal{G},\mathcal{G}'\}$ & $\theta_1=\theta_2$ & 52.53$\pm$4.89 & 62.97$\pm$0.78 & 42.42$\pm$1.57 & 74.29$\pm$1.79 & 70.56$\pm$1.27 & 37.56$\pm$2.25 & 39.33$\pm$3.60 & 32.85$\pm$0.90 & 57.60$\pm$7.53 & 66.49$\pm$10.37 & 71.57$\pm$4.46 & 64.82$\pm$2.11 & 67.55$\pm$1.80 & 86.58$\pm$0.72 & 9.50\\ 
SGC & kNN & $\{\mathcal{G},\mathcal{G}'\}$ & $\theta_1\neq\theta_2$ & \textcolor{blue}{80.78$\pm$1.08} & 64.59$\pm$1.93 & 47.48$\pm$0.99 & 83.17$\pm$1.43 & 76.80$\pm$1.09 & 36.53$\pm$2.06 & 40.17$\pm$4.24 & \textcolor{blue}{34.23$\pm$1.72} & 69.26$\pm$6.77 & 65.95$\pm$8.87 & 76.08$\pm$5.92 & 87.38$\pm$1.49 & 76.02$\pm$1.22 & 88.77$\pm$0.45 & 5.79\\ 
\toprule 

MLP & None & - & - & 79.55$\pm$1.23 & 65.45$\pm$0.99 & 46.65$\pm$0.83 & 75.94$\pm$1.38 & 74.92$\pm$1.39 & 39.29$\pm$2.22 & \textcolor{red}{43.57$\pm$4.18} & \textcolor{red}{35.40$\pm$1.38} & \textcolor{red}{80.46$\pm$6.44} & \textcolor{blue}{73.78$\pm$7.34} & \textcolor{blue}{85.88$\pm$7.78} & \textcolor{blue}{87.97$\pm$1.80} & \textcolor{red}{76.68$\pm$2.10} & 87.39$\pm$2.18 & 4.14\\ 
SAGE & None & - & - & \textcolor{blue}{90.66$\pm$0.88} & \textcolor{red}{85.02$\pm$0.97} & \textcolor{red}{52.93$\pm$0.83} & \textcolor{red}{83.31$\pm$1.12} & \textcolor{red}{75.95$\pm$1.41} & \textcolor{red}{40.43$\pm$2.64} & \textcolor{blue}{42.95$\pm$5.37} & 34.83$\pm$1.20 & \textcolor{blue}{80.17$\pm$6.90} & \textcolor{red}{75.68$\pm$7.52} & \textcolor{red}{86.27$\pm$6.67} & \textcolor{red}{88.13$\pm$1.77} & \textcolor{blue}{76.65$\pm$2.00} & \textcolor{red}{89.18$\pm$0.65} & \textcolor{red}{1.71}\\ 
SAGE & cos-graph & $\{\mathcal{G}'\}$ & - & 80.39$\pm$4.66 & 70.13$\pm$1.05 & 47.55$\pm$1.17 & 76.77$\pm$1.28 & 72.86$\pm$1.18 & 39.03$\pm$2.69 & 40.84$\pm$5.42 & 34.75$\pm$1.39 & 70.91$\pm$8.58 & 70.00$\pm$7.56 & 78.24$\pm$6.87 & 83.64$\pm$2.03 & 75.53$\pm$1.36 & \textcolor{blue}{89.18$\pm$0.35} & 6.07\\ 
SAGE & cos-graph & $\{\mathcal{G},\mathcal{G}'\}$ & $\theta_1=\theta_2$ & 53.02$\pm$6.49 & 59.98$\pm$1.73 & 39.99$\pm$2.29 & 71.57$\pm$2.28 & 66.01$\pm$3.58 & 35.05$\pm$2.41 & 38.49$\pm$3.68 & 31.32$\pm$1.04 & 60.30$\pm$7.05 & 67.57$\pm$4.59 & 76.47$\pm$5.92 & 64.58$\pm$1.74 & 67.77$\pm$1.31 & 85.53$\pm$0.51 & 9.93\\ 
SAGE & cos-graph & $\{\mathcal{G},\mathcal{G}'\}$ & $\theta_1\neq\theta_2$ & \textcolor{red}{90.67$\pm$0.66} & 79.02$\pm$1.21 & \textcolor{blue}{52.10$\pm$0.84} & \textcolor{blue}{82.17$\pm$0.89} & \textcolor{blue}{75.38$\pm$0.96} & \textcolor{blue}{39.36$\pm$2.14} & 40.64$\pm$6.06 & 35.14$\pm$1.08 & 76.08$\pm$6.30 & 70.27$\pm$6.62 & 79.41$\pm$5.71 & 83.60$\pm$1.78 & 74.39$\pm$1.35 & 88.88$\pm$0.50 & \textcolor{blue}{3.86}\\ 
SAGE & cos-node & $\{\mathcal{G}'\}$ & - & 85.26$\pm$4.64 & 71.25$\pm$1.76 & 48.96$\pm$0.87 & 78.39$\pm$1.75 & 73.01$\pm$1.11 & 38.68$\pm$2.75 & 40.81$\pm$4.51 & 35.10$\pm$1.26 & 71.47$\pm$9.47 & 68.11$\pm$7.87 & 75.49$\pm$6.32 & 84.88$\pm$1.90 & 75.58$\pm$1.04 & 89.17$\pm$0.35 & 5.64\\ 
SAGE & cos-node & $\{\mathcal{G},\mathcal{G}'\}$ & $\theta_1=\theta_2$ & 53.02$\pm$6.49 & 59.98$\pm$1.73 & 39.99$\pm$2.29 & 71.59$\pm$2.28 & 66.01$\pm$3.58 & 35.05$\pm$2.41 & 38.49$\pm$3.68 & 31.32$\pm$1.04 & 60.30$\pm$7.05 & 67.57$\pm$4.59 & 76.47$\pm$5.92 & 64.58$\pm$1.74 & 67.77$\pm$1.31 & 85.53$\pm$0.51 & 9.79\\ 
SAGE & cos-node & $\{\mathcal{G},\mathcal{G}'\}$ & $\theta_1\neq\theta_2$ & 90.64$\pm$0.65 & 78.60$\pm$0.98 & 52.08$\pm$0.90 & 82.02$\pm$0.88 & 75.31$\pm$1.12 & 39.18$\pm$2.54 & 40.86$\pm$6.17 & \textcolor{blue}{35.18$\pm$1.24} & 74.71$\pm$5.65 & 69.73$\pm$7.43 & 80.00$\pm$5.68 & 83.96$\pm$1.65 & 74.63$\pm$1.26 & 88.93$\pm$0.64 & 3.93\\ 
SAGE & kNN & $\{\mathcal{G}'\}$ & - & 82.86$\pm$3.14 & 70.74$\pm$0.80 & 48.40$\pm$1.01 & 78.12$\pm$2.17 & 72.70$\pm$1.15 & 38.93$\pm$2.84 & 39.68$\pm$5.40 & 35.09$\pm$1.14 & 70.91$\pm$9.05 & 68.92$\pm$6.88 & 75.69$\pm$6.73 & 84.40$\pm$1.75 & 75.68$\pm$1.43 & 88.86$\pm$0.44 & 6.50\\ 
SAGE & kNN & $\{\mathcal{G},\mathcal{G}'\}$ & $\theta_1=\theta_2$ & 53.02$\pm$6.49 & 59.98$\pm$1.73 & 39.99$\pm$2.29 & 71.59$\pm$2.28 & 66.01$\pm$3.58 & 35.05$\pm$2.41 & 38.49$\pm$3.68 & 31.32$\pm$1.04 & 60.30$\pm$7.05 & 67.57$\pm$4.59 & 76.47$\pm$5.92 & 64.58$\pm$1.74 & 67.77$\pm$1.31 & 85.53$\pm$0.51 & 9.86\\ 
SAGE & kNN & $\{\mathcal{G},\mathcal{G}'\}$ & $\theta_1\neq\theta_2$ & 90.61$\pm$0.63 & \textcolor{blue}{79.16$\pm$1.15} & 51.56$\pm$1.07 & 81.66$\pm$0.87 & 75.22$\pm$0.97 & 39.20$\pm$2.39 & 40.44$\pm$5.82 & 35.13$\pm$1.38 & 74.17$\pm$6.31 & 70.54$\pm$7.32 & 79.61$\pm$6.61 & 84.05$\pm$1.63 & 74.59$\pm$1.25 & 88.67$\pm$0.55 & 4.57\\ 
\toprule 

MLP & None & - & - & 79.55$\pm$1.23 & 65.45$\pm$0.99 & 46.65$\pm$0.83 & 75.94$\pm$1.38 & 74.92$\pm$1.39 & 39.29$\pm$2.22 & \textcolor{blue}{43.57$\pm$4.18} & \textcolor{red}{35.40$\pm$1.38} & \textcolor{red}{80.46$\pm$6.44} & \textcolor{red}{73.78$\pm$7.34} & \textcolor{red}{85.88$\pm$7.78} & \textcolor{blue}{87.97$\pm$1.80} & \textcolor{blue}{76.68$\pm$2.10} & 87.39$\pm$2.18 & \textcolor{blue}{3.86}\\ 
GAT & None & - & - & \textcolor{red}{90.41$\pm$1.34} & \textcolor{red}{84.51$\pm$0.84} & \textcolor{red}{52.00$\pm$2.84} & \textcolor{red}{84.37$\pm$0.96} & \textcolor{red}{77.78$\pm$1.27} & \textcolor{red}{41.67$\pm$2.51} & \textcolor{red}{43.83$\pm$3.66} & 33.73$\pm$1.77 & \textcolor{blue}{75.28$\pm$8.12} & 65.41$\pm$12.14 & 77.84$\pm$7.41 & \textcolor{red}{88.02$\pm$1.92} & \textcolor{red}{76.77$\pm$2.02} & \textcolor{red}{89.21$\pm$0.67} & \textcolor{red}{2.04}\\ 
GAT & cos-graph & $\{\mathcal{G}'\}$ & - & 80.78$\pm$8.24 & 67.68$\pm$1.25 & 45.79$\pm$1.10 & 74.84$\pm$1.84 & 72.34$\pm$1.49 & 38.74$\pm$2.54 & 40.21$\pm$3.53 & 33.37$\pm$1.10 & 62.73$\pm$9.06 & \textcolor{blue}{67.57$\pm$7.03} & 77.06$\pm$7.29 & 86.03$\pm$1.85 & 75.46$\pm$1.49 & \textcolor{blue}{88.63$\pm$0.59} & 6.29\\ 
GAT & cos-graph & $\{\mathcal{G},\mathcal{G}'\}$ & $\theta_1=\theta_2$ & 53.16$\pm$7.93 & 63.67$\pm$1.08 & 44.83$\pm$2.04 & 73.46$\pm$1.07 & 68.92$\pm$1.53 & 37.14$\pm$2.13 & 39.85$\pm$2.87 & 32.06$\pm$1.12 & 57.03$\pm$8.70 & 67.30$\pm$4.67 & 75.10$\pm$5.85 & 64.84$\pm$1.45 & 67.82$\pm$0.62 & 86.47$\pm$0.66 & 9.46\\ 
GAT & cos-graph & $\{\mathcal{G},\mathcal{G}'\}$ & $\theta_1\neq\theta_2$ & 89.97$\pm$0.80 & 76.08$\pm$1.70 & 49.61$\pm$0.73 & 82.75$\pm$0.90 & \textcolor{blue}{77.13$\pm$1.20} & 39.21$\pm$2.81 & 40.40$\pm$3.30 & 33.05$\pm$1.20 & 70.66$\pm$7.77 & 66.76$\pm$7.23 & \textcolor{blue}{78.82$\pm$6.76} & 86.60$\pm$1.75 & 75.05$\pm$1.36 & 87.85$\pm$0.72 & 4.71\\ 
GAT & cos-node & $\{\mathcal{G}'\}$ & - & 87.64$\pm$8.40 & 68.80$\pm$2.39 & 46.37$\pm$1.06 & 77.77$\pm$1.86 & 73.65$\pm$1.47 & 38.65$\pm$2.46 & 40.33$\pm$3.25 & 33.43$\pm$0.94 & 64.64$\pm$9.09 & 65.41$\pm$8.48 & 75.10$\pm$6.13 & 87.08$\pm$1.66 & 75.59$\pm$1.49 & 88.59$\pm$0.49 & 5.82\\ 
GAT & cos-node & $\{\mathcal{G},\mathcal{G}'\}$ & $\theta_1=\theta_2$ & 53.16$\pm$7.93 & 63.67$\pm$1.08 & 44.83$\pm$2.04 & 73.46$\pm$1.07 & 68.92$\pm$1.53 & 37.14$\pm$2.13 & 39.85$\pm$2.87 & 32.06$\pm$1.12 & 57.03$\pm$8.70 & 67.30$\pm$4.67 & 75.10$\pm$5.85 & 64.84$\pm$1.45 & 67.82$\pm$0.62 & 86.47$\pm$0.66 & 9.46\\ 
GAT & cos-node & $\{\mathcal{G},\mathcal{G}'\}$ & $\theta_1\neq\theta_2$ & \textcolor{blue}{90.03$\pm$0.78} & \textcolor{blue}{77.56$\pm$2.75} & \textcolor{blue}{50.36$\pm$0.70} & 82.72$\pm$1.16 & 76.83$\pm$1.16 & 38.97$\pm$3.12 & 40.56$\pm$3.77 & 33.49$\pm$1.35 & 70.39$\pm$7.34 & 65.95$\pm$6.77 & 78.63$\pm$6.59 & 86.64$\pm$1.78 & 75.32$\pm$1.04 & 87.87$\pm$0.61 & 4.21\\ 
GAT & kNN & $\{\mathcal{G}'\}$ & - & 84.27$\pm$5.25 & 68.73$\pm$1.47 & 46.05$\pm$0.90 & 77.57$\pm$1.75 & 71.58$\pm$1.62 & 38.82$\pm$2.33 & 40.12$\pm$3.69 & \textcolor{blue}{33.84$\pm$1.07} & 61.68$\pm$8.71 & 62.97$\pm$7.43 & 74.90$\pm$5.86 & 86.77$\pm$1.90 & 75.64$\pm$1.45 & 88.29$\pm$0.48 & 6.50\\ 
GAT & kNN & $\{\mathcal{G},\mathcal{G}'\}$ & $\theta_1=\theta_2$ & 53.16$\pm$7.93 & 63.67$\pm$1.08 & 44.83$\pm$2.04 & 73.46$\pm$1.07 & 68.92$\pm$1.53 & 37.14$\pm$2.13 & 39.85$\pm$2.87 & 32.06$\pm$1.12 & 57.03$\pm$8.70 & 67.30$\pm$4.67 & 75.10$\pm$5.85 & 64.84$\pm$1.45 & 67.82$\pm$0.62 & 86.47$\pm$0.66 & 9.46\\ 
GAT & kNN & $\{\mathcal{G},\mathcal{G}'\}$ & $\theta_1\neq\theta_2$ & 89.96$\pm$0.79 & 77.23$\pm$1.63 & 49.79$\pm$0.72 & \textcolor{blue}{82.78$\pm$0.95} & 76.67$\pm$1.13 & \textcolor{blue}{39.65$\pm$2.76} & 41.11$\pm$3.92 & 33.54$\pm$1.36 & 70.38$\pm$7.22 & 65.95$\pm$6.52 & 77.84$\pm$7.23 & 86.97$\pm$1.75 & 75.20$\pm$1.55 & 87.97$\pm$0.51 & 4.18\\ 
\bottomrule

\end{tabular}
}
\caption{Performance of GNNs with GNN+GSL.}\label{tab:baseline_plus_rewriting}
\end{table}%

Table \ref{tab:baseline_plus_rewriting} shows the performance of MLP, GNN baselines, and GNN+GSL across $8$ datasets, using the best-performing GSL bases. For each GNN backbone, the best-performing method is highlighted in \textcolor{red}{red}, while the second-best method is highlighted in \textcolor{blue}{blue}. Notably, under fair comparison conditions, all $4$ baseline GNNs outperform their GNN+GSL counterparts. This suggests that incorporating GSL into these GNN baselines does not consistently yield performance improvements and leads to worse results in many instances. However, these results alone are insufficient to conclusively determine the effectiveness of GSL, as the method may require specific training procedures or more complex model designs. Therefore, we further examine the performance of state-of-the-art (SOTA) GSL approaches to more fairly evaluate GSL's potential within GNNs.

\textbf{SOTA-GSL.} To fairly reassess the impact of GSL in state-of-the-art (SOTA) methods, we compare the performance of SOTA models with their SOTA-GSL counterparts within the same hyperparameter search space. Corresponding to the analysis of GCN and MLP in Section \ref{sec:ana_exp}, the SOTA-GSL methods include two variants: (1) SOTA, $\mathcal{G}'=\mathcal{G}$, which replaces the GSL graph $\mathcal{G}'$ with the original graph $\mathcal{G}$; and (2) SOTA, $\mathcal{G}'=\text{MLP}$, which substitutes the graph convolution layers of GSL $\mathcal{G}'$ with MLP layers. The results are presented in Table \ref{tab:sota_minus_gsl}, where "OOM" denotes out-of-memory. It is evident that removing GSL does not diminish model performance; in fact, it is often comparable to or even exceeds the original results. Furthermore, GSL-based SOTA methods require significantly more GPU memory and longer running times compared to their non-GSL counterparts. Based on these findings, we conclude that GSL not only fails to enhance performance across most datasets but also increases model complexity. In conjunction with the results in Table \ref{tab:baseline_plus_rewriting}, we assert that GSL is unnecessary for effective GNN design in most cases.

\begin{table}[h]
  \centering
\resizebox{1\hsize}{!}{
\begin{tabular}{lcccccccccccccccc}
\toprule
\multicolumn{1}{c}{} &
  \multicolumn{2}{c}{Questions} &
  \multicolumn{2}{c}{Minesweeper} &
  \multicolumn{2}{c}{Roman-empire} &
  \multicolumn{2}{c}{Amazon-ratings} &
  \multicolumn{2}{c}{Tolokers} &
  \multicolumn{2}{c}{Cora} &
  \multicolumn{2}{c}{Pubmed} &
  \multicolumn{2}{c}{Citeseer} \\ \cmidrule{2-17}
\multicolumn{1}{c}{\multirow{-2}{*}{Model}} &
  \multicolumn{1}{c}{AUC} &
  \multicolumn{1}{c}{Time} &
  \multicolumn{1}{c}{AUC} &
  \multicolumn{1}{c}{Time} &
  \multicolumn{1}{c}{Acc} &
  \multicolumn{1}{c}{Time} &
  \multicolumn{1}{c}{Acc} &
  \multicolumn{1}{c}{Time} &
  \multicolumn{1}{c}{AUC} &
  \multicolumn{1}{c}{Time} &
  \multicolumn{1}{c}{Acc} &
  \multicolumn{1}{c}{Time} &
  \multicolumn{1}{c}{Acc} &
  \multicolumn{1}{c}{Time} &
  \multicolumn{1}{c}{Acc} &
  \multicolumn{1}{c}{Time} \\ \midrule
GAug* &
  OOM &
  - &
  \textcolor{blue}{77.93$\pm$0.64} &
  - &
  \textcolor{red}{52.74$\pm$0.48} &
  - &
  \textcolor{blue}{48.42$\pm$0.39} &
  - &
  OOM &
  - &
  \textcolor{red}{82.48$\pm$0.66} &
  7s &
  \textcolor{blue}{78.73$\pm$0.77} &
  20s &
  71.66$\pm$1.14 &
  10s \\
GAug,~$\mathcal{G}'=\mathcal{G}$ &
  OOM &
  - &
  \textcolor{red}{80.56$\pm$0.36} &
  11s &
  OOM &
  - &
  \textcolor{red}{48.45$\pm$0.37} &
  12s &
  OOM &
  - &
  \textcolor{blue}{81.73$\pm$0.38} &
  1s &
  \textcolor{red}{79.17$\pm$0.23} &
  6s &
  \textcolor{blue}{72.34$\pm$0.18} &
  2s \\
GAug,~$\mathcal{G}'=\text{MLP}$ &
  OOM &
  - &
  64.31$\pm$1.40 &
  4.8s &
  OOM &
  - &
  48.05$\pm$0.66 &
  37s &
  OOM &
  - &
  78.90$\pm$0.00 &
  3.2s &
  77.40$\pm$0.00 &
  8.1s &
  \textcolor{red}{72.40$\pm$0.00} &
  9s \\ \midrule
GEN* &
  OOM &
  - &
  \textcolor{blue}{79.56$\pm$1.09} &
  260s &
  OOM &
  - &
  49.17$\pm$0.68 &
  - &
  \textcolor{red}{77.25$\pm$0.25} &
  - &
  \textcolor{blue}{81.66$\pm$0.91} &
  214s &
  \textcolor{blue}{80.40$\pm$1.85} &
  1384s &
  \textcolor{blue}{73.21$\pm$0.62} &
  470s \\
GEN,~$\mathcal{G}'=\mathcal{G}$ &
  OOM &
  - &
  \textcolor{red}{80.81$\pm$0.23} &
  75s &
  OOM &
  - &
  \textcolor{red}{50.08$\pm$0.30} &
  130s &
  OOM &
  - &
  \textcolor{red}{82.16$\pm$0.37} &
  39s &
  \textcolor{red}{80.49$\pm$0.13} &
  114s &
  71.52$\pm$0.34 &
  25s \\
GEN,~$\mathcal{G}'=\text{MLP}$ &
   OOM &
   - &
   71.81$\pm$0.98&
   12s &
   OOM &
   - &
  \textcolor{blue}{49.29$\pm$0.65} &
  49s &
   OOM &
   - &
  80.20$\pm$0.00 &
  140s &
  66.80$\pm$0.00 &
  1592s &
  \textcolor{red}{73.50$\pm$0.00} &
  310s \\ \midrule
GRCN* &
  \textcolor{blue}{74.50$\pm$0.84} &
  - &
  \textcolor{red}{72.57$\pm$0.49} &
  60s &
  44.31$\pm$0.34 &
  180s &
  \textcolor{red}{50.13$\pm$0.31} &
  220s &
  \textcolor{blue}{71.27$\pm$0.42} &
  37s &
  \textcolor{red}{83.82$\pm$0.25} &
  13s &
  \textcolor{blue}{78.84$\pm$0.32} &
  17s &
  \textcolor{red}{72.45$\pm$0.70} &
  20s \\
GRCN,~$\mathcal{G}'=\mathcal{G}$ &
  \textcolor{red}{75.69$\pm$0.52} &
  8s &
  71.15$\pm$0.05 &
  10s &
  \textcolor{blue}{45.84$\pm$0.52} &
  8s &
  46.07$\pm$1.02 &
  10s &
  \textcolor{red}{71.73$\pm$0.42} &
  10s &
  \textcolor{blue}{81.66$\pm$1.10} &
  2s &
  \textcolor{red}{79.12$\pm$0.26} &
  3s &
  69.55$\pm$1.28 &
  2s \\
GRCN,~$\mathcal{G}'=\text{MLP}$ &
  63.59$\pm$2.35 &
  3.9s &
  \textcolor{blue}{72.18$\pm$1.09} &
  2s &
  \textcolor{red}{45.89$\pm$0.83} &
  7.5s &
  \textcolor{blue}{48.77$\pm$0.60} &
  8.1s &
  70.45$\pm$1.39 &
  8s &
  79.40$\pm$0.00 &
  1.3s &
  78.10$\pm$0.00 &
  5s &
  71.40$\pm$0.00 &
  4.2s \\ \midrule
IDGL* &
  OOM &
  - &
  50.00$\pm$0.00 &
  157s &
  \textcolor{blue}{44.38$\pm$0.75} &
  186s &
  \textcolor{red}{45.87$\pm$0.58} &
  - &
  50.00$\pm$0.00 &
  279s &
  \textcolor{red}{84.40$\pm$0.49} &
  123s &
  \textcolor{blue}{76.63$\pm$1.55} &
  146s &
  \textcolor{blue}{72.88$\pm$0.59} &
  332s \\
IDGL,~$\mathcal{G}'=\mathcal{G}$ &
  OOM &
  - &
  \textcolor{blue}{50.00$\pm$0.00} &
  51s &
  41.24$\pm$0.86 &
  42s &
  OOM &
  - &
  \textcolor{blue}{50.00$\pm$0.00} &
  52s &
  82.43$\pm$0.45 &
  13s &
  73.50$\pm$1.85 &
  23s &
  \textcolor{red}{73.13$\pm$0.49} &
  36s \\
IDGL,~$\mathcal{G}'=\text{MLP}$ &
  OOM &
  - &
  \textcolor{red}{79.56$\pm$1.26} &
  13.7s &
  \textcolor{red}{50.35$\pm$0.36} &
  35s &
  \textcolor{blue}{39.93$\pm$0.88} &
  15s &
  \textcolor{red}{71.55$\pm$1.08} &
  11s &
  \textcolor{blue}{83.20$\pm$0.00} &
  6.6s &
  \textcolor{red}{79.20$\pm$0.00} &
  13s &
  72.60$\pm$0.00 &
  13.9s \\  \midrule
NodeFormer* &
  OOM &
  - &
  77.29$\pm$1.71 &
  - &
  \textcolor{blue}{56.54$\pm$3.73} &
  - &
  \textcolor{blue}{41.33$\pm$1.25} &
  - &
  OOM &
  - &
  \textcolor{blue}{78.81$\pm$1.21} &
  213s &
  \textcolor{red}{78.38$\pm$1.94} &
  - &
   70.39$\pm$2.04 &
  219s \\
NodeFormer,~$\mathcal{G}'=\mathcal{G}$ &
  OOM &
  - &
  \textcolor{red}{80.66$\pm$0.82} &
  215s &
  \textcolor{red}{68.37$\pm$1.95} &
  236s &
  OOM &
  - &
  OOM &
  - &
  77.01$\pm$1.99 &
  152s &
  OOM &
  - &
  \textcolor{blue}{70.82$\pm$0.13} &
  139s \\
NodeFormer,~$\mathcal{G}'=\text{MLP}$ &
  OOM &
  - &
  \textcolor{blue}{80.04$\pm$1.42} &
   21s &
  53.08$\pm$2.37 &
  7.2s &
  \textcolor{red}{71.55$\pm$1.08} &
  26s &
  OOM &
  - &
  \textcolor{red}{78.82$\pm$0.00} &
  8s &
  \textcolor{blue}{76.30$\pm$0.00} &
  127s &
  \textcolor{red}{72.80$\pm$0.00} &
  15s \\ \midrule
GloGNN &
  \textcolor{blue}{68.67$\pm$1.07} &
  66.6s &
  \textcolor{red}{52.45$\pm$0.30} &
  13.0s &
  \textcolor{blue}{66.21$\pm$0.17} &
  26.1s &
  \textcolor{red}{50.72$\pm$0.88} &
  31.1s &
  \textcolor{blue}{79.81$\pm$0.20} &
  47.4s &
  \textcolor{red}{78.07$\pm$1.66} &
  6.6s &
  \textcolor{red}{87.88$\pm$0.26} &
  18.2s &
  71.95$\pm$1.90 &
  21.8s \\  
GloGNN,~$\mathcal{G}'=\mathcal{G}$ &
  68.32$\pm$1.23 &
  49.4s &
  52.30$\pm$0.21 &
  3.6s &
  66.03$\pm$0.14 &
  15.3s &
  \textcolor{blue}{50.23$\pm$0.83} &
  21.7s &
  \textcolor{red}{80.02$\pm$0.16} &
   25.1s &
  73.49$\pm$2.01 &
  5.1s &
  87.62$\pm$0.20 &
  14.4s &
  \textcolor{red}{72.27$\pm$2.08} &
21.2s \\
GloGNN,~$\mathcal{G}'=\text{MLP}$ &
  \textcolor{red}{69.69$\pm$0.22} &
  25.7s &
  \textcolor{blue}{52.30$\pm$0.20} &
   2.1s &
  \textcolor{red}{66.49$\pm$0.16} &
  12.4s &
  49.56$\pm$0.73 &
  12.3s &
  74.85$\pm$0.12 &
  2.8s &
  \textcolor{blue}{73.93$\pm$1.81} &
  3.2s &
  \textcolor{blue}{87.64$\pm$0.27} &
   10.2s &
  \textcolor{blue}{72.09$\pm$1.81} &
   13.8s\\   \midrule
WRGAT &
  OOM &
  - &
  \textcolor{red}{90.22$\pm$0.64} &
  168.0s &
  OOM &
  - &
  OOM &
  - &
  \textcolor{blue}{78.69$\pm$1.21} &
  153.0s &
  \textcolor{red}{84.28$\pm$1.52} &
  19.5s &
  \textcolor{blue}{88.82$\pm$0.50} &
   421.6s &
  \textcolor{red}{73.50$\pm$1.41} &
   22.1s \\
WRGAT,~$\mathcal{G}'=\mathcal{G}$ &
  \textcolor{red}{74.67$\pm$0.95} &
  64.1s &
  \textcolor{blue}{89.79$\pm$0.37} &
  18.6s &
  OOM &
  - &
  \textcolor{red}{50.41$\pm$0.53} &
   49.9s &
  \textcolor{red}{78.81$\pm$0.89} &
  47.0s &
  \textcolor{blue}{83.48$\pm$1.48} &
   3.4s &
  \textcolor{red}{88.92$\pm$0.43} &
 26.5s &
  \textcolor{blue}{73.22$\pm$1.90} &
  4.7s \\
WRGAT,~$\mathcal{G}'=\text{MLP}$ &
  \textcolor{blue}{68.07$\pm$2.62} &
  75.8s &
  87.08$\pm$2.11 &
  16.2s &
  OOM &
  - &
  \textcolor{blue}{41.38$\pm$1.46} &
   24.4s &
  76.41$\pm$1.25 &
  37.7s &
  76.99$\pm$1.10 &
  2.9s &
  80.27$\pm$6.23 &
  23.9s &
  65.28$\pm$2.11 &
  4.5s \\  \midrule
WRGCN &
  \textcolor{blue}{74.70$\pm$1.71} &
  358.3s &
  \textcolor{blue}{90.63$\pm$0.64} &
   40.9s &
  OOM &
  - &
  \textcolor{red}{52.76$\pm$0.95} &
   508.4s &
  \textcolor{red}{82.68$\pm$0.82} &
   52.3s &
  \textcolor{blue}{88.30$\pm$1.46} &
   23.7s &
  OOM &
  - &
  \textcolor{blue}{73.74$\pm$1.60} &
   54.2s \\
WRGCN,~$\mathcal{G}'=\mathcal{G}$ &
  \textcolor{red}{75.91$\pm$1.30} &
  43.3s &
  \textcolor{red}{90.65$\pm$0.49} &
  5.5s &
  OOM &
  - &
  \textcolor{blue}{52.54$\pm$0.56} &
   50.1s &
  \textcolor{blue}{82.65$\pm$0.86} &
  15.6s &
  \textcolor{red}{88.32$\pm$0.79} &
  3.9s &
  \textcolor{red}{89.26$\pm$0.45} &
   19.4s &
  \textcolor{red}{74.45$\pm$1.51} &
 10.5s \\
WRGCN,~$\mathcal{G}'=\text{MLP}$ &
  64.59$\pm$1.48 &
  23.1s &
  70.66$\pm$1.37 &
  7.7s &
  OOM &
  - &
  37.05$\pm$0.46 &
  8.0s &
  69.10$\pm$0.91 &
  12.2s &
  70.00$\pm$3.59 &
  2.2s &
  \textcolor{blue}{67.29$\pm$2.49} &
   9.9s &
  70.84$\pm$1.36 &
  4.1s\\
\bottomrule
\end{tabular}
}
\caption{Model Performance and training time per epoch of SOTA methods and SOTA-GSL. The results for methods marked with ``*" are reported in \citet{OpenGSL}.
% , where OOM means out-of-memory and the best and second models are highlighted in \textcolor{red}{red} and \textcolor{blue}{blue}.
}\label{tab:sota_minus_gsl}
\end{table}

\vspace{-0.3cm}
\subsection{Quality of GSL Graphs}\label{sec:exp_gsl_quality}

\begin{figure}[h]
    \centering
    \subfloat{%
       \includegraphics[width=0.25\textwidth]{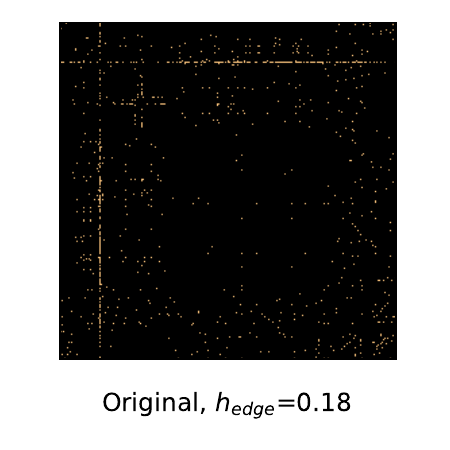}}
    \hfill
    \subfloat{%
        \includegraphics[width=0.25\textwidth]{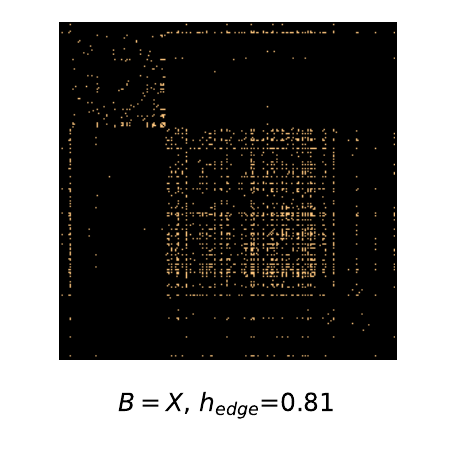}}
    \hfill
    \subfloat{%
       \includegraphics[width=0.25\textwidth]{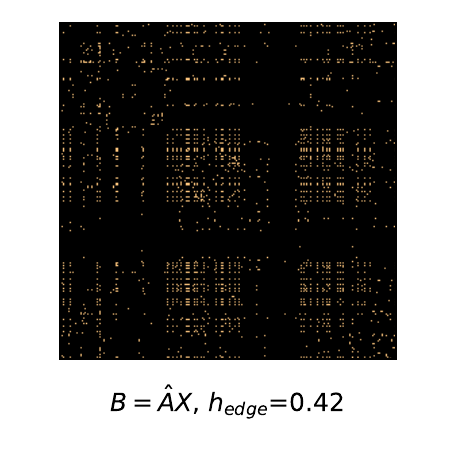}}
    \hfill
    \subfloat{%
        \includegraphics[width=0.25\textwidth]{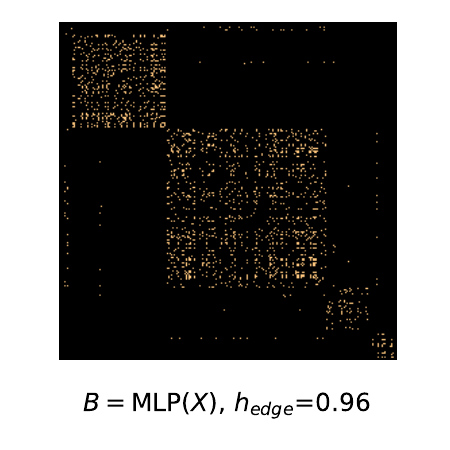}}
    \hfill
    \vspace{-0.4cm}
    \subfloat{%
       \includegraphics[width=0.25\textwidth]{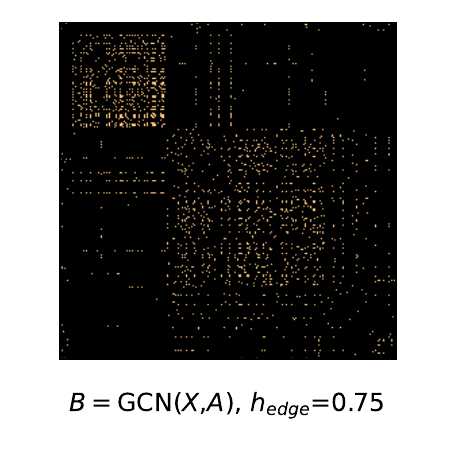}}
    \hfill
    \subfloat{%
        \includegraphics[width=0.25\textwidth]{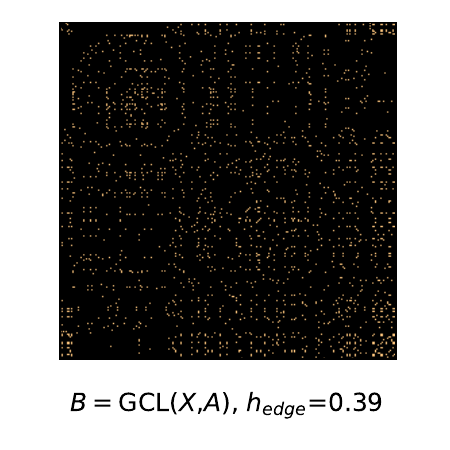}}
    \hfill
    \subfloat{%
        \includegraphics[width=0.25\textwidth]{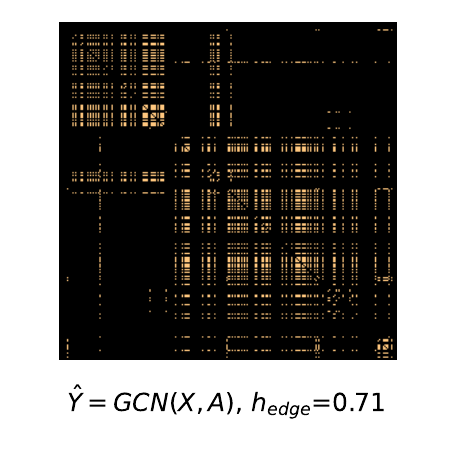}}
    \hfill
    \subfloat{%
       \includegraphics[width=0.25\textwidth]{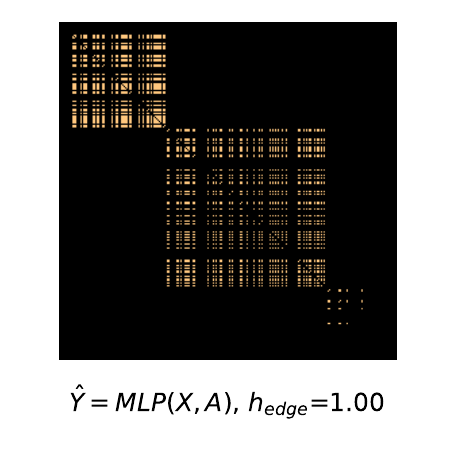}}
    \hfill
\caption{Visualization of original graph and reconstructed graphs on Wisconsin}
\label{fig:gsl_quality}
\end{figure}

Previous studies \citep{GloGNN,HiGNN} suggest that GSL constructs graphs with properties that improve intra-class node connectivity, which can be measured by homophily. This improvement can be visualized by inspecting graph structures with nodes sorted by their class labels. A graph that appears closer to a block diagonal matrix indicates stronger intra-class connectivity. However, this enhancement may not always be essential and can be achieved through non-GSL methods as well. In Figure \ref{fig:gsl_quality}, we visualize the original and reconstructed structures of a heterophilous graph from the Wisconsin dataset. The GSL graphs are constructed using various bases: $\mathbf{X}$, $\mathbf{\hat{A}}\mathbf{X}$, $\text{MLP}(\mathbf{X})$, $\text{GCN}(\mathbf{X},\mathbf{A})$, and $\text{GCL}(\mathbf{X},\mathbf{A})$. We also include reconstructed graphs using a simple method that samples edges between nodes of the same class based on label predictions, \ie{} $\mathbf{\hat{Y}}=\text{GCN}(\mathbf{X},\mathbf{A})$ or $\mathbf{\hat{Y}}=\text{MLP}(\mathbf{X},\mathbf{A})$. Figure \ref{fig:gsl_quality} demonstrates that, although GSL improves intra-class connectivity, the improvement is not as substantial as that achieved by non-GSL methods, as seen in the last two subfigures. Thus, the improvement in homophily within GSL graphs is unnecessary, as it can be easily achieved through simple methods.

\vspace{-0.3cm}
\subsection{GSL Components}\label{sec:exp_gsl_components}

\begin{figure}[h]
    \centering
  \subfloat{%
       \includegraphics[width=0.5\textwidth]{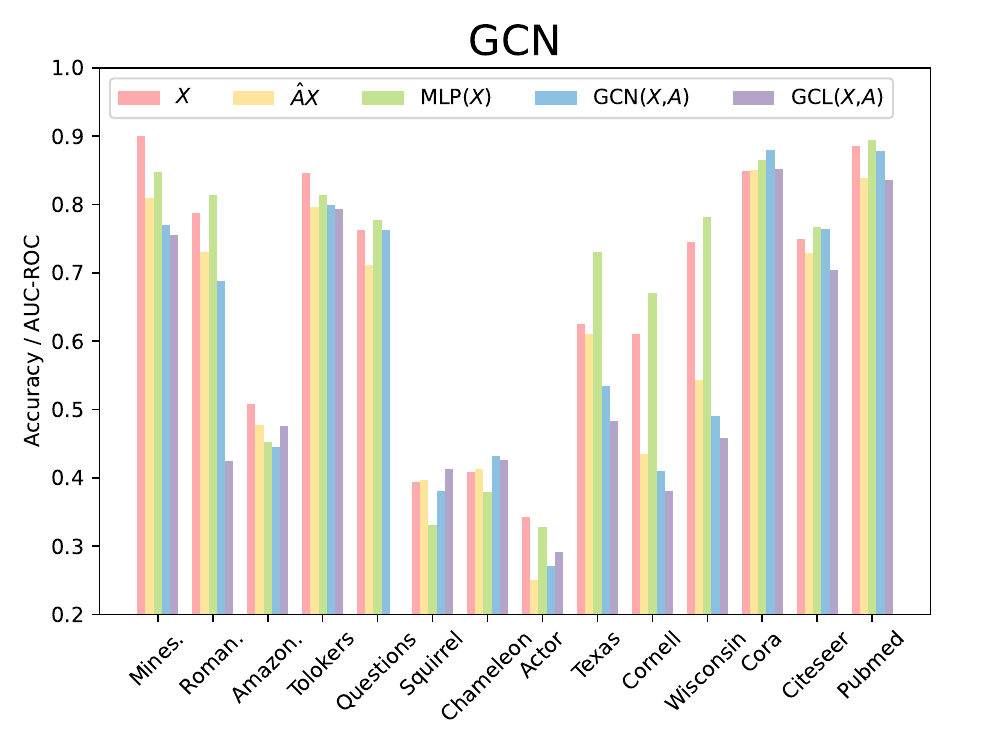}}
    \hfill
  \subfloat{%
        \includegraphics[width=0.5\textwidth]{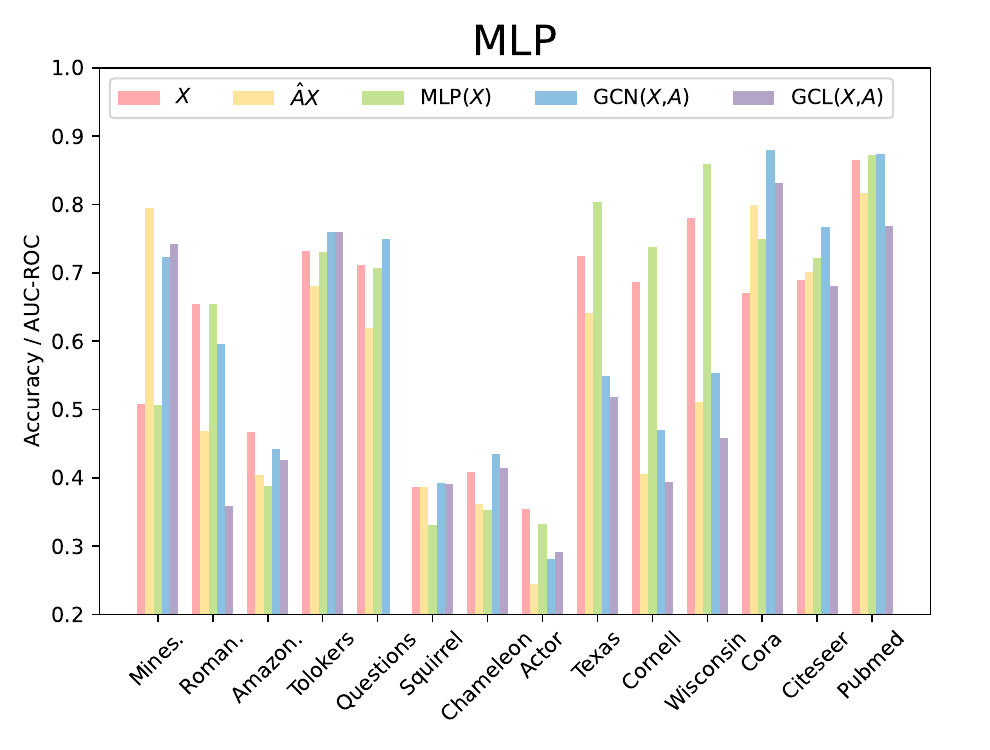}}
    \hfill
    \caption{Influences of different GSL bases to model performance.}
    \label{fig:gsl_basis}
\end{figure}

Since the performance of GNN and GNN+GSL models is comparable under the same GSL bases, as shown in Table \ref{tab:baseline_plus_rewriting}, we further investigate how different types of GSL bases influence GNNs. As illustrated in Figure \ref{fig:gsl_basis}, the original node features are not always the optimal input for GCN and MLP, which explains why prior comparisons of GNNs are unfair. Pretrained node representations, such as $\text{MLP}(\mathbf{X})$ and $\text{GCN}(\mathbf{X},\mathbf{A})$, significantly enhance model performance on several datasets, including Texas, Cornell, and Wisconsin. This improvement stems from self-training, a key component in many GSL approaches. As a result, incorporating self-training methods may be more advantageous for future GNN designs compared to GSL. For more results and analysis of GSL modules, please refer to Appendix \ref{apd:additional_exp}.

% TASK LIST:

% 1. MLP, GNN, GNN + GSL

% 2. SOTA, SOTA - GSL

% 3. Compare effectiveness of different GSL design on GNN+GSL

% 4. Compare different GNNs, different GSL basis on GNN

% 5. Measure the quality of reconstructed graph (ratio of homophily edges), measure the sparsity, connectivity, radius, 

% 6. Training time comparison.

\vspace{-0.3cm}

\section{Conclusion}

In this paper, we revisit the role of Graph Structure Learning (GSL) in Graph Neural Networks (GNNs) with our proposed GSL framework. Motivated by the controversy of GSL, we demonstrate that graph convolution over GSL-constructed graphs does not improve mutual information, as confirmed by both empirical observations and theoretical analysis. By either adding GSL to baseline GNNs or removing it from state-of-the-art (SOTA) methods, we find that GSL does not enhance GNN performance when evaluated under the same GSL bases and hyperparameter tuning. These results suggest that the improvements attributed to GSL may stem from components other than GSL. Our findings contribute to a better understanding of GSL and offer insights into re-evaluating the essential components in future GNN design.

\newpage
\bibliography{iclr2025_conference}
\bibliographystyle{iclr2025_conference}

\appendix

\newpage
\section{Taxonomy of Graph Structure Learning Methods}\label{apd:gsl_method_explanation}

We present several representative GSL-based GNNs within our proposed GSL framework in Table \ref{tab:gsl_framework}. Below, we provide a detailed description of each method.

\begin{table}[h]
    \centering
    \resizebox{1\hsize}{!}{
    \begin{tabular}{cccccc}
        \toprule
        Method & Bases & Construct & Refinement & View Fusion & Training Mode \\ \midrule
        
        LDS \citep{LDS} 
        & $\b{X}$  
        & $\{\mathcal{E}'=kNN(\b{B})\}$ +Opt. 
        & Bernoulli($\mathcal{E}'$) 
        & Late Fusion, $\{\mathcal{G}'_1,\mathcal{G}'_2,\dots,\mathcal{G}'_m\}$,
        $\theta_1 = \theta_2$
        & 2-stage \\
        
        Geom-GCN \citep{Geom-GCN} 
        & \makecell{Isomap/Poincare/ \\$\text{Struc2vec}(\b{X},\b{A})$}
        & $\{\mathcal{E}'|e_{ij}'=\abs{\b{B}_i-\b{B}_j}\}$ 
        & $\text{threshold}(\mathcal{E}')$
        & Late Fusion, $\{\mathcal{G},\mathcal{G}'\}$,
        $\theta_1 \neq \theta_2$
        & Static \\
        
        ProGNN \citep{ProGNN} 
        & $\b{\epsilon}$
        & $\{\mathcal{E}'=\text{Opt}(\b{\epsilon})\}$
        & \makecell{Low Rank+Sparsity\\+Original}
        & No Fusion, $\{\mathcal{G}'\}$
        & Joint \\
        
        IDGL \citep{IDGL} 
        & MLP($\b{X}$) 
        & $\{\mathcal{E}'|e_{ij}'=\text{cos}(\b{B}_i$,$\b{B}_j)\}$ 
        & topk($\mathcal{E}'$)
        & Early Fusion, $\{\mathcal{G}+\mathcal{G}'\}$
        & Joint
        \\

        GRCN \citep{GRCN}
        & GCN($\b{X}$,$\b{A}$)
        & $\{\mathcal{E}'|e_{ij}'=\sigma(\b{B}_i\b{B}_j^T)\}$ 
        & topk($\mathcal{E}'$), sym($\mathcal{E}'$)
        & Early Fusion, $\{\mathcal{G}+\mathcal{G}'\}$
        & Joint
        \\

        GAug-M \citep{GAug}
        & $\text{GCN}^{(2)}$($\b{X}$,$\b{A}$)
        & $\{\mathcal{E}'|e_{ij}'=\sigma(\b{B}_i\b{B}_j^T)\}$ 
        & \makecell{$\mathcal{G}'_+ = \text{topk}(\mathcal{E}')$, \\ $\mathcal{G}'_- = \text{bottom}(\mathcal{E}')$}
        & Early Fusion, $\{\mathcal{G}+\mathcal{G}'_+ - \mathcal{G}'_-\}$
        & Joint 
        \\
        
        GAug-O \citep{GAug}
        & $\b{X}$
        & $\{\mathcal{E}'|e_{ij}'=p(e_{ij}|\text{GAE}(\b{B},\b{A}))\}$ 
        & Gumbel($\mathcal{E}'$)
        & Early Fusion, $\{\mathcal{G}+\mathcal{G}'\}$
        & Joint 
        \\
        
        SLAPS \citep{SLAPS}
        & $\text{MLP}(\b{X})$
        & $\{\mathcal{E}'=kNN(\b{B})\}$
        & norm($\mathcal{E}'$),sym($\mathcal{E}'$)
        & No Fusion, $\{\mathcal{G}'\}$
        & Joint 
        \\
        
        CoGSL \citep{CoGSL}
        & \makecell{$\text{GCN}(\b{X},\{\b{A},\text{kNN}(\b{X}),$\\$\text{PPR}(\b{X}),\text{Subgraph}(\b{X})\})$}
        & $\{\mathcal{E}'|e_{ij}'=p(e_{ij}|\text{MLP}(\b{B},\b{A}))\}$ 
        & -
        & Early Fusion, $\{\mathcal{G}^*|\min{\mathcal{L}_\text{CL}(\mathcal{G},\mathcal{G}')}\}$,
        $\theta_1 \neq \theta_2$
        & 2-stage
        \\
        
        GEN \citep{GEN}
        & $\text{GCN}(\b{X},\b{A})$
        & $\{\mathcal{E}'=kNN(\b{B})\}$
        & -
        & Late Fusion, $\{\mathcal{G}'_1,\mathcal{G}'_2,\dots,\mathcal{G}'_m\}$ ,
        $\theta_1 \neq \theta_2$
        & 2-stage 
        \\
        
        STABLE \citep{STABLE}
        & $\text{GCL}(\b{X},\b{A})$
        & \makecell{$\{\mathcal{E}'|e_{ij}'=\text{cos}(\b{B}_i$,$\b{B}_j)$ \\ or $\text{cos}(\b{B}_i$,$\b{B}_j)\}$}
        & \makecell{$\mathcal{G}'_+ = \text{topk}(\mathcal{E}')$, \\ $\mathcal{G}'_- = \text{threshod}(\mathcal{E}')$}
        & Early Fusion, $\{\mathcal{G}+\mathcal{G}'_+ - \mathcal{G}'_-\}$
        & Joint
        \\
        
        SEGSL \citep{SEGSL}
        & $\b{X}$
        & \makecell{ $\{\mathcal{E}'| \min{\mathcal{H}_S}, $ \\$e_{ij}'\in\text{EncTree}(\text{kNN}(\b{B}))\}$}
        & -
        & No Fusion, $\{\mathcal{G}'\}$
        & Joint
        \\
        
        SUBLIME \citep{SUBLIME}
        & $\text{GCN}(\b{X},\b{A})$
        & \makecell{$\{\mathcal{E}'=\text{Opt}(\b{\epsilon})\}$ or \\ $\{\mathcal{E}'|e_{ij}'=\text{cos}/\text{Minkowski}(\b{B}_i$,$\b{B}_j)\}$ }
        & topk($\mathcal{E}'$),sym($\mathcal{E}'$),norm($\mathcal{E}'$)
        & Separation, $\{\mathcal{G},\mathcal{G}'\}$, 
        $\theta_1 = \theta_2$
        & Joint
        \\
        
        BM-GCN \citep{BM-GCN}
        & \makecell{$\hat{\b{Y}} = \text{MLP}(\b{X})$, \\$\min\mathcal{L}_{\text{CE}}(\hat{\b{Y}}, \b{Y})$}
        & $\{\mathcal{E}'=\b{BQB}^T\}$
        & norm($\mathcal{E}'$)
        & Early Fusion, $\{\mathcal{G}\odot\mathcal{G}'\}$
        & Joint
        \\
        
        WSGNN \citep{WSGNN}
        & MLP($\b{X}$)
        & $\{\mathcal{E}'|e_{ij}'=cos(\b{B}_i,\b{B}_j)\}$
        & -
        & Early Fusion, $\{\mathcal{G}+\mathcal{G}'\}$
        & Joint
        \\
        
        GLCN \citep{GLCN}
        & $\b{X}$
        & $\{\mathcal{E}'|e_{ij}'=\phi(\abs{\b{B}_i-\b{B}_j})\}$
        & \makecell{norm($\mathcal{E}'$), Original \\ +Sparsity+Smoothness}
        & No Fusion, $\{\mathcal{G}'\}$
        & Joint
        \\
        
        ASC \citep{ASC} 
        & SpectralCluster($\b{X}$)
        & $\{\mathcal{E}'|e_{ij}'=\norm{\b{B}_i-\b{B}_j}\}$ 
        & $\text{topk}(\mathcal{E}')$
        & No Fusion, $\{\mathcal{G}'\}$
        & Static \\
        
        WRGAT \citep{WRGAT}
        & GCN($\b{X}$, $\b{A}$)
        & $\{\mathcal{E}'| e_{ij}' \cdot Opt(\b{B})\}$
        & Sparsity + MultiHop
        & Early Fusion $\{\mathcal{G}+\mathcal{G}'\}$
        & Static \\

        HOG-GCN \citep{HOG-GCN}
        & GCN($\b{X}$, $\b{A}$)
        & $\{\mathcal{E}'| e_{ij}' = \sigma(\b{B_i}\b{B_j}^T) \}$
        & Sparsity + Smoothness
        & No Fusion $\{\mathcal{G}'\}$
        & Joint \\

        GGCN \citep{GGCN}
        & MLP($\b{X}$)
        & $\{\mathcal{E}'|  e_{ij}' = \text{cos}(\b{B_i}, \b{B_j})\}$
        & Low Rank + Sparsity
        & Early Fusion, $\{\mathcal{G}+\mathcal{G}'\}$
        & Joint \\
        
        GloGNN \citep{GloGNN} 
        & $\text{MLP}(\b{X})$
        & $\{\mathcal{E}'=\text{Opt}(\b{B})\}$
        & Sparsity+MultiHop
        & No Fusion, $\{\mathcal{G}'\}$
        & Joint \\
        
        HiGNN \citep{HiGNN} 
        & \makecell{$\hat{\b{Y}} = \text{GCN}(\b{X},\b{A})$,\\ $\min\mathcal{L}_{CE}(\hat{\b{Y}},\b{Y})$}
        & $\{\mathcal{E}'=e_{ij}'=\text{cos}(\b{B_i},\b{B_j}))\}$
        & topk($\mathcal{E}'$), sym($\mathcal{E}'$)
        & Late Fusion, $\{\mathcal{G},\mathcal{G}'\}$, $\theta_1 \neq \theta_2$
        & Static \\
        \bottomrule
    \end{tabular}
    }
    \caption{Representative GSL methods under our proposed GSL framework}
    \label{tab:gsl_framework}
\end{table}

\textbf{LDS} \citep{LDS}. The GSL bases in LDS is constructed as node features $\b{X}$ and the GSL graph $\mathcal{G}'$ is initialized using a k-Nearest-Neighbors algorithm based on $\b{B}$. Then, $\mathcal{G}'$ is updated with a loss function of node classification. Then multiple graphs are sampled based on $\mathcal{G}'$ with a Bernoulli function and used to update the model parameters. The $\mathcal{G}'$ construction and model parameters are updated as a 2-stage mode.

\textbf{Geom-GCN} \citep{Geom-GCN}. Geom-GCN constructs the GSL bases from several graph-aware node embedding strategies using both of the $\b{X}$ and $\b{A}$: Isomap \citep{}, Poincare \citep{}, and struc2vec \citep{}. Then, new graphs are constructed by filtering node pairs with a higher similarity measured by Euclidean distance $\{\mathcal{E}'|e_{ij}' = \abs{\b{B}_i-\b{B}_j} < \delta \}$ where $\delta$ is a threshold. Finally, both of the aggregated message from $\mathcal{G}$ and $\mathcal{G}'$ are fused after applying graph convolution layers with no parameter sharing. The $\mathcal{G}'$ is not updated through the training process.

\textbf{ProGNN} \citep{ProGNN}. The $\mathcal{G}'$ in ProGNN is purely learned by optimization without GSL bases. It optimizes the $\mathcal{G}'$ using low rank, sparsity, and similarity with the original graphs $\mathcal{G}$. It outputs a single graph $\mathcal{G}'$ without fusion and updates the $\mathcal{G}'$ together with model parameters.

\textbf{IDGL} \citep{IDGL}. The GSL bases in LDS is constructed by linear transformation of node features $\text{MLP}(\b{X})$. Then, a GSL graph $\mathcal{G}'$ is constructed using cosine similarity with topk threshold refinement. The early fusion is applied by fusing GSL graph $\mathcal{G}'$ with original graph $\mathcal{G}$ before training. The GSL graph $\mathcal{G}'$ is trained with model parameters jointly.

\textbf{GRCN} \citep{GRCN}. GRCN constructs GSL bases by node embeddings of graph convolution $\text{GCN}(\b{X},\b{A})$. Then, the GSL graph $\mathcal{G}'$ is constructed by a kernel function with topk and symmetrization refinement $\{\mathcal{E}'|e_{ij}' = \sigma(\b{B}_i\b{B}_j) > \delta \}$. The final graph is obtained by early fusion and the GSL graph $\mathcal{G}'$ is updated together with model parameters.

\textbf{GAug-M} and \textbf{GAug-O} \citep{GAug}. GAug-M constructs GSL bases using a 2-layer graph convolution $\text{GCN}^{(2)}(\b{X},\b{A})$. Then, the GSL graph $\mathcal{G}'$ is constructed by a kernel function. The final graph is obtained by adding some edges with highest probabilities and removing some edges with lowest probabilities on $\mathcal{G}$. GAug-O selects node features as GSL bases $\b{X}$, then trains a Graph Auto-Encoder to predict edges as $\mathcal{G}'$. Then, after gumbel sampling, the GSL graph $\mathcal{G}'$ is fused with original graph $\mathcal{G}$ before training. The $\mathcal{G}'$ in both of the GAug-M and GAug-O is updated together with model parameters.

\textbf{SLAPS} \citep{SLAPS}. SLAPS constructs the GSL bases by applying MLP(X) followed by a k-nearest neighbors (kNN) algorithm based on node feature similarities. The GSL graph $\mathcal{G}'$ is then processed by an adjacency processor that symmetrizes and normalizes the adjacency matrix to ensure non-negativity and symmetry. The final graph is obtained of the generated graph $\mathcal{G}'$ with the node features without fusion. Additionally, a self-supervised denoising autoencoder $L_{DAE}=L(X_i,GNN_{DAE}(\hat{X}_i;\theta_{GNN_{DAE}}))$ is introduced to address the supervision starvation problem, updating $\mathcal{G}'$ together with the model parameters.

\textbf{CoGSL} \citep{CoGSL}. CoGSL constructs GSL bases using two views, one of them is the Origin graph. Another is  selected from the Adjacency matrix $A$, Diffusion matrix $PPR(X)$, the KNN graph $KNN(X)$ and the Subgraph of the Origin. GCNs are applied to these views to obtain node embeddings. The GSL graph  is constructed by applying a linear transformation to the node embeddings of each node pair to estimate the connection probability between them. This connection probability is then added to the original view to finalize the graph. The refinement $\mathcal{E}'|e_{ij}'=p(e_{ij}|\text{MLP}(\mathbf{B},\mathbf{A}))$  step involves maximizing the mutual information between the two selected views and the newly constructed graph. InfoNCE loss is used to optimize the connection probability, where the same node serves as a positive sample, and different nodes serve as negative samples. The final graph $\mathcal{G}'$ is obtained via early fusion of the selected views, and the GSL graph is updated with model parameters.

\textbf{GEN} \citep{GEN}. GEN constructs the GSL bases by generating kNN graphs though several GCN layer, utilizing node representations from different layers. These kNN graphs are then combined using a Stochastic Block Model (SBM) to create a new graph $\mathcal{G}'$. The GSL graph $\mathcal{G}'$ is refined iteratively through Bayesian inference to maximize posterior probabilities $P(G,\alpha,\beta|O,Z,Y_{l})=\frac{P(O|G,\alpha,\beta)P(G,\alpha,\beta)P(O,Z,Y_{l})}{P(O,Z,Y_l)}$, considering both the original graph and node embeddings. The final graph is obtained by feeding the graph $Q$ back into the GCN for further optimization. The iterative process updates both the GSL graph and GCN parameters as a 2-stage mode, providing mutual reinforcement between the graph estimation and model learning.

\textbf{STABLE} \citep{STABLE}. STABLE constructs the GSL bases by generating augmentations based on node similarity through kNN graph and perturbing edges to simulate adversarial attacks. The GSL graph $\mathcal{G}'$ is constructed by refining the structure using contrastive learning between positive samples (slightly perturbed graphs) and negative samples (undesirable views generated by feature shuffling). The refinement step applies a top-k filtering strategy on the node similarity matrix to retain helpful edges while removing adversarial ones. The final graph is obtained through early fusion, and the GSL graph $\mathcal{G}'$ is updated together with model parameters during joint training

\textbf{SE-GSL} \citep{SEGSL}. SE-GSL constructs the GSL bases using a kNN graph fused with the original graph. The GSL graph $\mathcal{G}'$ is constructed through a structural entropy minimization process that extracts hierarchical community structures in the form of an encoding tree. The final graph is optimized by sampling node pairs from the encoding tree and generating new edges based on the minimized entropy structure. The refined graph is then used for downstream tasks, and the GSL graph $\mathcal{G}'$ is updated jointly with model parameters during training.

\textbf{SUBLIME} \citep{SUBLIME}. SUBLIME constructs the GSL bases using both an anchor view (original graph) and a  learner view (new graph). The new graph is initialized through kNN and further optimized either by parameter-based methods (using models like MLP, GCN, or GAT) or by non-parameter-based approaches (using cosine similarity or  Minkowski distance). After obtaining the new graph, post-processing operations such as top-k filtering, symmetrization, and degree-based regularization are applied to ensure the graph’s sparsity and structure. The GSL graph $\mathcal{G}'$ is refined by applying contrastive learning between the anchor and learner views, incorporating edge drop and feature masking to generate node embeddings. The final graph is used in downstream tasks, and both views are updated together with model parameters in a joint training process.

\textbf{BM-GCN} \citep{BM-GCN}. BM-GCN constructs the GSL bases by introducing soft labels for nodes enbedding $\mathbf{B}=softmax(\sigma(MLP(X)))$ via a multilayer perceptron $\mathcal{L}_{MLP}=\sum_{v_i\in\mathcal{V}}f(B_i,Y_i)$. These soft labels are then used to compute a  block matrix (H) , which models the connection probabilities between different node classes. The GSL graph $\mathcal{G}'$ is constructed by creating a block similarity matrix $Q=HH^{T}$  from the block matrix $Y_s={Y_i,B_i|\forall v_i\in\mathcal{T}_{y}, \forall v_j\notin\mathcal{T}_{y}}, H=(Y_{s}^{T}AY_s)\circ (Y_{s}^{T}AE)$, reflecting similarities between classes. The new graph is optimized using $BQB^{T}$ and further fused with the original graph $A+\beta I$ for downstream tasks. The final graph is obtained by optimizing $\mathcal{G}'$ through degree-based regularization and  top-k filtering. The GSL graph $\mathcal{G}'$ is updated together with model parameters during joint training.

\textbf{WSGNN} \citep{WSGNN}. WSGNN introduces a two-branch graph structure learning method, where each branch operates on different aspects of the graph: Branch AZ learns node labels from the new graph structure, while Branch ZA learns the new graph structure from the labels. The GSL bases is constructed using the observed graph $A_{obs}$ and node features $X$. The new graph $A'$ is inferred via cosine similarity between node embeddings. After constructing two separate views from each branch, the final graph is obtained by averaging the graphs from both branches. The refinement process ensures sparsity through cosine-based edge calculation  $\mathcal{E}'|e_{ij}'=cos(\mathbf{B}_i,\mathbf{B}_j)$. Finally, both views undergo early fusion, with graph structure and node labels optimized jointly using a composite loss function that includes ELBO for structure prediction and cross-entropy loss for label prediction. The final GSL graph $\mathcal{G}'$ is updated during joint training.

\textbf{GLCN} \citep{GLCN}. GLCN constructs the GSL bases by computing pairwise distances between node features and passing them through an MLP to obtain a block similarity score. This score is then processed with a softmax function to generate an $n \times n$ probability matrix that serves as the learned graph structure. The graph is refined using regularization techniques to ensure sparsity and feature smoothness $L_{GL}=\sum_{i,j=1}^{n}||x_i-x_j||_{2}^{2}S_{ij}+\gamma||S||_{F}^{2}+\beta||S-A||_{F}^{2}$. The learned graph is then used for downstream graph tasks, where the task loss and the graph regularization loss are jointly optimized during joint training

\textbf{ASC} \citep{ASC}. ASC constructs the GSL bases is formed by using pseudo-eigenvectors from spectral clustering. They divide the Laplacian spectrum into slices, with each slice corresponding to an embedding matrix. The GSL graph $\mathcal{G}'$ is constructed by adaptive spectral clustering, where pseudo-eigenvectors are weighted based on alignment with node labels  Where $f^\mathcal{Z}_i$. For refinement, they apply top-K edge selection by minimizing node embedding distance and maximizing homophily $\underset{\mathcal{Z}}{\text{argmin}}\sum_{i,j\in V_Y}(d(f^Z_i ,f^Z_j ), 1(y_i, y_j))$. This final restructured graph is training without fusion. Finally, the GSL graph is updated together with the model parameters.

\textbf{WRGAT} \citep{WRGAT}. WRGAT constructs the GSL bases using the node features and a weighted relational GNN (WRGNN) framework that fuses structural and proximity information. A multi-relational graph is built by assigning different types of edges based on the structural equivalence of nodes at various neighborhood levels. This framework adapts to both assortative and disassortative mixing patterns, which helps improve node classification tasks. The GSL graph $\mathcal{G}'$ is refined through attention-based message passing across these relational edges, and early fusion of proximity and structural features is used. The GSL graph $\mathcal{G}'$ is trained jointly with the model parameters to optimize the node classification task.

\textbf{HOG-GCN} \citep{HOG-GCN}. HOG-GCN constructs the GSL bases by incorporating both topological information and node attributes to estimate a homophily degree matrix $S=BB^T,B=softmax(Z_m),Z_{m}^{(l)}=\sigma(Z_{m}^{(l-1)W_{m}^{(l)}})$. The GSL graph $\mathcal{G}'$ is constructed using a homophily-guided propagation mechanism, which adapts the feature propagation weights between neighborhoods based on the homophily degree matrix $Z^{(l)}=\sigma(\mu Z^{(l-1)}W_e^{(l)}+\xi\hat{D}^{(-1)}A_k\odot H Z^{(l-1)}W^{(l)}_n )$. For refinement, the graph incorporates both k-order structures and class-aware information to model the homophily and heterophily relationships between nodes. The final graph is obtained through joint fusion of topological and attribute-based homophily degrees, and both graph structure and model parameters are updated during  joint training.

\textbf{GGCN} \citep{GGCN}. GGCN constructs the GSL bases using node features and structural properties such as node-level homophily $h_i$ and relative degree $\bar{r_i} $. It incorporates structure-based edge correction by learning new edge weights derived from structural properties like node degree, and feature-based edge correction by learning signed edge weights from node features, allowing for positive and negative influences between neighbors. The GSL graph $\mathcal{G}'$ is constructed by combining signed and unsigned edge information, aiming to capture both homophily and heterophily. The refinement process uses edge correction and decaying aggregation to mitigate oversmoothing and heterophily problems. The final graph is updated with early fusion, and the GSL graph $\mathcal{G}'$ is optimized during joint training

\textbf{GloGNN} \citep{GloGNN}. GloGNN constructs its GSL bases using node embeddings derived from MLP, combining both low-pass and high-pass convolutional filters. A coefficient matrix $Z^{(l)}$ is used to characterize the relationship between nodes and is optimized to capture both feature and structural similarities $H_X^{(0)}=(1-\alpha)H_X^{(0)}+\alpha H_A^{(0)}$. Refinement is achieved via top-k selection based on the multi-hop adjacency matrix, and the matrix is symmetrized. The final graph is obtained through global aggregation of nodes, capturing both local and distant homophilous nodes. This graph is then used in downstream tasks, where the GSL graph $\mathcal{G}'$ is jointly optimized with the model parameters.

\textbf{HiGNN} \citep{HiGNN}. HiGNN constructs its GSL bases by utilizing heterophilous information as node neighbor distributions, which represent the likelihood of neighboring nodes belonging to different classes $\mathcal{H}_u=[p_1,p_2,...,p_c],where\ p_i=\frac{|{v|v\in\mathcal{N}_u,y_v=i}|}{|\mathcal{N}_u|}$. A new graph structure $\mathcal{G}'$ is constructed by linking nodes with similar heterophilous distributions using cosine similarity. The refinement involves selecting top-k edges based on the similarity score and applying symmetrization. The final graph is fused with the original adjacency matrix $A$ and the newly constructed adjacency matrix $A'$ via late fusion during message passing, where the node embeddings from both $A$ and $A'$ are combined with a balance parameter $\lambda$. The graph $\mathcal{G}'$ and node embeddings are updated during static training.

\newpage
\section{Proof of Theorem}\label{apd:proof}

\begin{theorem} 1
Given a graph $\mathcal{G} = \{\mathcal{V}, \mathcal{E}\}$ with node labels $\mathbf{Y}$ and node features $\mathbf{X}$, the accuracy of graph convolution in node classification is upper bounded by the mutual information between the node label $Y$ and the aggregated node features $H$:
\begin{equation}
    P_A \le \frac{I(Y; H) + \log 2}{\log(C)}
\end{equation}
\end{theorem}

\textit{Proof.} For an arbitrary node $u$, the aggregated node features can be derived as $H_u = \frac{1}{|\mathcal{N}_u|} \sum_{v \in \mathcal{N}_u} X_v$ following the graph convolution operation. For a classifier predicting labels based on $H_u$, we have $\hat{Y}_u = \text{cls}(H_u)$. Consequently, the Markov chain $Y \to H \to \hat{Y}$ holds. By applying Fano's inequality \citep{}, we obtain
\begin{equation}
    H(Y | H) \le H_b(P_E) + P_E \log(C-1)
\end{equation}
where $P_E$ represents the error rate and $H_b(\cdot)$ is the binary entropy function. Rearranging this inequality gives us a lower bound on $P_E$:
\begin{equation}
    P_E \ge \frac{H(Y | H) - H_b(P_E)}{\log(C-1)}
\end{equation}

Since $H(Y | H) = H(Y) - I(Y; H) = \log(C) - I(Y; H)$ and $H_b(P_E) \le \log 2$, we can substitute these terms into the equation:
\begin{equation}
    P_E \ge 1 - \frac{I(Y; H) + \log 2}{\log(C)}
\end{equation}

Finally, by expressing the accuracy rate $P_A$, we find:
\begin{equation}
    P_A = 1 - P_E \le \frac{I(Y; H) + \log 2}{\log(C)}
\end{equation}
This concludes the proof of Theorem 1.

%%%%%%%%%%%%%%%%%% B_u vs B for variable ?
\begin{theorem} 2
    Given a new graph $\mathcal{G}'=\{\mathcal{V},\mathcal{E}'\}$ constructed by a non-parametric graph structure learning method on bases $B$, the mutual information $I(Y;B')$ between node label $Y$ and aggregated bases $B'_u = \frac{1}{\abs{\mathcal{N}_u}}\sum_{v\in\mathcal{N}_u}B_v$ is upper bounded by $I(Y;B)$.
\end{theorem}

\textit{Proof.} For node in class $k$, we can get its GSL bases $B_k$ using node features $X_k$ and topological information $\mathcal{N}_k$ (if possible), which follows a class-wised distribution $B_u \sim p_B(y)$. Then, for a non-parametric GSL method, we have the probability that class $k$ connects with class $j$ as:
\begin{equation}
    p_{k,j} = \frac{g(B_k,B_j)}{\sum_{i\in\mathcal{N}_k} g(B_k,B_i)}
\end{equation}
where $g(\cdot)$ is a non-parametric measurement of the probability of new connections, such as cosine similarity or Minkowski Distance. Then, we can get an aggregated bases from the new graph by the operation of graph convolution \citep{is_hom_necessary,when_do_graph_help}:
\begin{equation}
    B_k' = \sum_{j=1}^C p_{k,j} B_j
\end{equation}
Therefore, the Markow chain $Y\to B\to B'$ holds. From data processing inequality \citep{proof_DPI}, we have
\begin{equation}
    I(Y;B')\le I(Y,B)
\end{equation}
This concludes the proof of Theorem 2.

\newpage
\section{Dataset Details}\label{apd:dataset_details}

The datasets used in our experiments include heterophilous graphs: Squirrel, Chameleon, Actor, Texas, Cornell, and Wisconsin \citep{Geom-GCN, dataset_hetero}, homophilous graphs: Cora, PubMed, and Citeseer \citep{dataset_cora}, and Minesweeper, Roman-empire, Amazon-ratings, Tolokers, and Questions \citep{hom_gnn_progress}. The dataset statistics are shown in \ref{tab:dataset_statistics}. The descriptions of all the datasets are given below:

\begin{table}[htbp]
\centering
\begin{tabular}{*{13}{c}}
\toprule
\textbf{Dataset} & \textbf{\#Nodes} & \textbf{\#Edges}& \textbf{\#Classes} & \textbf{\#Features} &\textbf{Edge Homophily} \\
\midrule
Cora  & 2,708  & 5,278 & 7  & 1,433 & 0.81  \\
Pubmed & 19,717 & 44,324 & 3 & 500 & 0.80 \\
Citeseer & 3,327 & 4,552 & 6 & 3,703 & 0.74  \\
\midrule
Roman-empire & 22,662 & 32,927 & 18 & 300 & 0.05  \\
Amazon-ratings & 24,492  & 93,050  & 5 & 300 & 0.38    \\
Minesweeper & 10,000 & 39,402 & 2 & 7  & 0.68 \\
Tolokers & 11,758 & 529,000 & 2 & 10   & 0.59 \\
Questions & 48,921 & 153,540 & 2 & 301 & 0.84 \\
\midrule
Cornell & 183 & 295 & 5 & 1,703  & 0.30   \\
Chameleon & 2,277 & 36,101 & 5  & 2,325 & 0.23 \\
Wisconsin & 251 & 466 & 5 & 1,703  & 0.21 \\
Texas & 183 & 309 & 5 & 1,703  & 0.11 \\
Squirrel & 5,201 & 216,933 & 5 & 2,089  & 0.22  \\
Actor & 7,600   & 33,544   & 5   & 931  & 0.22  \\
\bottomrule
\end{tabular}
\caption{Dataset Statistics}
\label{tab:dataset_statistics}
\end{table}

% \begin{table}[h]
% \centering
% \caption{Details of all homophilous and heterophilous datasets used in our experiments}
% \begin{tabular}{lccccc}
% \toprule
% Method         & Nodes  & Edges  & Classes & Features & Edge Homophily \\
% \midrule
% Cora           & 2,708   & 5,278   & 7       & 1,433     & 0.81           \\
% Pubmed         & 19,717  & 44,324  & 3       & 500      & 0.80           \\
% Citeseer       & 3,327   & 4,552   & 6       & 3,703     & 0.74           \\
% \midrule
% Roman-empire   & 22,662  & 32,927  & 18      & 300      & 0.05           \\
% Amazon-ratings & 24,492  & 93,050  & 5       & 300      & 0.38           \\
% Minesweeper    & 10,000  & 39,402  & 2       & 7        & 0.68           \\
% Tolokers       & 11,758  & 529,000 & 2       & 10       & 0.59           \\
% Questions      & 48,921  & 153,540 & 2       & 301      & 0.84           \\
% Cornell        & 183    & 295    & 5       & 1,703     & 0.30           \\
% Chameleon      & 2,277   & 36,101  & 5       & 2,325     & 0.23           \\
% Wisconsin      & 251    & 466    & 5       & 1,703     & 0.21           \\
% Texas          & 183    & 309    & 5       & 1,793     & 0.11           \\
% Squirrel       & 5,201   & 216,933 & 5       & 2,089     & 0.22           \\
% Actor          & 7,600   & 33,544  & 5       & 931      & 0.22           \\
% \bottomrule
% \end{tabular}
% \end{table}

\textbf{Cora}, \textbf{Citeseer}, and \textbf{Pubmed} datasets are widely used citation networks in graph structure learning research. In each dataset, nodes represent academic papers, while edges capture citation relationships between them. The node features are bag-of-words vectors derived from the paper's content, and each node is assigned a label based on its research topic. These datasets offer a structured framework to evaluate GNN models on classification tasks within citation networks.

\textbf{Roman-Empire} is constructed from the Roman Empire Wikipedia article, with nodes representing words and edges formed by either word adjacency or dependency relations. It contains 22.7K nodes and 32.9K edges. The task is to classify words by their syntactic roles, and node features are fastText embeddings. The graph is chain-like, with an average degree of 2.9 and a large diameter of 6824. Adjusted homophily is low ($h_{adj}$ = -0.05), making it useful for GNN evaluation under low homophily and sparse connectivity.

\textbf{Amazon-Ratings} is based on Amazon's product co-purchasing network, this dataset includes nodes as products (books, CDs, DVDs, etc.) and edges linking frequently co-purchased items. It consists of the largest connected component of the graph's 5-core. The goal is to predict product ratings grouped into five classes.

\textbf{Minesweeper} is a synthetic dataset resembling the Minesweeper game, nodes in a 100x100 grid represent cells, with edges connecting adjacent cells. The task is to identify mines (20\% of nodes). Node features indicate neighboring mine counts, with 50\% of features missing. The average degree is 7.88, and the graph has near-zero homophily due to random mine placement.

\textbf{Tolokers} is derived from the Toloka crowdsourcing platform, where nodes represent workers connected by shared tasks. The graph has 11.8K nodes and an average degree of 88.28. The task is to predict which workers have been banned, using profile and task performance features. The graph is much denser than others in the benchmark.

\textbf{Questions} is based on user interactions from Yandex Q, this dataset focuses on users interested in medicine. Nodes are users, and edges represent questions answered between users. It contains 48.9K nodes with an average degree of 6.28. The task is to predict user activity at the end of a one-year period, with fastText embeddings from user descriptions as features. The graph is highly imbalanced (97\% active users).

\textbf{Texas}, \textbf{Wisconsin}, \textbf{Cornell} are part of the WebKB project, representing web pages from university computer science departments. Nodes correspond to web pages, and edges represent hyperlinks between them. The node features are bag-of-words vectors from the web page content, and the labels classify each page into one of five categories: student, project, course, staff, and faculty.

\textbf{Chameleon}, \textbf{Squirrel} are page-page networks based on specific topics from Wikipedia. Nodes represent web pages, and edges correspond to mutual links between them. Node features are derived from the page content, and the classification task is based on average monthly traffic. These datasets are characterized by high heterophily, making them challenging for traditional GNN models.

\textbf{Actor} is an induced subgraph from a film-director-actor-writer network. Nodes represent actors, and edges are created when two actors co-occur on the same Wikipedia page. The task is to classify actors into five categories based on the keywords associated with their Wikipedia pages.

\newpage
\section{Implementation Details}\label{apd:implement_detail}

All the experiments are conducted on a linux server(Operation system: Ubuntu 16.04.7 LTS) with one NVIDIA Tesla V100 card.

\subsection{GNN+GSL}\label{apd:gnn_plus_gsl}
We implement GSL on $4$ baseline GNNs with a variety of GSL approaches from the perspective of GSL bases, GSL graph construction, and view fusion. The baseline GNNs include:

\begin{itemize}
\item \textbf{GCN} \citep{GCN} performs layer-wise propagation of node features and aggregates information from neighboring nodes to capture local graph structures. Each layer applies a convolution operation to update node embeddings, combining the node’s features with its neighbors.
\item \textbf{GAT} \citep{GAT} employs self-attention to learn dynamic attention coefficients between nodes and their neighbors. These coefficients are normalized using softmax, and the final node representation is computed as a weighted sum of the neighbor features. Multi-head attention is used to enhance stability and expressiveness, with the number of attention heads set to $8$ by default in our experiments.
\item \textbf{SAGE} \citep{GraphSage} uses an inductive framework to aggregate features from a node’s local neighborhood, allowing it to generalize to unseen nodes. The aggregation function, set to mean in our experiments, efficiently combines neighbor information at each layer.
\item \textbf{SGC} \citep{SGC} simplifies the GCN model by removing non-linear activations and collapsing multiple layers into a single linear transformation. This reduction in complexity accelerates training. Node features are propagated using precomputed matrices, making the model faster and more efficient. In our experiments, the number of k-hops in SGC is set to 2 by default.
\end{itemize}

The GSL bases $\b{B}$ includes the following options:
\begin{itemize} 
    \item $\b{B} = \b{X}$: The original node features are used as the GSL bases. 
    \item $\b{B} = \b{\hat{A}X}$: Aggregated node features from 1-hop neighbors, normalized by node degree, are used as the GSL bases. 
    \item $\b{B} = \text{MLP}(\b{X})$: Pretrained MLP embeddings are used as the GSL bases. A 2-layer MLP is trained using node features and labels on the training set for 1000 epochs per run. The hidden layer size is set to 128, the learning rate to $1e^{-2}$, the dropout rate to 0.5, and the weight decay to $5e^{-4}$. All parameters are optimized with Adam. After training, node embeddings are extracted from the last hidden layer, with a dimension of 128, prior to classifier input. 
    \item $\b{B} = \text{GCN}(\b{X}, \b{A})$: Pretrained node embeddings are obtained from a 2-layer GCN model, following the same training procedure as for the MLP embeddings. 
    \item $\b{B} = \text{GCL}(\b{X}, \b{A})$: Pretrained node embeddings are derived from a Graph Contrastive Learning (GCL) model without supervision, following the same training process as the MLP embeddings. GRACE \citep{GRACE} is used as the GCL model, with 2 views and 2 layers. The edge and feature dropout rates in each view are set to 0.2. 
\end{itemize}

The approaches for the construction of GSL graph $\mathcal{G}'$ includes:
\begin{itemize} 
    \item Cos-graph: $\mathcal{G}'=\{e_{ij}|\text{cos}(\b{B_i}, \b{B_j}) > \delta, i \in \mathcal{V}, j \in \mathcal{V}\}$. This method calculates the cosine similarity between all node pairs in the original graph $\mathcal{G}$. Node pairs with a similarity higher than the threshold $\delta$ are selected as the edge set for the GSL graph $\mathcal{G}'$. 
    \item Cos-node: $\mathcal{G}'=\bigcup_{i \in \mathcal{V}}\{ \{e_ij\}|\text{cos}(\b{B_i}, \b{B_j}) > \delta_i, j \in \mathcal{N}_i\}$. Unlike Cos-graph, which operates at the graph level, Cos-node constructs $\mathcal{G}'$ at the node level. To prevent nodes from being left without neighbors (which may occur in Cos-graph), Cos-node selects neighbors based on node-level cosine similarity, ensuring each node has sufficient connections. 
    \item kNN: $\mathcal{G}' = \text{kNN}(\b{B})$. This method constructs a kNN graph using the k-Nearest Neighbors algorithm based on the GSL bases $\b{B}$. 
\end{itemize}

The view fusion in GSL includes:
\begin{itemize}
    \item $\{\mathcal{G}'\}$: This approach uses only the GSL graph $\mathcal{G}'$ for subsequent GNN training, completely ignoring the original graph $\mathcal{G}$.
    \item $\{\mathcal{G},\mathcal{G}'\}, \theta_1=\theta_2$. Both the GSL graph $\mathcal{G}'$ and the original graph $\mathcal{G}$ are used for GNN training, with parameter sharing across each layer of the GNN.
    \item $\{\mathcal{G},\mathcal{G}'\}, \theta_1\neq\theta_2$. Both the GSL graph $\mathcal{G}'$ and the original graph $\mathcal{G}$ are used for GNN training, but with separate model parameters for each graph.
\end{itemize}

Especially, for graphs with two views, the fusion stage in GSL includes:
\begin{itemize}
    \item Early Fusion: $\mathcal{G}+\mathcal{G}'$.Combine the two graphs, $\mathcal{G}$ and $\mathcal{G}'$, into a single new graph prior to GNN training.
    \item Late Fusion: $\b{H}+\b{H}'$. After training the GNN on the original graph $\mathcal{G}$ and the GSL graph $\mathcal{G}'$, merge the node embeddings, $\mathbf{H}$ and $\mathbf{H}'$, before passing them to the classifiers.
\end{itemize}

In addition to the original models based on $4$ baseline GNNs, we implement GNN+GSL (GSL-augmented GNNs) by combining the aforementioned GSL modules, resulting in multiple variants for each type of GNN. For all models, we explore hyperparameters including hidden dimensions from the set $\{64, 128, 256\}$, learning rates from \{1\text{e}{-2}, 1\text{e}{-3}, 1\text{e}{-4}\}, weight decay values from \{0, 1\text{e}{-5}, 1\text{e}{-3}\}, the number of layers from $\{2, 3\}$, and dropout rates from $\{0.2, 0.4, 0.6, 0.8\}$.

For GSL graph generation, we also search for additional hyperparameters to ensure the performance quality of the GSL-augmented GNN. Specifically, for Cos-graph and Cos-node, we control the parameter $\delta$ to vary the ratio of the number of edges in $\mathcal{G}'$ to the number of edges in $\mathcal{G}$ across the set $\{0.1, 0.5, 1, 5\}$. For kNN, we investigate the number of neighbors from the set $\{2, 3, 5, 10\}$..

\subsection{SOTA-GSL}
To fairly re-evaluate the effectiveness of GSL in state-of-the-art (SOTA) models, two methods are employed to compare performance within the same search space. The first method (SOTA, $\mathcal{G}'=\mathcal{G}$) replaces the GSL graph with the original graph. The second method (SOTA, $\mathcal{G}'=\text{MLP}$) substitutes the GSL graph with a linear transformation, connecting it to the subsequent model structures and ensuring the continuity of channels within the original network structure. We train each model for 1000 epochs and search the hidden dimensions from the set \{16, 32, 64, 128, 256, 512\}, learning rate from \{1\text{e}{-1}, 1\text{e}{-2}, 1\text{e}{-3}, 1\text{e}{-4}, 1\text{e}{-5}\}, weight decay values from \{5\text{e}{-4}, 5\text{e}{-5}, 5\text{e}{-6}, 5\text{e}{-7}, 0\}, the number of layers from \{1, 2, 3\}, and dropout rates from \{0.2, 0.4, 0.6, 0.8\}. The model-specific hyperparameters are shown as follows:

In \textbf{GRCN}, the hyperparameter K determines the number of nearest neighbors used to create a sparse graph from a dense similarity graph which helps balance efficiency and accuracy.We set the k as 5.

In \textbf{GAug}, the alpha is a hyperparameter that regulates the influence of the edge predictor on the original graph. We set the alpha as 0.1.

In \textbf{IDGL}, The parameter graph\_learn\_num\_pers defines the number of perspectives for evaluating node similarities in the graph learning process. The parameter num\_anchors specifies the number of anchor points used to reduce computational complexity and improve scalability in graph structure learning. The graph\_skip\_conn parameter controls the proportion of skip connections, preserving information from the original graph during new graph structure learning. The update\_adj\_ratio parameter determines the proportion of the adjacency matrix updated at each iteration, influencing the dynamic adjustment of the graph structure. We set the graph\_learn\_num\_pers as 6, num\_anchors as 500, graph\_skip\_conn as 0.7, and update\_adj\_ratio as 0.3.

In \textbf{NodeFormer}, The parameter k determines the number of neighbors considered for each node in constructing the local graph structure, influencing the strength of node connections and the propagation of features. The parameter tolerance controls the degree of error tolerance during optimization. A larger tolerance allows more flexibility in the search space near local optima, while a smaller one results in stricter convergence. The number of attention heads in a graph attention network (GAT). Multi-head attention enables the model to focus on different subspace representations simultaneously, enhancing the diversity and stability of the representations. We set the k as 10, lambda as 0.01, and n\_heads as 4.

In \textbf{GEN},the parameter K in KNN refers to the number of nearest neighbors used to construct the graph structure, determining how many adjacent nodes are selected. The parameter tolerance defines the acceptable range of error during optimization, controlling the convergence criteria of the model. The parameter threshold determines the edge weight threshold in the graph, deciding which edges to retain in the graph structure.We set the k as 10, tolerance as 0.01, and threshold as 0.5.

In \textbf{GloGNN}, we set the Delta as 0.9, Gamma as 0.8, alpha as 0.5, beta as 2000, and orders as 5. Delta adjusts the balance between local and global node embeddings. Gamma controls the significance of global aggregation versus local information. Alpha balances the contributions of node features and graph structure. Beta regularizes the model, preventing overfitting. Order defines how many layers of neighbors are considered.

In \textbf{WRGAT}, we set the number of attention heads as 2 and the negative slope as 0.2. The number of attention heads determines how many attention mechanisms are used. The negative slope modifies the LeakyReLU activation.

% the number of layers from $\{1, 2, 3\}$, the dimensions of hidden embeddings from $\{16, 32, 64, 128, 256, 512\}$, learning rates from $\{1\times 10^{-5}, 1\times 10^{-4}, 1\times 10^{-3},  5\times 10^{-3},1\times 10^{-2}, 5\times 10^{-2},1\times 10^{-1}\}$, weight decay from $\{0, 5\times 10^{-7}, 5\times 10^{-6}, 5\times 10^{-5}, 5\times 10^{-4}, 5\times 10^{-3}\}$, and dropout rate from $\{0, 0.2, 0.4, 0.6, 0.8\}$ for all the models. 

The tables below show the best combination of hyperparameters based on the accuracy of test set.
%C:datasets
%baseline\    training
% 5  3

\begin{table}[htbp]
\resizebox{1\hsize}{!}{
\centering
\begin{tabular}{llccccc}
\toprule
Dataset & Model & Learning Rate & Weight Decay & Dropout & Hidden Dim & Num of Layers \\
\midrule
\multirow{24}{*}{Cora}    & GAug                   & 1e-4                        & 5e-7                  & 0.8              & 512                   & 2                         \\
                          & GAug,~$\mathcal{G}'=\mathcal{G}$       & 1e-4                        & 5e-7                  & 0.8              & 512                   & 2                         \\
                          & GAug,~$\mathcal{G}'=\text{MLP}$          & 1e-4                        & 5e-7                  & 0.8              & 512                   & 2                         \\
                          & GEN                    & 1e-2                        & 5e-4                  & 0.5              & 16                    & 2                         \\
                          & GEN,~$\mathcal{G}'=\mathcal{G}$        & 1e-2                        & 5e-4                  & 0.5              & 16                    & 2                         \\
                          & GEN,~$\mathcal{G}'=\text{MLP}$           & 1e-2                        & 5e-4                  & 0.5              & 16                    & 2                         \\
                          & GRCN                   & 1e-3                        & 5e-3                  & 0.5              & 256                   & 2                         \\
                          & GRCN,~$\mathcal{G}'=\mathcal{G}$       & 1e-3                        & 5e-3                  & 0.5              & 256                   & 2                         \\
                          & GRCN,~$\mathcal{G}'=\text{MLP}$          & 1e-3                        & 5e-3                  & 0.5              & 256                   & 2                         \\
                          & IDGL                   & 1e-2                        & 5e-4                  & 0.5              & 512                   & 2                         \\
                          & IDGL,~$\mathcal{G}'=\mathcal{G}$       & 1e-2                        & 5e-4                  & 0.5              & 512                   & 2                         \\
                          & IDGL,~$\mathcal{G}'=\text{MLP}$          & 1e-2                        & 5e-4                  & 0.5              & 512                   & 2                         \\
                          & NodeFormer             & 1e-2                        & 5e-4                  & 0.2              & 64                    & 2                         \\
                          & NodeFormer,~$\mathcal{G}'=\mathcal{G}$ & 1e-2                        & 5e-4                  & 0.2              & 64                    & 2                         \\
                          & NodeFormer,~$\mathcal{G}'=\text{MLP}$    & 1e-2                        & 5e-4                  & 0.2              & 64                    & 2                         \\
                          & GloGNN                 & 1e-2                        & 5e-5                  & 0.5              & 64                   & 1                         \\
                          & GloGNN,~$\mathcal{G}'=\mathcal{G}$     & 1e-2                        & 5e-5                  & 0.5              & 64                   & 1                         \\
                          & GloGNN,~$\mathcal{G}'=\text{MLP}$        & 1e-2                        & 5e-5                  & 0.5              & 64                   & 1                         \\
                          & WRGAT                  & 1e-2                        & 1e-5                  & 0.5              & 128                   & 2                         \\
                          & WRGAT,~$\mathcal{G}'=\mathcal{G}$      & 1e-2                        & 5e-5                  & 0.5              & 128                   & 2                         \\
                          & WRGAT,~$\mathcal{G}'=\text{MLP}$         & 1e-2                        & 1e-5                  & 0.5              & 128                   & 2                         \\
                          & WRGCN                  & 1e-2                        & 1e-5                  & 0.5              & 128                   & 2                         \\
                          & WRGCN,~$\mathcal{G}'=\mathcal{G}$      & 1e-2                        & 5e-5                  & 0.5              & 128                   & 2                         \\
                          & WRGCN,~$\mathcal{G}'=\text{MLP}$         & 1e-2                        & 1e-5                  & 0.5              & 128                   & 2                         \\
\bottomrule
\end{tabular}
}
\caption{Hyperparameters for SOTA-GSL on Cora.}
\end{table}
\begin{table}[htbp]
\resizebox{1\hsize}{!}{
\centering
\begin{tabular}{llccccc}
\toprule
Dataset & Model & Learning Rate & Weight Decay & Dropout & Hidden Dim & Num of Layers \\
\midrule
\multirow{24}{*}{PubMed} & GAug                   & 1e-2                        & 5e-4                  & 0.5              & 128                   & 2                         \\
                          & GAug,~$\mathcal{G}'=\mathcal{G}$       & 1e-2                        & 5e-4                  & 0.5              & 128                   & 2                         \\
                          & GAug,~$\mathcal{G}'=\text{MLP}$          & 1e-2                        & 5e-4                  & 0.5              & 128                   & 2                         \\
                          & GEN                    & 1e-3                        & 5e-4                  & 0.2              & 32                    & 2                         \\
                          & GEN,~$\mathcal{G}'=\mathcal{G}$        & 1e-3                        & 5e-4                  & 0.2              & 32                    & 2                         \\
                          & GEN,~$\mathcal{G}'=\text{MLP}$           & 1e-3                        & 5e-4                  & 0.2              & 32                    & 2                         \\
                          & GRCN                   & 1e-3                        & 5e-3                  & 0.5              & 32                   & 2                         \\
                          & GRCN,~$\mathcal{G}'=\mathcal{G}$       & 1e-3                        & 5e-3                  & 0.5              & 32                   & 2                         \\
                          & GRCN,~$\mathcal{G}'=\text{MLP}$          & 1e-3                        & 5e-3                  & 0.5              & 32                    & 2                         \\
                          & IDGL                   & 1e-2                        & 5e-4                  & 0.5              & 16                   & 2                         \\
                          & IDGL,~$\mathcal{G}'=\mathcal{G}$       & 1e-2                        & 5e-4                  & 0.5              & 16                   & 2                         \\
                          & IDGL,~$\mathcal{G}'=\text{MLP}$          & 1e-2                        & 5e-4                  & 0.5              & 16                    & 2                         \\
                          & NodeFormer             & 1e-3                        & 5e-4                  & 0.2              & 64                    & 2                         \\
                          & NodeFormer,~$\mathcal{G}'=\mathcal{G}$  & 1e-3                      & 5e-4                  & 0.2              & 64                    & 2                         \\
                          & NodeFormer,~$\mathcal{G}'=\text{MLP}$    & 1e-3                        & 5e-4                  & 0.2              & 32                    & 2                         \\
                          & GloGNN                 & 1e-3                        & 5e-5                  & 0.7              & 64                   & 3                         \\
                          & GloGNN,~$\mathcal{G}'=\mathcal{G}$     & 1e-3                        & 5e-5                  & 0.7              & 64                   & 3                         \\
                          & GloGNN,~$\mathcal{G}'=\text{MLP}$        & 1e-3                        & 5e-5                  & 0.7              & 64                   & 3                         \\
                          & WRGAT                  & 1e-2                        & 5e-5                  & 0.5              & 64                   & 2                         \\
                          & WRGAT,~$\mathcal{G}'=\mathcal{G}$      & 1e-2                        & 1e-5                  & 0.5              & 64                   & 2                         \\
                          & WRGAT,~$\mathcal{G}'=\text{MLP}$         & 1e-2                        & 5e-5                  & 0.5              & 64                   & 2                         \\
                          & WRGCN                  & 1e-2                       & 5e-5                  & 0.5              & 64                   & 2                         \\
                          & WRGCN,~$\mathcal{G}'=\mathcal{G}$      & 1e-2                        & 5e-5                  & 0.5              & 64                   & 2                         \\
                          & WRGCN,~$\mathcal{G}'=\text{MLP}$         & 1e-2                        & 5e-5                  & 0.5              & 64                   & 2                         \\
\bottomrule
\end{tabular}
}
\caption{Hyperparameters for SOTA-GSL on PubMed.}
\end{table}
\begin{table}[htbp]
\resizebox{1\hsize}{!}{
\centering
\begin{tabular}{llccccc}
\toprule
Dataset & Model & Learning Rate & Weight Decay & Dropout & Hidden Dim & Num of Layers \\
\midrule
\multirow{24}{*}{Citeseer} & GAug                   & 1e-4                        & 5e-7                  & 0.8              & 512                   & 2                         \\
                          & GAug,~$\mathcal{G}'=\mathcal{G}$       & 1e-4                        & 5e-7                  & 0.8              & 512                   & 2                         \\
                          & GAug,~$\mathcal{G}'=\text{MLP}$          & 1e-4                        & 5e-7                  & 0.8              & 512                   & 2                         \\
                          & GEN                    & 1e-2                        & 5e-4                  & 0.5              & 16                    & 2                         \\
                          & GEN,~$\mathcal{G}'=\mathcal{G}$        & 1e-2                        & 5e-4                  & 0.5              & 16                    & 2                         \\
                          & GEN,~$\mathcal{G}'=\text{MLP}$           & 1e-2                        & 5e-4                  & 0.5              & 16                    & 2                         \\
                          & GRCN                   & 1e-3                        & 5e-3                  & 0.8              & 512                   & 3                         \\
                          & GRCN,~$\mathcal{G}'=\mathcal{G}$       & 1e-3                        & 5e-3                  & 0.8              & 512                   & 3                         \\
                          & GRCN,~$\mathcal{G}'=\text{MLP}$          & 1e-2                        & 5e-3                  & 0.5              & 256                   & 3                         \\
                          & IDGL                   & 1e-2                        & 5e-4                  & 0.5              & 32                   & 2                         \\
                          & IDGL,~$\mathcal{G}'=\mathcal{G}$       & 1e-3                        & 5e-4                  & 0.5              & 16                   & 2                         \\
                          & IDGL,~$\mathcal{G}'=\text{MLP}$         & 1e-3                        & 5e-4                  & 0.5              & 16                   & 2                         \\
                          & NodeFormer             & 1e-2                        & 5e-4                  & 0.2              & 64                    & 2                         \\
                          & NodeFormer,~$\mathcal{G}'=\mathcal{G}$ & 1e-2                        & 5e-4                  & 0.2              & 64                    & 2                         \\
                          & NodeFormer,~$\mathcal{G}'=\text{MLP}$    & 1e-2                        & 5e-4                  & 0.2              & 64                    & 2                         \\
                          & GloGNN                 & 1e-2                        & 1e-5                  & 0.7              & 64                   & 2                         \\
                          & GloGNN,~$\mathcal{G}'=\mathcal{G}$     & 1e-2                        & 1e-5                  & 0.7              & 64                   & 2                         \\
                          & GloGNN,~$\mathcal{G}'=\text{MLP}$       & 1e-2                        & 1e-5                  & 0.7              & 64                   & 2                         \\
                          & WRGAT                  & 1e-2                        & 5e-5                  & 0.5              & 128                   & 2                         \\
                          & WRGAT,~$\mathcal{G}'=\mathcal{G}$      & 1e-2                        & 5e-5                  & 0.5              & 128                   & 2                         \\
                          & WRGAT,~$\mathcal{G}'=\text{MLP}$         & 1e-2                        & 5e-5                  & 0.5              & 128                   & 2                         \\
                          & WRGCN                  & 1e-2                        & 5e-5                  & 0.3              & 128                   & 2                         \\
                          & WRGCN,~$\mathcal{G}'=\mathcal{G}$      & 1e-2                        & 5e-5                  & 0.5              & 128                   & 2                         \\
                          & WRGCN,~$\mathcal{G}'=\text{MLP}$         & 1e-2                        & 1e-5                  & 0.5              & 128                   & 1                         \\
\bottomrule
\end{tabular}
}
\caption{Hyperparameters for SOTA-GSL on Citeseer.}
\end{table}
\begin{table}[htbp]
\resizebox{1\hsize}{!}{
\centering
\begin{tabular}{llccccc}
\toprule
Dataset & Model & Learning Rate & Weight Decay & Dropout & Hidden Dim & Num of Layers \\
\midrule
\multirow{24}{*}{Minesweeper}    & GAug                   & 1e-3 & 5e-6         & 0.8     & 256    & 3        \\
                                 & GAug,~$\mathcal{G}'=\mathcal{G}$       & 1e-3 & 5e-6         & 0.8     & 256    & 3        \\
                                 & GAug,~$\mathcal{G}'=\text{MLP}$          & 1e-3 & 5e-6         & 0.8     & 256    & 3        \\
                                 & GEN                    & 1e-4 & 5e-6         & 0.8     & 256    & 3        \\
                                 & GEN,~$\mathcal{G}'=\mathcal{G}$        & 1e-4 & 5e-6         & 0.8     & 256    & 3        \\
                                 & GEN,~$\mathcal{G}'=\text{MLP}$           & 1e-4 & 5e-6         & 0.8     & 256    & 3        \\
                                 & GRCN                   & 1e-3 & 5e-7         & 0.2     & 128    & 2        \\
                                 & GRCN,~$\mathcal{G}'=\mathcal{G}$       & 1e-3 & 5e-6         & 0.2     & 128    & 2        \\
                                 & GRCN,~$\mathcal{G}'=\text{MLP}$          & 1e-3 & 5e-6         & 0.2     & 128    & 2        \\
                                 & IDGL                   & 1e-1 & 5e-6         & 0.2     & 128    & 3        \\
                                 & IDGL,~$\mathcal{G}'=\mathcal{G}$       & 1e-1 & 5e-6         & 0.2     & 128    & 3        \\
                                 & IDGL,~$\mathcal{G}'=\text{MLP}$          & 1e-1 & 5e-6         & 0.2     & 128    & 3        \\
                                 & NodeFormer             & 1e-2 & 5e-4         & 0.8     & 32     & 2        \\
                                 & NodeFormer,~$\mathcal{G}'=\mathcal{G}$ & 1e-2 & 5e-4         & 0.8     & 32     & 2        \\
                                 & NodeFormer,~$\mathcal{G}'=\text{MLP}$    & 1e-2 & 5e-4         & 0.8     & 32     & 2        \\
                                 & GloGNN                 & 1e-2 & 5e-4         & 0.5     & 512    & 5        \\
                                 & GloGNN,~$\mathcal{G}'=\mathcal{G}$     & 1e-2 & 5e-4         & 0.5     & 512    & 5        \\
                                 & GloGNN,~$\mathcal{G}'=\text{MLP}$        & 1e-2 & 5e-4         & 0.5     & 512    & 5        \\
                                 & WRGAT                  & 1e-2 & 5e-5         & 0.5     & 128    & 2        \\
                                 & WRGAT,~$\mathcal{G}'=\mathcal{G}$      & 1e-2 & 5e-5         & 0.5     & 128    & 2        \\
                                 & WRGAT,~$\mathcal{G}'=\text{MLP}$         & 1e-2 & 5e-5         & 0.5     & 128    & 2        \\
                                 & WRGCN                  & 1e-2 & 5e-5         & 0.5     & 128    & 2        \\
                                 & WRGCN,~$\mathcal{G}'=\mathcal{G}$      & 1e-2 & 5e-5         & 0.5     & 128    & 2        \\
                                 & WRGCN,~$\mathcal{G}'=\text{MLP}$         & 1e-2 & 5e-5         & 0.5     & 128    & 2        \\
\bottomrule
\end{tabular}
}
\caption{Hyperparameters for SOTA-GSL on Minesweeper.}
\end{table}
\begin{table}[htbp]
\resizebox{1\hsize}{!}{
\centering
\begin{tabular}{llccccc}
\toprule
Dataset & Model & Learning Rate & Weight Decay & Dropout & Hidden Dim & Num of Layers \\
\midrule
\multirow{24}{*}{Roman-empire}   & GAug                   & 1e-1 & 5e-5         & 0.5     & 32     & 2        \\
                                 & GAug,~$\mathcal{G}'=\mathcal{G}$       & 1e-1 & 5e-5         & 0.5     & 32     & 2        \\
                                 & GAug,~$\mathcal{G}'=\text{MLP}$          & 1e-1 & 5e-5         & 0.5     & 32     & 2        \\
                                 & GEN                    & 1e-2 & 5e-7         & 0.2     & 128    & 2        \\
                                 & GEN,~$\mathcal{G}'=\mathcal{G}$        & 1e-2 & 5e-7         & 0.2     & 128    & 2        \\
                                 & GEN,~$\mathcal{G}'=\text{MLP}$           & 1e-2 & 5e-7         & 0.2     & 128    & 2        \\
                                 & GRCN                   & 1e-3 & 5e-5         & 0.5     & 128    & 2        \\
                                 & GRCN,~$\mathcal{G}'=\mathcal{G}$       & 1e-2 & 5e-5         & 0.5     & 128    & 2        \\
                                 & GRCN,~$\mathcal{G}'=\text{MLP}$        & 1e-2 & 5e-5         & 0.5     & 128    & 2        \\
                                 & IDGL                   & 1e-1 & 5e-5         & 0.5     & 128    & 2        \\
                                 & IDGL,~$\mathcal{G}'=\mathcal{G}$       & 1e-1 & 5e-5         & 0.5     & 128    & 2        \\
                                 & IDGL,~$\mathcal{G}'=\text{MLP}$          & 1e-1 & 5e-5         & 0.5     & 128    & 2        \\
                                 & NodeFormer             & 1e-3 & 5e-6         & 0.2     & 128    & 3        \\
                                 & NodeFormer,~$\mathcal{G}'=\mathcal{G}$ & 1e-3 & 5e-6         & 0.2     & 128    & 3        \\
                                 & NodeFormer,~$\mathcal{G}'=\text{MLP}$    & 1e-3 & 5e-5         & 0.8     & 128    & 3        \\
                                 & GloGNN                 & 1e-2 & 5e-5         & 0.7     & 128    & 3        \\
                                 & GloGNN,~$\mathcal{G}'=\mathcal{G}$     & 1e-2 & 5e-5         & 0.7     & 128    & 3        \\
                                 & GloGNN,~$\mathcal{G}'=\text{MLP}$        & 1e-2 & 5e-5         & 0.7     & 128    & 3        \\
                                 & WRGAT                  & 1e-2 & 5e-5         & 0.5     & 128    & 2        \\
                                 & WRGAT,~$\mathcal{G}'=\mathcal{G}$     & 1e-2 & 1e-5         & 0.5     & 128    & 2        \\
                                 & WRGAT,~$\mathcal{G}'=\text{MLP}$         & 1e-2 & 5e-5         & 0.5     & 128    & 2        \\
                                 & WRGCN                  & 1e-2 & 5e-5         & 0.5     & 128    & 2        \\
                                 & WRGCN,~$\mathcal{G}'=\mathcal{G}$      & 1e-2 & 5e-5         & 0.5     & 128    & 2        \\
                                 & WRGCN,~$\mathcal{G}'=\text{MLP}$         & 1e-2 & 5e-5         & 0.5     & 128    & 2   \\
\bottomrule
\end{tabular}
}
\caption{Hyperparameters for SOTA-GSL on Roman-Empire.}
\end{table}
\begin{table}[htbp]
\resizebox{1\hsize}{!}{
\centering
\begin{tabular}{llccccc}
\toprule
Dataset & Model & Learning Rate & Weight Decay & Dropout & Hidden Dim & Num of Layers \\
\midrule
\multirow{24}{*}{Amazon-ratings} & GAug                   & 1e-2 & 5e-7         & 0.2     & 128    & 2        \\
                                 & GAug,~$\mathcal{G}'=\mathcal{G}$       & 1e-2 & 5e-7         & 0.2     & 128    & 2        \\
                                 & GAug,~$\mathcal{G}'=\text{MLP}$         & 1e-2 & 5e-7         & 0.2     & 128    & 2        \\
                                 & GEN                    & 1e-2 & 5e-7         & 0.2     & 128    & 2        \\
                                 & GEN,~$\mathcal{G}'=\mathcal{G}$        & 1e-2 & 5e-7         & 0.2     & 128    & 2        \\
                                 & GEN,~$\mathcal{G}'=\text{MLP}$           & 1e-2 & 5e-7         & 0.2     & 128    & 2        \\
                                 & GRCN                   & 1e-3 & 5e-7         & 0.2     & 128    & 2        \\
                                 & GRCN,~$\mathcal{G}'=\mathcal{G}$       & 1e-2 & 5e-7         & 0.2     & 64     & 2        \\
                                 & GRCN,~$\mathcal{G}'=\text{MLP}$          & 1e-2 & 5e-7         & 0.2     & 128    & 2        \\
                                 & IDGL                   & 1e-2 & 5e-7         & 0.2     & 128    & 2        \\
                                 & IDGL,~$\mathcal{G}'=\mathcal{G}$       & 1e-2 & 5e-7         & 0.2     & 128    & 2        \\
                                 & IDGL,~$\mathcal{G}'=\text{MLP}$          & 1e-2 & 5e-7         & 0.2     & 128    & 2        \\
                                 & NodeFormer             & 1e-4 & 5e-5         & 0.5     & 128    & 3        \\
                                 & NodeFormer,~$\mathcal{G}'=\mathcal{G}$ & 1e-4 & 5e-5         & 0.5     & 64     & 3        \\
                                 & NodeFormer,~$\mathcal{G}'=\text{MLP}$    & 1e-4 & 5e-5         & 0.5     & 64     & 3        \\
                                 & GloGNN                 & 1e-2 & 5e-5         & 0.3     & 128    & 3        \\
                                 & GloGNN,~$\mathcal{G}'=\mathcal{G}$     & 1e-2 & 5e-5         & 0.3     & 128    & 3        \\
                                 & GloGNN,~$\mathcal{G}'=\text{MLP}$        & 1e-2 & 5e-5         & 0.3     & 128    & 3        \\
                                 & WRGAT                  & 1e-2 & 5e-5         & 0.3     & 128    & 2        \\
                                 & WRGAT,~$\mathcal{G}'=\mathcal{G}$      & 1e-2 & 1e-5         & 0.3     & 128    & 2        \\
                                 & WRGAT,~$\mathcal{G}'=\text{MLP}$         & 1e-2 & 1e-5         & 0.3     & 128    & 2        \\
                                 & WRGCN                  & 1e-2 & 5e-5         & 0.7     & 128    & 3        \\
                                 & WRGCN,~$\mathcal{G}'=\mathcal{G}$      & 1e-2 & 5e-5         & 0.7     & 128    & 3        \\
                                 & WRGCN,~$\mathcal{G}'=\text{MLP}$         & 1e-2 & 1e-5         & 0.7     & 128    & 3        \\
\bottomrule
\end{tabular}
}
\caption{Hyperparameters for SOTA-GSL on Amazon-ratings.}
\end{table}
\begin{table}[htbp]
\resizebox{1\hsize}{!}{
\centering
\begin{tabular}{llccccc}
\toprule
Dataset & Model & Learning Rate & Weight Decay & Dropout & Hidden Dim & Num of Layers \\
\midrule
\multirow{24}{*}{Questions}  & GAug                   & 1e-2 & 5e-4         & 0.5     & 64     & 3        \\  
                             & GAug,~$\mathcal{G}'=\mathcal{G}$       & 1e-2 & 5e-4         & 0.5     & 64     & 3        \\
                             & GAug,~$\mathcal{G}'=\text{MLP}$          & 1e-2 & 5e-4         & 0.5     & 64     & 3        \\  
                             & GEN                    & 1e-2 & 5e-7         & 0.2     & 256    & 2        \\  
                             & GEN,~$\mathcal{G}'=\mathcal{G}$        & 1e-2 & 5e-7         & 0.2     & 256    & 2        \\
                             & GEN,~$\mathcal{G}'=\text{MLP}$           & 1e-2 & 5e-7         & 0.2     & 256    & 2        \\ 
                             & GRCN                   & 1e-2 & 5e-6         & 0.5     & 64     & 2        \\
                             & GRCN,~$\mathcal{G}'=\mathcal{G}$       & 1e-2 & 5e-6         & 0.5     & 64     & 2        \\ 
                             & GRCN,~$\mathcal{G}'=\text{MLP}$          & 1e-2 & 5e-6         & 0.5     & 64     & 2        \\ 
                             & IDGL                   & 1e-2 & 5e-7         & 0.2     & 128    & 2        \\ 
                             & IDGL,~$\mathcal{G}'=\mathcal{G}$       & 1e-2 & 5e-7         & 0.2     & 128    & 2        \\  
                             & IDGL,~$\mathcal{G}'=\text{MLP}$          & 1e-2 & 5e-7         & 0.2     & 128    & 2        \\ 
                             & NodeFormer             & 1e-4 & 5e-3         & 0.5     & 128    & 3        \\  
                             & NodeFormer,~$\mathcal{G}'=\mathcal{G}$ & 1e-4 & 5e-3         & 0.5     & 64     & 3        \\  
                             & NodeFormer,~$\mathcal{G}'=\text{MLP}$    & 1e-4 & 5e-3         & 0.5     & 64     & 3        \\  
                             & GloGNN                 & 1e-2 & 5e-5         & 0.7     & 128    & 3        \\ 
                             & GloGNN,~$\mathcal{G}'=\mathcal{G}$     & 1e-2 & 5e-5         & 0.7     & 128    & 3        \\ 
                             & GloGNN,~$\mathcal{G}'=\text{MLP}$        & 1e-2 & 5e-5         & 0.7     & 128    & 3        \\ 
                             & WRGAT                  & 5e-3 & 5e-5         & 0.3     & 64     & 2        \\ 
                             & WRGAT,~$\mathcal{G}'=\mathcal{G}$      & 5e-3 & 1e-5         & 0.3     & 64     & 2        \\ 
                             & WRGAT,~$\mathcal{G}'=\text{MLP}$         & 5e-3 & 5e-5         & 0.3     & 64     & 2        \\ 
                             & WRGCN                  & 5e-3 & 5e-5         & 0.7     & 64     & 2        \\ 
                             & WRGCN,~$\mathcal{G}'=\mathcal{G}$      & 5e-3 & 5e-5         & 0.7     & 64     & 2        \\ 
                             & WRGCN,~$\mathcal{G}'=\text{MLP}$         & 5e-3 & 1e-5         & 0.7     & 64     & 2        \\
\bottomrule
\end{tabular}
}
\caption{Hyperparameters for SOTA-GSL on Questions.}
\end{table}
\begin{table}[htbp]
\resizebox{1\hsize}{!}{
\centering
\begin{tabular}{llccccc}
\toprule
Dataset & Model & Learning Rate & Weight Decay & Dropout & Hidden Dim & Num of Layers \\
\midrule
% \multirow{16}{*}{} & GCN & 1e-2 & 5e-5 & 3e-5 & 32 & 0.2 \\
\multirow{24}{*}{Tolokers}   & GAug                   & 1e-1 & 5e-5         & 0.5     & 32     & 2        \\ 
                             & GAug,~$\mathcal{G}'=\mathcal{G}$       & 1e-1 & 5e-5         & 0.5     & 32     & 2        \\  
                             & GAug,~$\mathcal{G}'=\text{MLP}$          & 1e-1 & 5e-5         & 0.5     & 32     & 2        \\  
                             & GEN                    & 1e-2 & 5e-5         & 0.2     & 128    & 2        \\  
                             & GEN,~$\mathcal{G}'=\mathcal{G}$        & 1e-2 & 5e-6         & 0.2     & 128    & 2        \\  
                             & GEN,~$\mathcal{G}'=\text{MLP}$           & 1e-2 & 5e-6         & 0.2     & 128    & 2        \\  
                             & GRCN                   & 1e-2 & 5e-5         & 0.5     & 32     & 2        \\  
                             & GRCN,~$\mathcal{G}'=\mathcal{G}$       & 1e-2 & 5e-6         & 0.5     & 32     & 2        \\  
                             & GRCN,~$\mathcal{G}'=\text{MLP}$          & 1e-1 & 5e-6         & 0.5     & 64     & 2        \\  
                             & IDGL                   & 1e-2 & 5e-4         & 0.5     & 64     & 2        \\ 
                             & IDGL,~$\mathcal{G}'=\mathcal{G}$       & 1e-2 & 5e-4         & 0.5     & 64     & 2        \\  
                             & IDGL,~$\mathcal{G}'=\text{MLP}$          & 1e-2 & 5e-4         & 0.5     & 64     & 2        \\ 
                             & NodeFormer             & 1e-2 & 5e-4         & 0.2     & 64     & 2        \\  
                             & NodeFormer,~$\mathcal{G}'=\mathcal{G}$ & 1e-2 & 5e-4         & 0.2     & 64     & 2        \\  
                             & NodeFormer,~$\mathcal{G}'=\text{MLP}$    & 1e-2 & 5e-4         & 0.2     & 64     & 2        \\  
                             & GloGNN                 & 1e-2 & 5e-5         & 0.3     & 128    & 3        \\ 
                             & GloGNN,~$\mathcal{G}'=\mathcal{G}$     & 1e-2 & 5e-5         & 0.3     & 128    & 3        \\ 
                             & GloGNN,~$\mathcal{G}'=\text{MLP}$        & 1e-2 & 5e-5         & 0.3     & 128    & 3        \\  
                             & WRGAT                  & 1e-2 & 5e-5         & 0.5     & 128    & 2        \\  
                             & WRGAT,~$\mathcal{G}'=\mathcal{G}$      & 1e-2 & 1e-5         & 0.5     & 128    & 2        \\  
                             & WRGAT,~$\mathcal{G}'=\text{MLP}$         & 1e-2 & 5e-5         & 0.5     & 128    & 2        \\ 
                             & WRGCN                  & 1e-2 & 5e-5         & 0.5     & 128    & 1        \\  
                             & WRGCN,~$\mathcal{G}'=\mathcal{G}$      & 1e-2 & 5e-5         & 0.5     & 128    & 2        \\ 
                             & WRGCN,~$\mathcal{G}'=\text{MLP}$         & 1e-2 & 5e-5         & 0.5     & 128    & 2        \\
\bottomrule
\end{tabular}
}
\caption{Hyperparameters for SOTA-GSL on Tolokers.}
\end{table}

\newpage
\section{Additional Experiment Results}\label{apd:additional_exp}

In this section, we examine the impact of different GSL modules on GNN models. The GSL modules include graph bases, GSL graph generation, view fusion methods, and fusion stages, with details provided in Appendix \ref{apd:gnn_plus_gsl}.

\subsection{GSL Bases}
\begin{figure}[b]
    \centering
  \subfloat{%
       \includegraphics[width=0.5\textwidth]{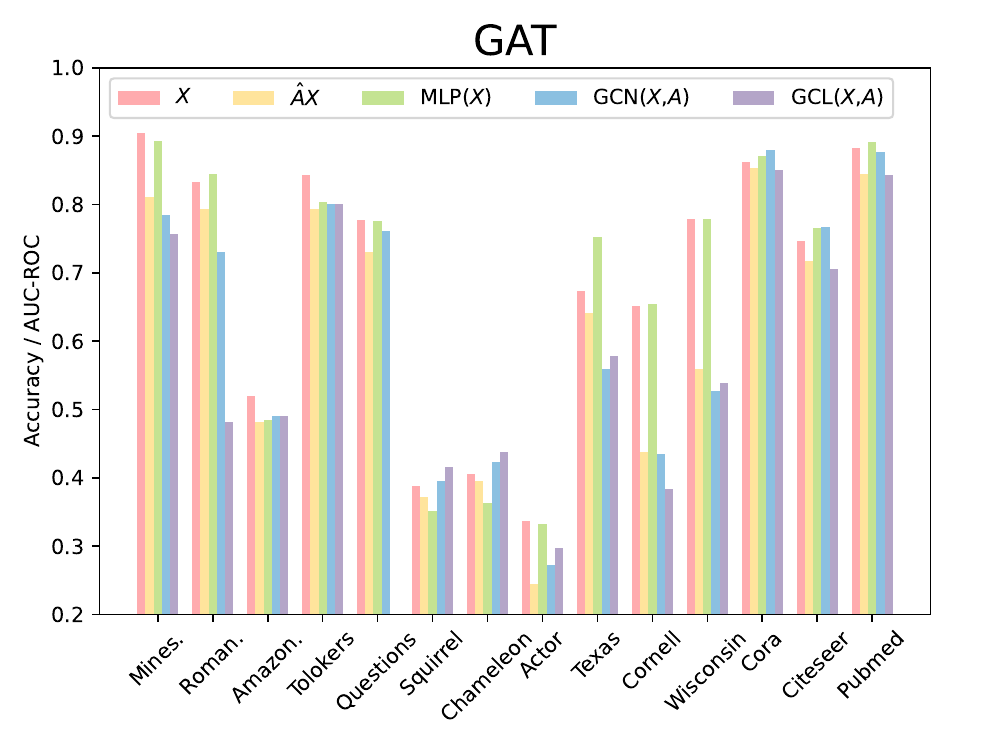}}
    \hfill
  \subfloat{%
        \includegraphics[width=0.5\textwidth]{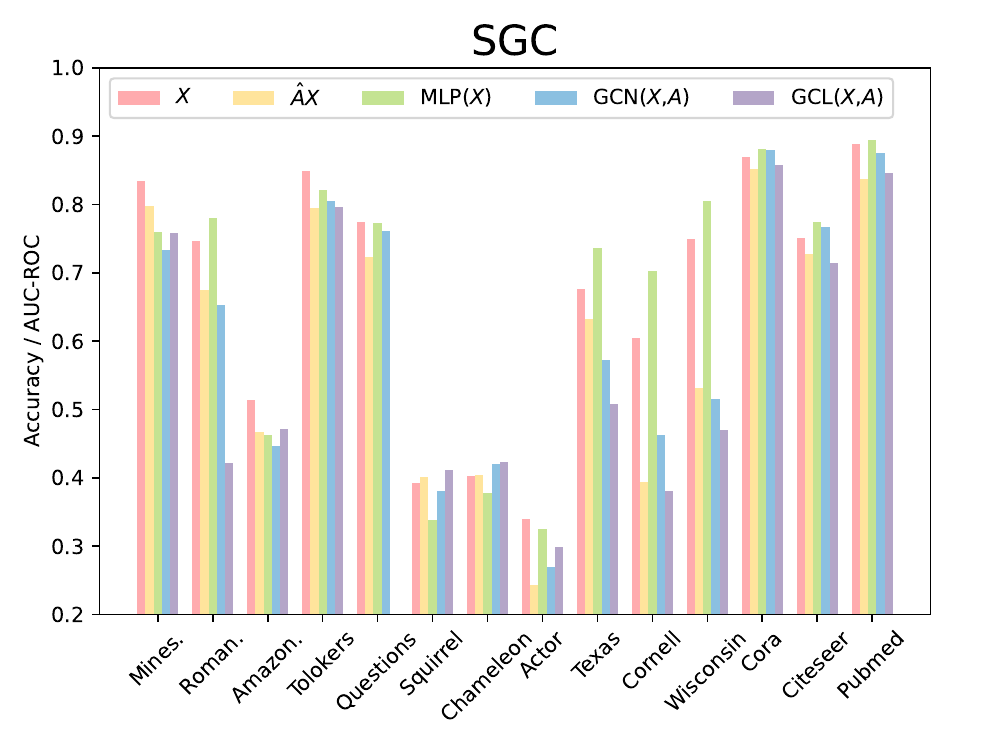}}
    \hfill
    \vspace{-0.7cm}
    \hspace*{\fill}
  \subfloat{%
        \includegraphics[width=0.5\textwidth]{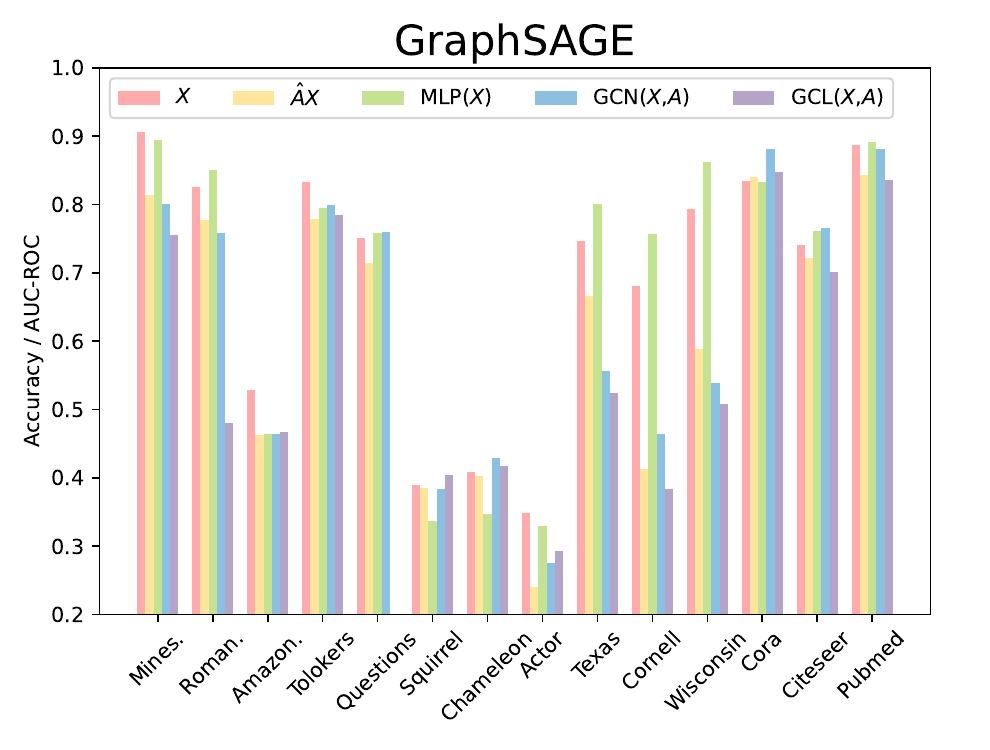}}
    % \hfill
    \hspace*{\fill}
    \vspace{-0.3cm}
    \caption{Influences of different GSL bases to more GNNs.}
    \label{fig:apd_gsl_basis_more}
\end{figure}

In addition to the analysis of the impact of GSL bases shown in Figure \ref{fig:gsl_basis}, Figure \ref{fig:apd_gsl_basis_more} presents further results on the performance of various GSL bases ($\mathbf{X}$, $\mathbf{\hat{A}}\mathbf{X}$, $MLP(\mathbf{X})$, $GCN(\mathbf{X}, \mathbf{A})$, $GCL(\mathbf{X}, \mathbf{A})$) across GAT, SGC, and GraphSAGE. The results are consistent with those observed in GCN and MLP, where the original node features do not always yield the best input. Some pretrained features, such as $MLP(\mathbf{X})$ on the Texas, Cornell, and Wisconsin datasets, demonstrate significant improvement compared to the original features $\mathbf{X}$, highlighting the necessity of self-training. Since many GSL methods \citep{HiGNN, WRGAT} utilize self-training during the training process, a fair comparison of these GSL methods and baseline GNNs should be conducted in the context of self-training, such as by using pretrained node features as input, as shown in Table \ref{tab:baseline_plus_rewriting}.

\subsection{GSL Graph Generation}
\begin{figure}[b]
    \centering
  \subfloat{%
       \includegraphics[width=0.5\textwidth]{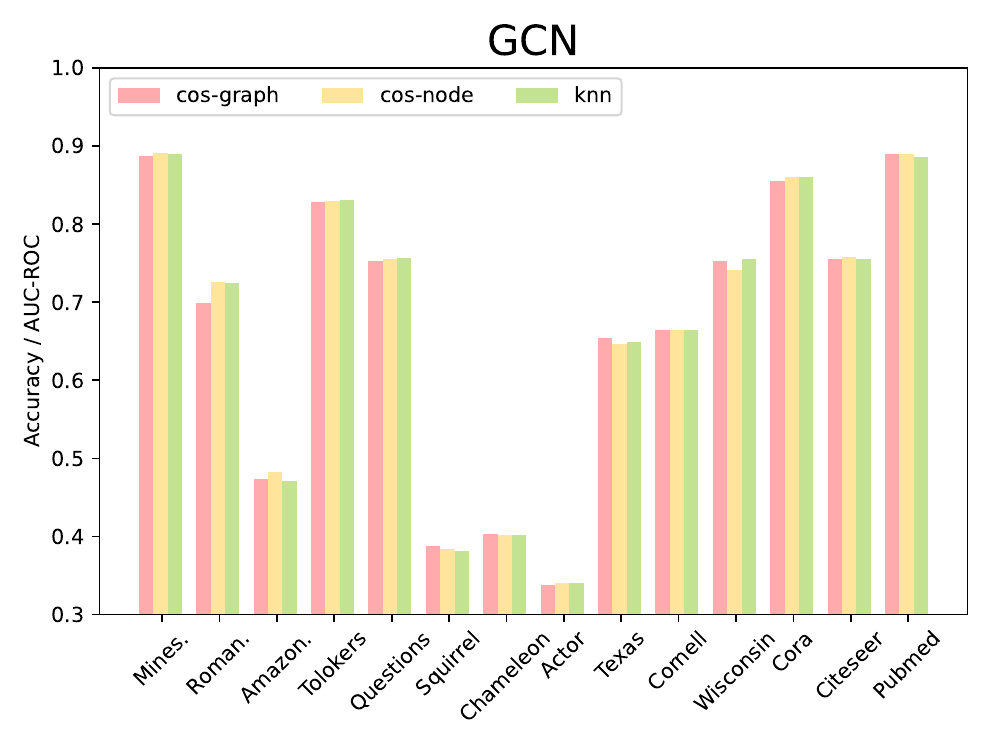}}
    \hfill
  \subfloat{%
        \includegraphics[width=0.5\textwidth]{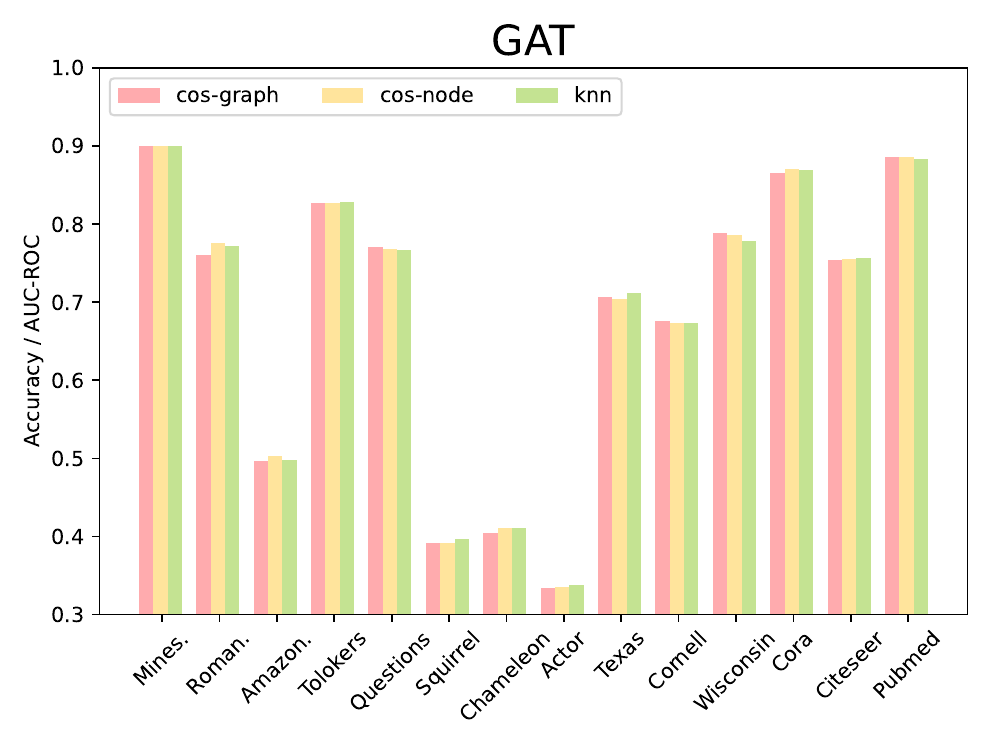}}
    \hfill
    \vspace{-0.7cm}
  \subfloat{%
       \includegraphics[width=0.5\textwidth]{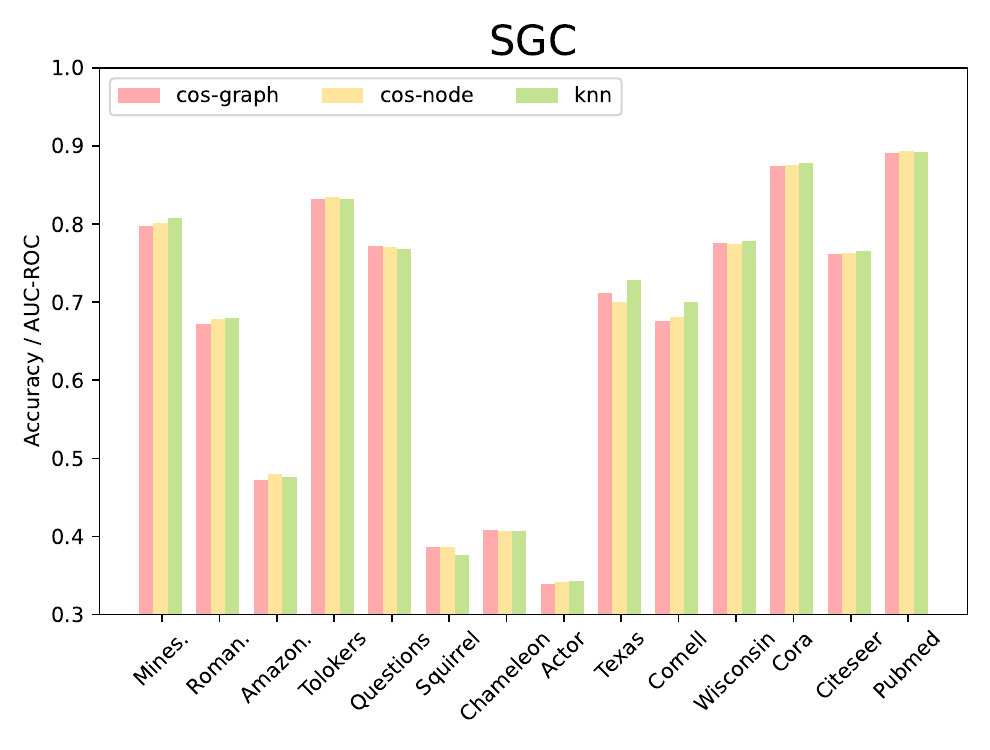}}
    \hfill
  \subfloat{%
        \includegraphics[width=0.5\textwidth]{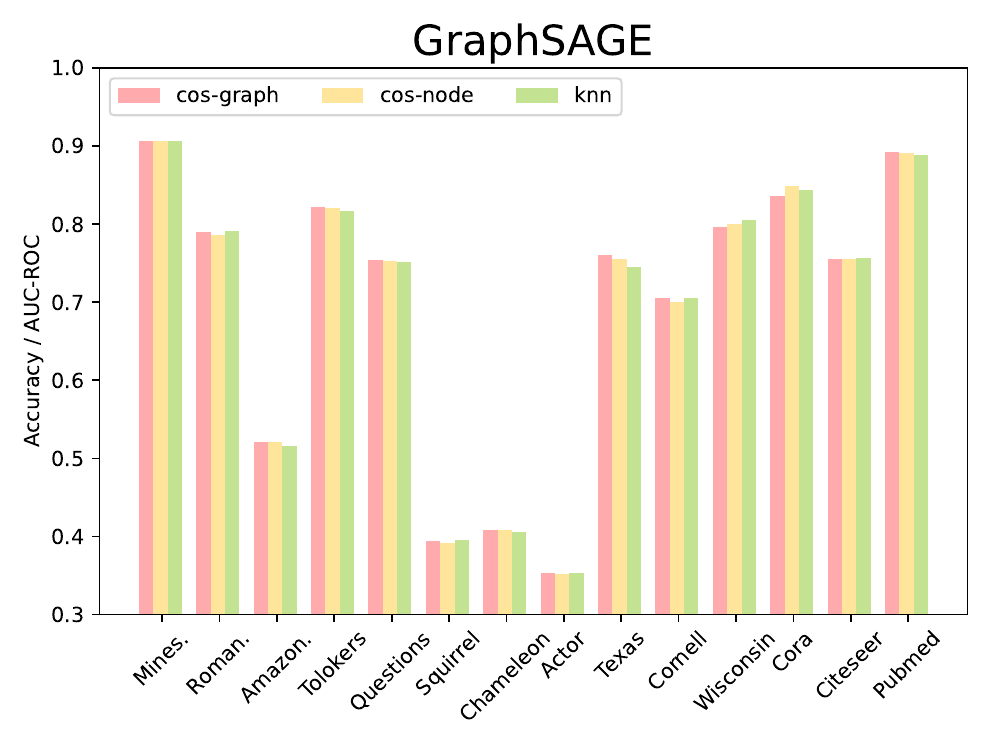}}
    \hfill
    % \hspace*{\fill}
    \vspace{-0.3cm}
    \caption{Influences of the approaches of GSL generation to GNN+GSL.}
    \label{fig:apd_basis_gsl_construct}
\end{figure}

Figure \ref{fig:apd_basis_gsl_construct} compares the Cos-graph, Cos-node, and kNN methods for GSL graph generation. Across most datasets, the performance differences among these methods are minimal. In certain datasets, such as Roman-empire and Pubmed, the models exhibit comparable performance regardless of the graph generation technique employed. This suggests that variations in graph generation have a limited effect on overall performance.

\subsection{View Fusion}
\begin{figure}[b]
    \centering
  \subfloat{%
       \includegraphics[width=0.5\textwidth]{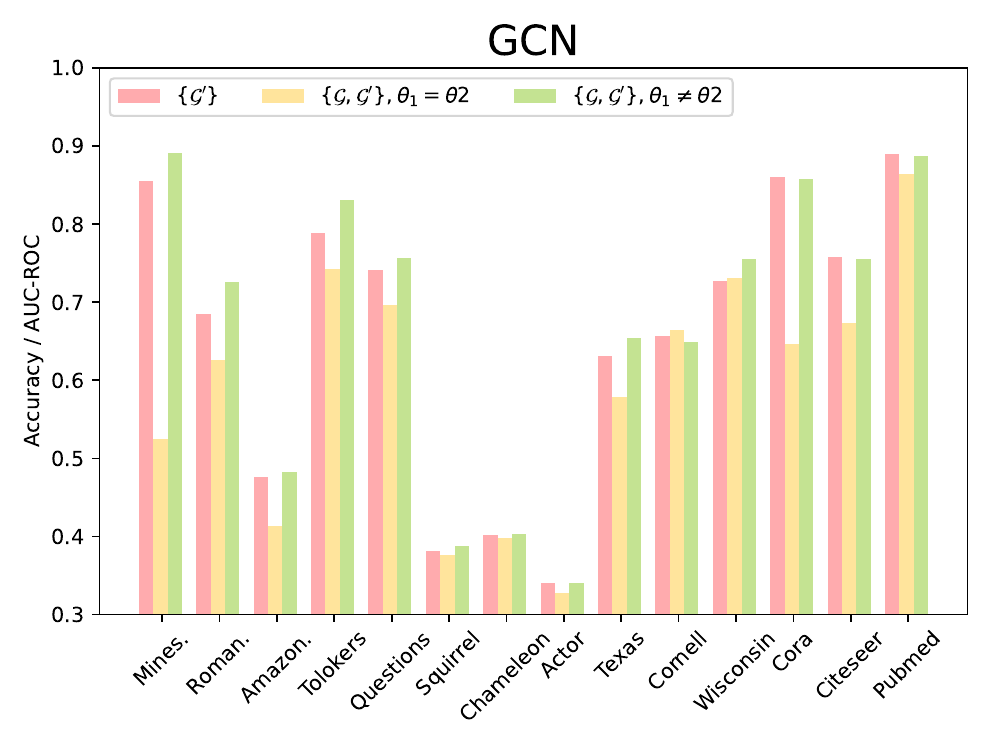}}
    \hfill
  \subfloat{%
        \includegraphics[width=0.5\textwidth]{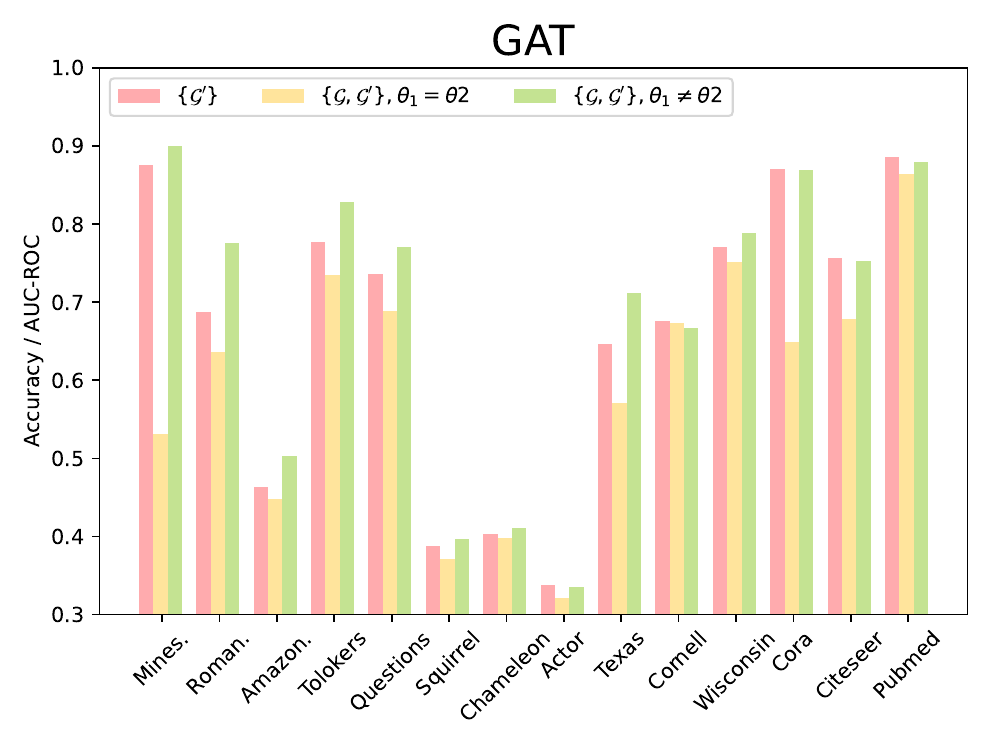}}
    \hfill
    \vspace{-0.7cm}
  \subfloat{%
       \includegraphics[width=0.5\textwidth]{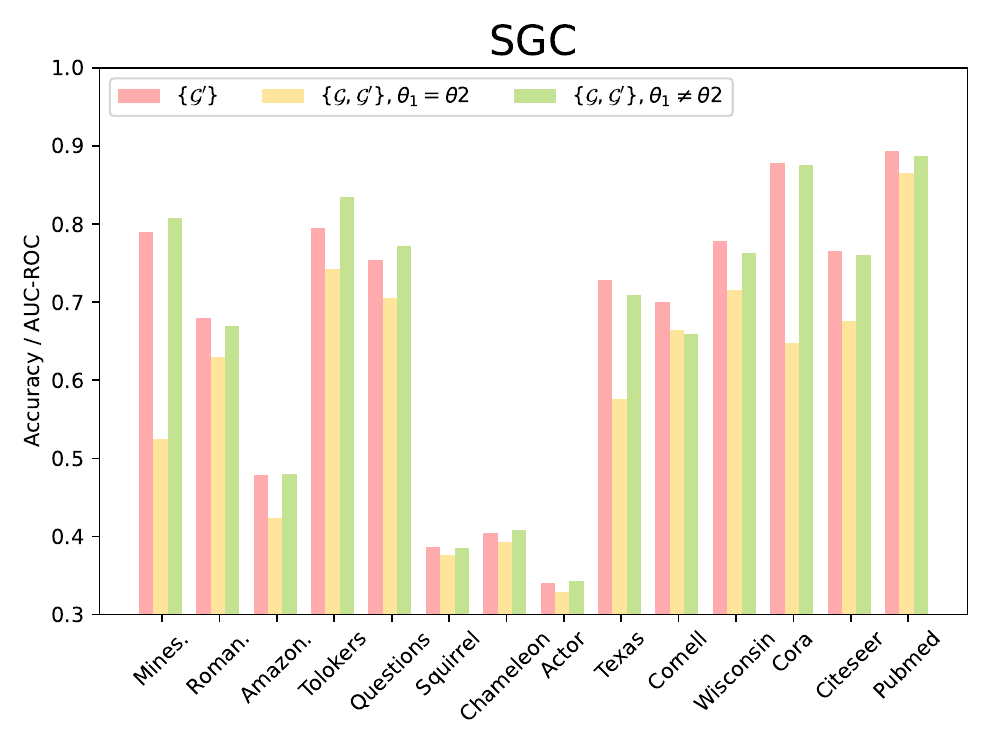}}
    \hfill
  \subfloat{%
        \includegraphics[width=0.5\textwidth]{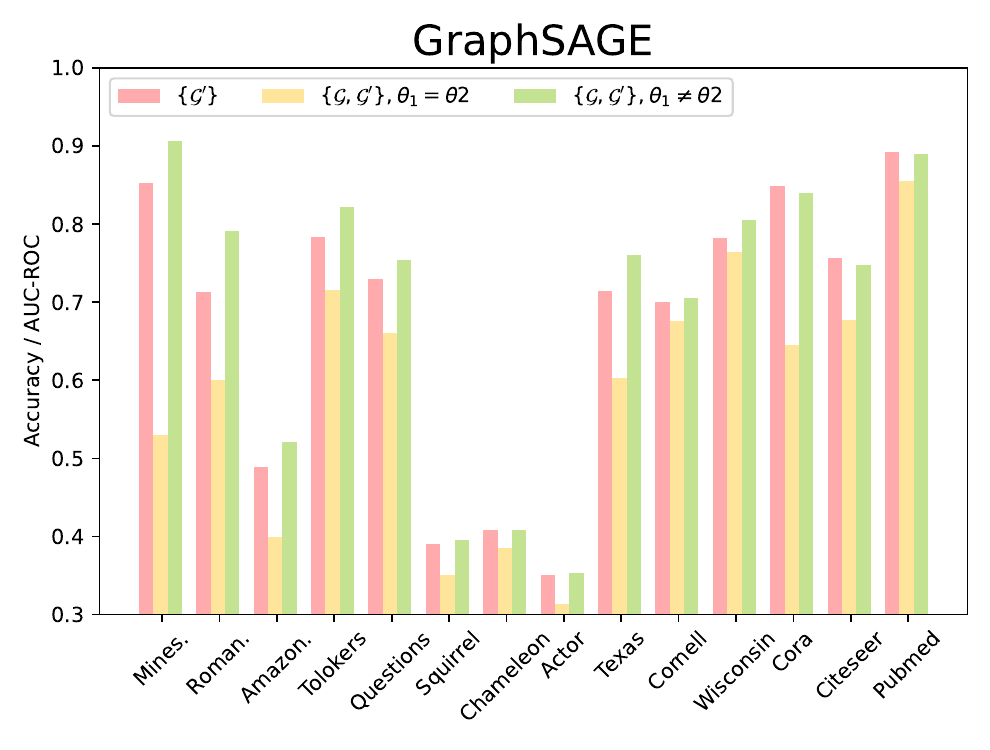}}
    \hfill
    % \hspace*{\fill}
    \vspace{-0.3cm}
    \caption{Influences of the approaches of view fusion in GSL to GNN+GSL.}
    \label{fig:apd_basis_gsl_fusion}
\end{figure}

Figure \ref{fig:apd_basis_gsl_fusion} illustrates the impact of different view fusion approaches, comparing the use of only the GSL graph $\mathcal{G}'$, the combination of the original graph $\mathcal{G}$ with $\mathcal{G}'$ using shared parameters $\theta_1 = \theta_2$, and the use of separate parameters $\theta_1 \neq \theta_2$. Notably, using only the GSL graph $\mathcal{G}'$ underperforms compared to employing both graph views with separate model parameters. This indicates that incorporating information from the original graph $\mathcal{G}$ is beneficial for maximizing GNN+GSL performance. Furthermore, for the two graph views, parameter sharing significantly underperforms parameter separation. We speculate that the messages aggregated under $\mathcal{G}$ and $\mathcal{G}'$ differ considerably, suggesting that different graphs should be treated with distinct model parameters.

\subsection{Fusion Stage}
\begin{figure}[b]
    \centering
  \subfloat{%
       \includegraphics[width=0.5\textwidth]{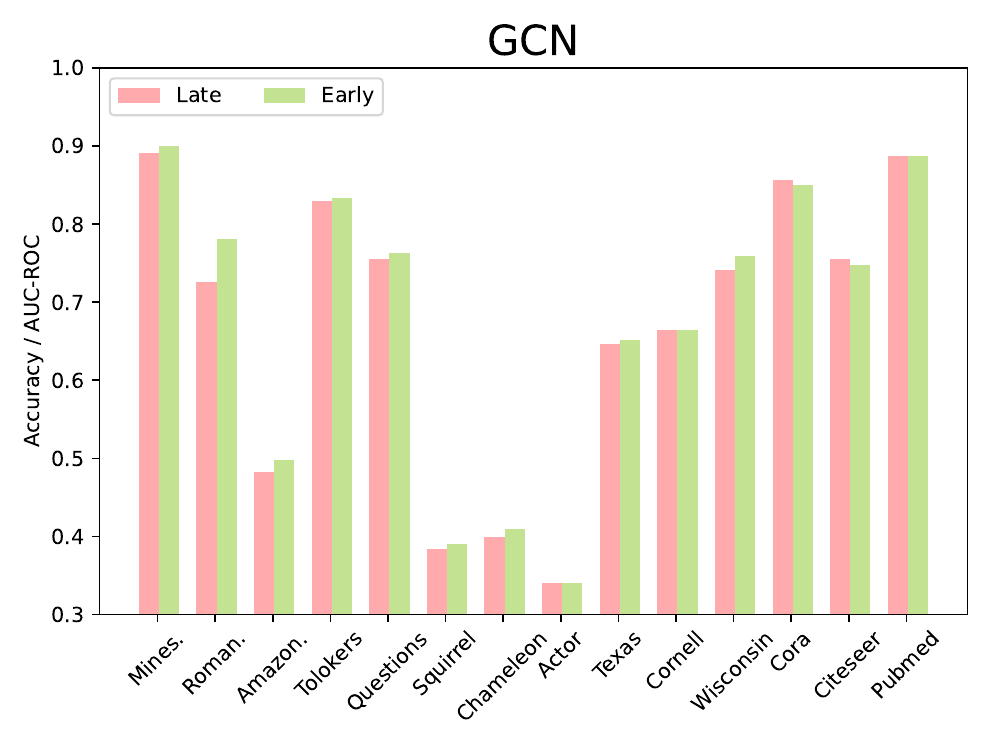}}
    \hfill
  \subfloat{%
        \includegraphics[width=0.5\textwidth]{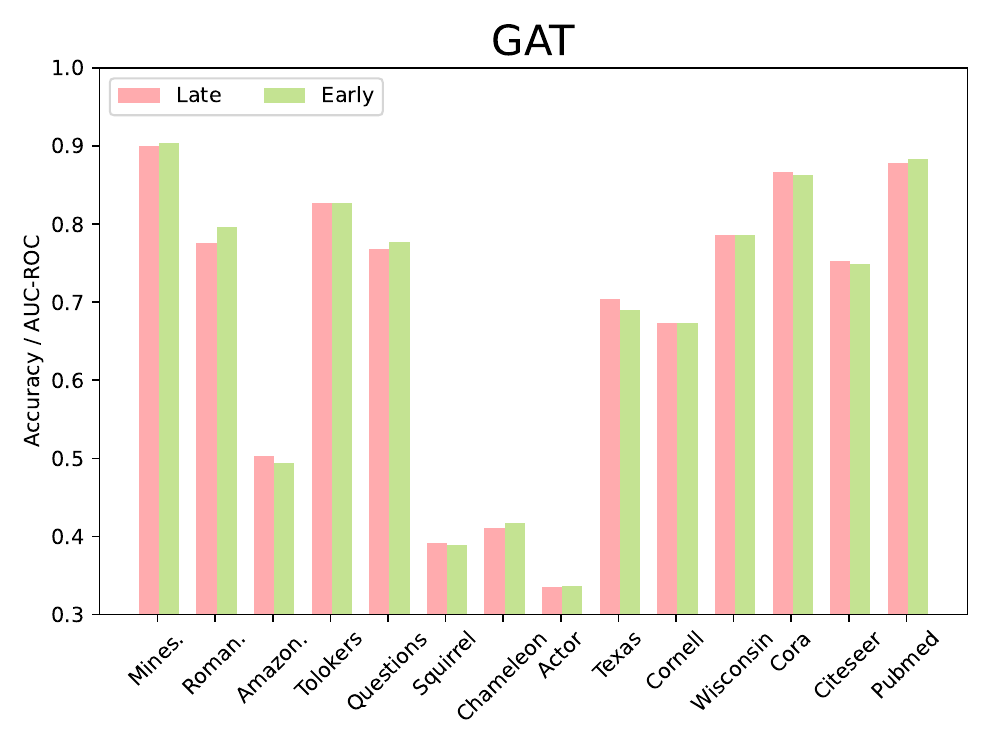}}
    \hfill
    \vspace{-0.7cm}
  \subfloat{%
       \includegraphics[width=0.5\textwidth]{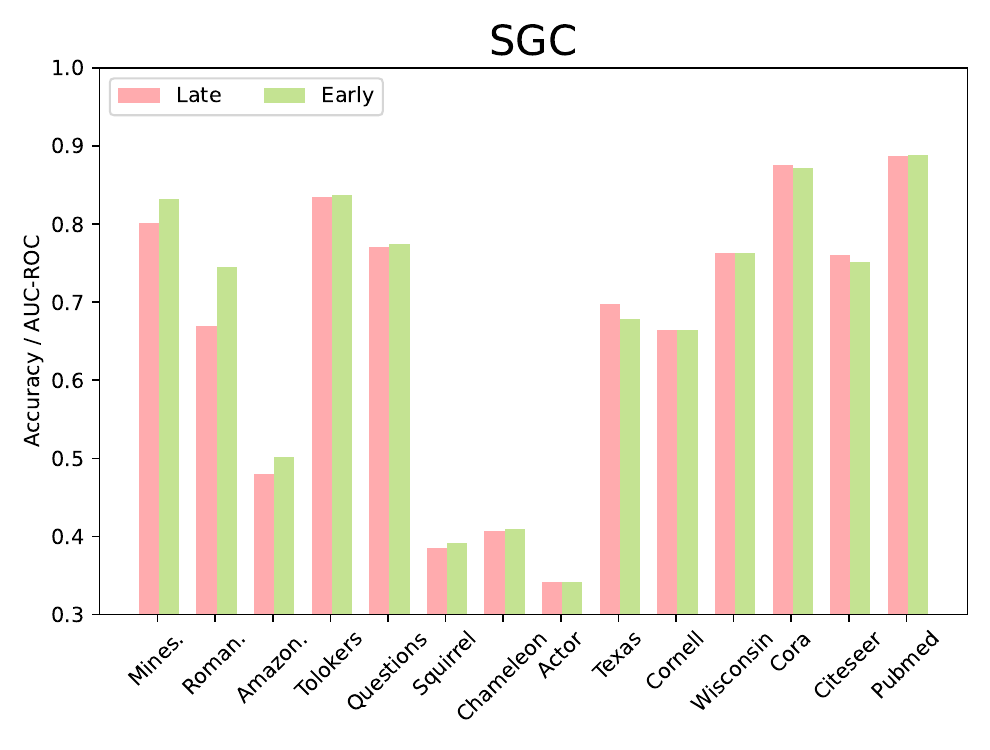}}
    \hfill
  \subfloat{%
        \includegraphics[width=0.5\textwidth]{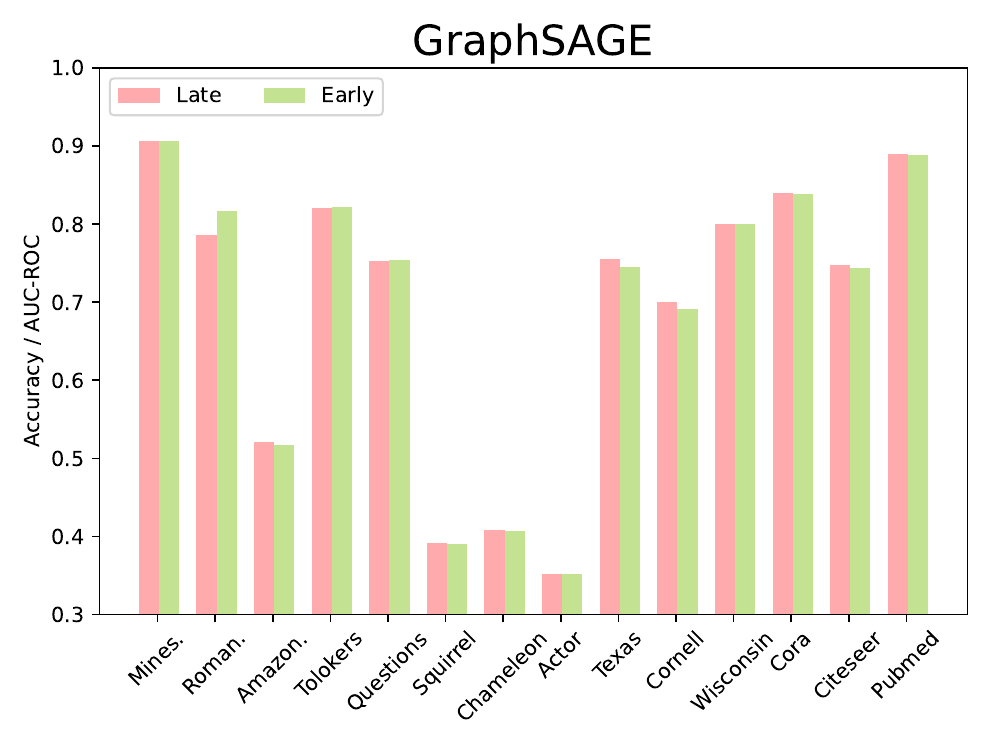}}
    \hfill
    % \hspace*{\fill}
    \vspace{-0.3cm}
    \caption{Influences of the states of view fusion in GSL to GNN+GSL.}
    \label{fig:apd_basis_gsl_fusion_state}
\end{figure}

Figure \ref{fig:apd_basis_gsl_fusion_state} compares early fusion and late fusion for GNN+GSL with multiple graph views. The performance difference between the two fusion states is often minimal. While early fusion tends to perform slightly better on complex datasets like Actor and Pubmed, the overall impact of switching between early and late fusion is limited across most datasets. For simpler datasets like Minesweeper and Amazon, both fusion methods yield nearly identical performance, indicating that the choice of fusion state does not drastically alter the model's outcome in most cases.

% Impact of different GSL components

\end{document}